%% file: _main.tex
\input{_constants}
\arxiv

\documentclass[10pt,twocolumn,letterpaper]{article}
\input{cvpr_header}
\input{_math_commands}

\begin{document}
\title{\paperTitle}
\author{\authorBlock}
\maketitle

\input{00_abstract}
\input{01_intro}
\input{02_related}
\input{03_method}
\input{04_analysis}
\input{05_results}
\input{10_conclusion}
\input{14_acknowledgements}

{\small
\bibliographystyle{ieee_fullname}
\bibliography{11b_shortstrings,11a_references}
}

\ifarxiv \clearpage \input{12_appendix} \fi

\end{document}

%% file: _constants.tex
\def\cvprPaperID{10889}
\def\confName{CVPR}
\def\confYear{2023}

\def\paperTitle{Independent Component Alignment for Multi-Task Learning}

\def\authorBlock{
    Dmitry Senushkin\qquad
    Nikolay Patakin\qquad
    Arseny Kuznetsov \qquad 
    Anton Konushin \\
    Samsung Research \\
    {\tt\small \{d.senushkin, n.patakin, a.konushin\}@samsung.com}
}

\newif\ifreview 
\newif\ifarxiv \newcommand{\arxiv}{\arxivtrue}
\newif\ifcamera 
\newif\ifrebuttal 

%% file: cvpr_header.tex
\ifreview \usepackage[review]{cvpr} \fi
\ifarxiv \usepackage[pagenumbers]{cvpr} \fi
\ifrebuttal \usepackage[rebuttal]{cvpr} \fi
\ifcamera \usepackage{cvpr} \fi

\usepackage{graphicx}
\usepackage{amsmath}
\usepackage{amssymb}
\usepackage{booktabs}

\input{_macros}  

\usepackage{xr-hyper}

\makeatletter
\newcommand*{\addFileDependency}[1]{
  \typeout{(#1)}
  \@addtofilelist{#1}
  \IfFileExists{#1}{}{\typeout{No file #1.}}
}

\makeatother

\crefname{section}{Sec.}{Secs.}
\crefname{table}{Table}{Tables}
\crefname{figure}{Fig.}{Figs.}

%% file: _macros.tex
\usepackage{times}
\usepackage{microtype}
\usepackage{epsfig}
\usepackage[table,xcdraw]{xcolor}
\usepackage{tikz,pgfplots}
\usepackage{tkz-euclide}
\usepackage{environ}
\usepackage{adjustbox}
\usepackage{tabularx}
\usepackage{enumitem}
\usepackage{xpatch}
\usepackage{colortbl}
\usepackage{caption}
\usepackage{float}
\usepackage{placeins}
\usepackage{color}
\usepackage{stfloats}
\usepackage{enumitem}
\usepackage{tabularx}
\usepackage{xstring}
\usepackage{multirow}
\usepackage{xspace}
\usepackage{url}
\usepackage{subcaption}
\usepackage{xcolor}
\usepackage{algorithm}
\usepackage{algpseudocode}
\usepackage{svg}
\usepackage{xr-hyper}
\usepackage{hyperref}
\usepackage{rotating}
\usepackage{graphicx}
\usepackage{amsmath}
\usepackage{amssymb}
\usepackage{booktabs}
\usepackage{xcolor}
\usepackage{ntheorem}
\usepackage[hang,flushmargin]{footmisc}
\usepackage[accsupp]{axessibility}

\usepgfplotslibrary{groupplots,dateplot}
\usetikzlibrary{patterns,shapes.arrows}
\usetikzlibrary{shapes.geometric}
\usepgfplotslibrary{colormaps}
\pgfplotsset{compat=newest}
\pgfkeys{/pgf/number format/.cd,1000 sep={}}

\usepackage[
    separate-uncertainty=true, 
    output-exponent-marker = \text{e},
    exponent-product={},
    binary-units=true
]{siunitx}

\colorlet{mygray}{black!10}
\colorlet{darkgray}{black!40}
\colorlet{myorange}{orange!80}
\newcommand{\maxf}[1]{{\cellcolor{mygray}} #1}
\definecolor{myyellow}{rgb}{0.858, 0.6, 0.1}
\definecolor{mydarkred}{rgb}{0.69,0.0,0.1098}
\definecolor{CornflowerBlue}{HTML}{0071BC}

\ifcamera \usepackage[accsupp]{axessibility} \fi

% ...

\usepackage[capitalize,nameinlink]{cleveref}
\crefname{section}{Sec.}{Sects.}
\crefname{proposition}{Proposition}{Propositions}
\crefname{lemma}{Lemma}{Lemmas}
\crefname{model}{Model}{Models}
\crefname{definition}{Def.}{Defs.}
\crefname{appendix}{App.}{Apps.}
\crefname{algorithm}{Alg.}{Algs.}

\newcommand{\nbf}[1]{{\noindent \textbf{#1.}}}

\newtheorem{theorem}{Theorem}

\newtheorem*{proof-non}{Proof}
\newtheorem{definition}{Definition}

\newtheorem{lemma}{Lemma}
\newtheorem{collorary}{Collorary}

\ifarxiv  \fi

\newcommand{\R}[1]{{%
    \textbf{%
        \ifstrequal{#1}{2RCZ}{\textcolor{red}{#1}}{%
        \ifstrequal{#1}{ckvx}{\textcolor{blue}{#1}}{%
        \ifstrequal{#1}{h3JY}{\textcolor{orange}{#1}}{%
        \ifstrequal{#1}{P5SY}{\textcolor{cyan}{#1}}{%
        }}}}%
    }%
}}

\algdef{SE}{Begin}{End}{\textbf{begin}}{\textbf{end}}

%% file: _math_commands.tex
\usepackage{amsmath,amsfonts,bm}

\DeclareMathOperator{\rank}{rank}
\DeclareMathOperator{\diag}{diag}  

\newcommand{\defeq}{\stackrel{\text{def}}{=}}

\def\lbr{\langle\hspace{0.5mm}}
\def\rbr{\hspace{0.5mm}\rangle}

\def\ctan{\qopname\relax o{ctan}}

\def\calL{{\mathcal{L}}}

\def\eqref#1{equation~\ref{#1}}

\def\1{\bm{1}}

\def\valpha{{\bm{\alpha}}}
\def\vtheta{{\bm{\theta}}}
\def\vdelta{{\bm{\delta}}}

\def\vg{{\bm{g}}}

\def\vr{{\bm{r}}}

\def\vu{{\bm{u}}}
\def\vv{{\bm{v}}}
\def\vw{{\bm{w}}}
\def\vx{{\bm{x}}}

\def\mB{{\bm{B}}}

\def\mD{{\bm{D}}}

\def\mG{{\bm{G}}}
\def\mH{{\bm{H}}}
\def\mI{{\bm{I}}}
\def\mJ{{\bm{J}}}

\def\mM{{\bm{M}}}

\def\mU{{\bm{U}}}
\def\mV{{\bm{V}}}

\def\mZ{{\bm{Z}}}

\def\mSigma{{\bm{\Sigma}}}

\DeclareMathAlphabet{\mathsfit}{\encodingdefault}{\sfdefault}{m}{sl}
\SetMathAlphabet{\mathsfit}{bold}{\encodingdefault}{\sfdefault}{bx}{n}



%% file: 00_abstract.tex
\begin{abstract}
In a multi-task learning (MTL) setting, a single model is trained to tackle a diverse set of tasks jointly. Despite rapid progress in the field, MTL remains challenging due to optimization issues such as conflicting and dominating gradients. In this work, we propose using a condition number of a linear system of gradients as a stability criterion of an MTL optimization. We theoretically demonstrate that a condition number reflects the aforementioned optimization issues. Accordingly, we present Aligned-MTL, a novel MTL optimization approach based on the proposed criterion, that eliminates instability in the training process by aligning the orthogonal components of the linear system of gradients. While many recent MTL approaches guarantee convergence to a minimum, task trade-offs cannot be specified in advance. In contrast, Aligned-MTL provably converges to an optimal point with pre-defined task-specific weights, which provides more control over the optimization result. Through experiments, we show that the proposed approach consistently improves performance on a diverse set of MTL benchmarks, including semantic and instance segmentation, depth estimation, surface normal estimation, and reinforcement learning. The source code is publicly available at \href{https://github.com/SamsungLabs/MTL}{https://github.com/SamsungLabs/MTL}.
\end{abstract}

%% file: 01_intro.tex
\section{Introduction}
In a multi-task learning (MTL), several tasks are solved jointly by a single model~\cite{caruana1993,Doersch17MTL}. In such a scenario, information can be shared across tasks, which may improve the generalization and boost the performance for all objectives. Moreover, MTL can be extremely useful when computational resources are constrained, so it is crucial to have a single model capable of solving various tasks~\cite{Liu19MTLNLP,Kendall15Posenet,Kokkinos17Ubernet}. In reinforcement learning~\cite{Parisotto16Actor-Mimic,Teh17DistralRL}, MTL setting arises naturally, when a single agent is trained to perform multiple tasks.

Several MTL approaches~\cite{liu2019DWA, misra2016crossstich,Liu19MTLwAtt, li2022url, houlsby19nlpICML,pfeiffer2021AdapterFusion,Lu202012in1} focus on designing specific network architectures and elaborate strategies of sharing parameters and representations across tasks for a given set of tasks. Yet, such complicated and powerful models are extremely challenging to train. 

Direct optimization of an objective averaged across tasks might experience issues~\cite{tianhe2020pcgrad} related to conflicting and dominating gradients. Such gradients destabilize the training process and degrade the overall performance. Accordingly, some other MTL approaches address these issues with multi-task gradient descent: either using gradient altering~\cite{tianhe2020pcgrad, sener2018MGDAUB, desideri2012MGDA, liu2021_imtl} or task balancing~\cite{liu2019DWA, kendall2018, lin2021RLW}. Many recent MTL methods~\cite{sener2018MGDAUB, liu2021_imtl, navon22a_nashmtl} guarantee convergence to a minimum, yet task trade-offs cannot be specified in advance. 
Unfortunately, the lack of control over relative task importance may cause some tasks to be compromised in favor of others~\cite{navon22a_nashmtl}.

In this work, we analyze the multi-task optimization challenges from the perspective of stability of a linear system of gradients. Specifically, we propose using a \textit{condition number} of a linear system of gradients as a stability criterion of an MTL optimization. According to our thorough theoretical analysis, there is a strong relation between the condition number and conflicting and dominating gradients issues. We exploit this feature to create Aligned-MTL, a novel gradient manipulation approach, which is the major contribution of this work. Our approach resolves gradient conflicts and eliminates dominating gradients by aligning principal components of a gradient matrix, which makes the training process more stable. In contrast to other existing methods (\eg~\cite{navon22a_nashmtl, sener2018MGDAUB, tianhe2020pcgrad, liu2021_imtl}), Aligned-MTL has a provable guarantee of convergence to an optimum with pre-defined task weights. 

We provide an in-depth theoretical analysis of the proposed method and extensively verify its effectiveness. Aligned-MTL consistently outperforms previous methods on various benchmarks. First, we evaluate the proposed approach on the problem of scene understanding; specifically, we perform joint instance segmentation, semantic segmentation, depth and surface normal estimation on two challenging datasets -- Cityscapes~\cite{Cordts2016Cityscapes} and NYUv2~\cite{Silberman:ECCV12:nyuv2}. Second, we apply our method to multi-task reinforcement learning and conduct experiments with the MT10 dataset~\cite{yu2019metaworld}. Lastly, in order to analyze generalization performance, Aligned-MTL has been applied to two different network architectures, namely PSPNet~\cite{sener2018MGDAUB} and MTAN~\cite{liu2019DWA}, in the scene understanding experiments.

\label{sec:intro}

%% file: 02_related.tex
\section{Related Work}\label{sec:related}
\input{figs/toy_1b}

A multi-task setting~\cite{caruana1993,ruder17MTLSurvey,crawshaw2020MTLSurvey} is leveraged in computer vision \cite{bilen2016,Kokkinos17Ubernet,zamir2018,nekrasov2019,kendall2018}, natural language processing \cite{collobert2008,luong2015,dong2015}, speech processing \cite{seltzer2013}, and robotics \cite{wulfmeier2020, levine2016} applications.
Prior MTL approaches formulate the total objective as a weighted sum of task-specific objectives, with weights being manually tuned~\cite{Kendall15Posenet,Laskar17Rel2Abs,Melekhov17Loc}. However, finding optimal weights via grid search is computationally inefficient. Kendall et al.~\cite{kendall2018} overcame this limitation, assigning task weights according to the homoscedastic uncertainty of each task. Other recent methods, such as GradNorm~\cite{chen2018GradNorm} and DWA~\cite{liu2019DWA}, optimize weights based on task-specific learning rates or by random weighting~\cite{lin2021RLW}.

The most similar to Aligned-MTL approaches (\eg~\cite{tianhe2020pcgrad, desideri2012MGDA, liu2021_imtl, liu2021cagrad, navon22a_nashmtl}) aim to mitigate effects of conflicting or dominating gradients. Conflicting gradients having opposing directions often induce a negative transfer (\eg~\cite{Hae18DAMTL}). Among all approaches tackling this problem, the best results are obtained by those based on an explicit gradient modulation~\cite{liu2021cagrad,tianhe2020pcgrad,liu2021_imtl} where a gradient of a task which conflicts with a gradient of some other task is replaced with a modified, non-conflicting, gradient. Specifically, PCGrad~\cite{tianhe2020pcgrad} proposes a "gradient surgery" which decorrelates a system of vectors, while CAGrad~\cite{liu2021cagrad} aims at finding a conflict-averse direction to minimize overall conflicts. GradDrop~\cite{chen2020graddrop} forces task gradients sign consistency. Other methods also address an issue of dominating gradients. Nash-MTL~\cite{navon22a_nashmtl} leverages advances of game theory~\cite{navon22a_nashmtl}, while IMTL~\cite{liu2021_imtl} searches for a gradient direction where all the cosine similarities are equal.

Several recent works~\cite{peitz2018,poirion2017} investigate a multiple-gradient descent algorithm (MGDA~\cite{desideri2012MGDA,fliege2000,schaffler2002}) for MTL: these methods search for a direction that decreases all objectives according to multi-objective Karush--Kuhn--Tucker (KKT) conditions~\cite{kuhn1951}. Sener and Koltun \cite{sener2018MGDAUB} propose extending the classical MGDA~\cite{desideri2012MGDA} so it scales well to high-dimensional problems for a specific use case. However, all the described approaches converge to an arbitrary Pareto-stationary solution, leading to a risk of imbalanced task performance.

%% file: figs/toy_1b.tex
\begin{figure*}[t!]
\centering

\begin{subfigure}[t]{.19\textwidth}
  \centering
  \includegraphics[width=1.0\textwidth]{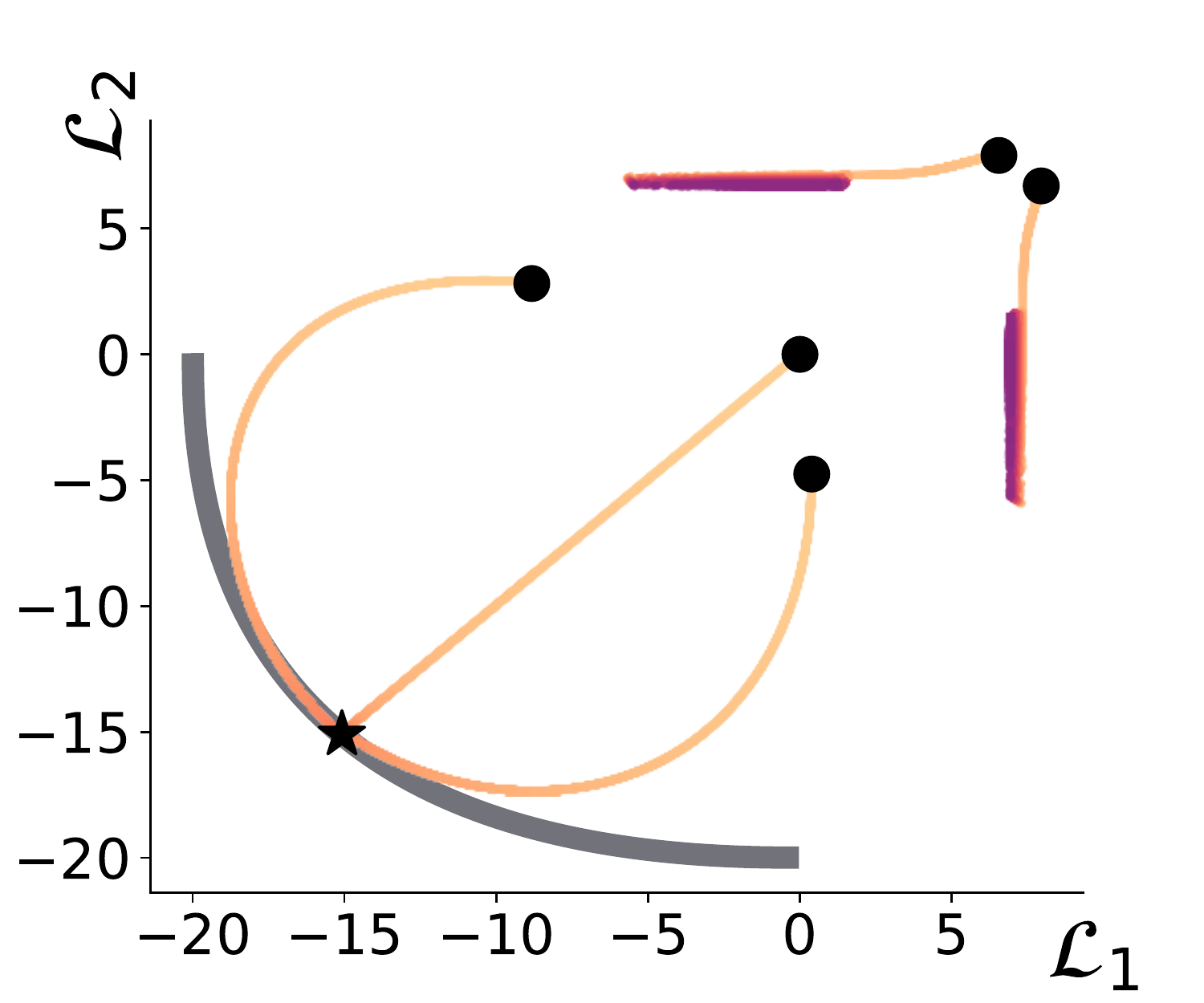}
\end{subfigure}
\hfill
\begin{subfigure}[t]{.19\textwidth}
  \centering
  \includegraphics[width=1.0\textwidth]{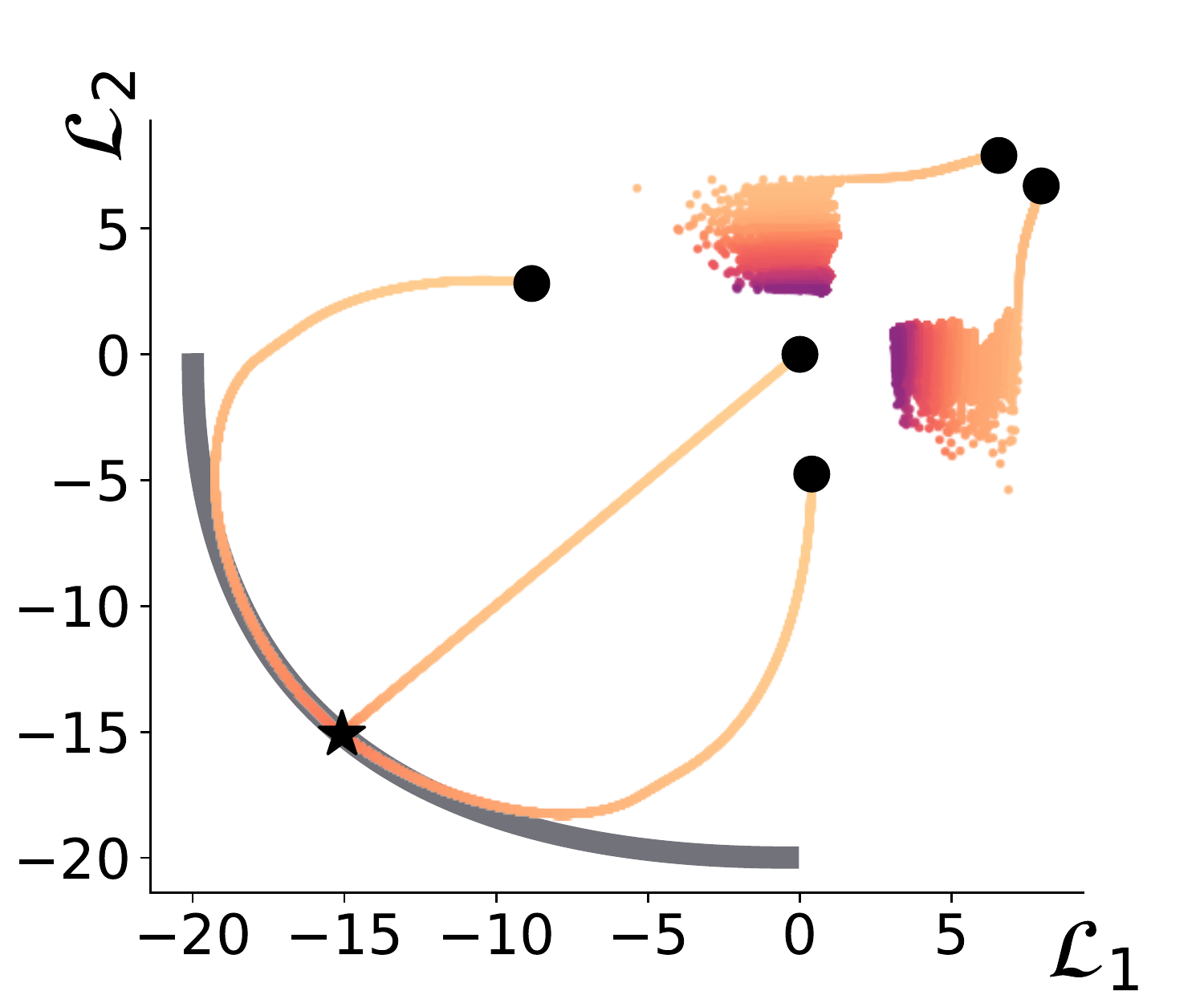}
\end{subfigure}
\hfill
\begin{subfigure}[t]{.19\textwidth}
  \centering
  \includegraphics[width=1.0\textwidth]{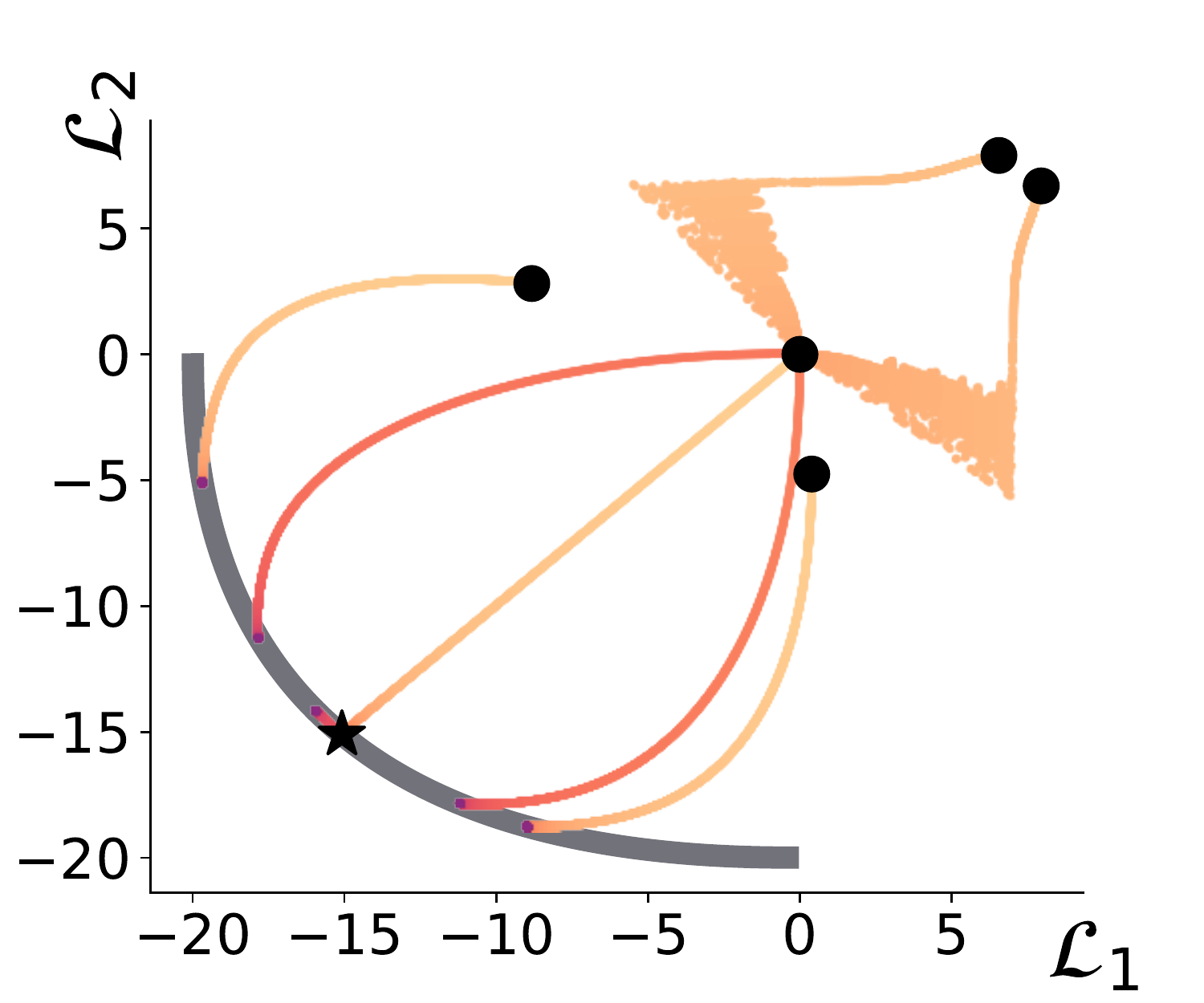}
\end{subfigure}
\hfill
\begin{subfigure}[t]{.19\textwidth}
  \centering
  \includegraphics[width=1.0\textwidth]{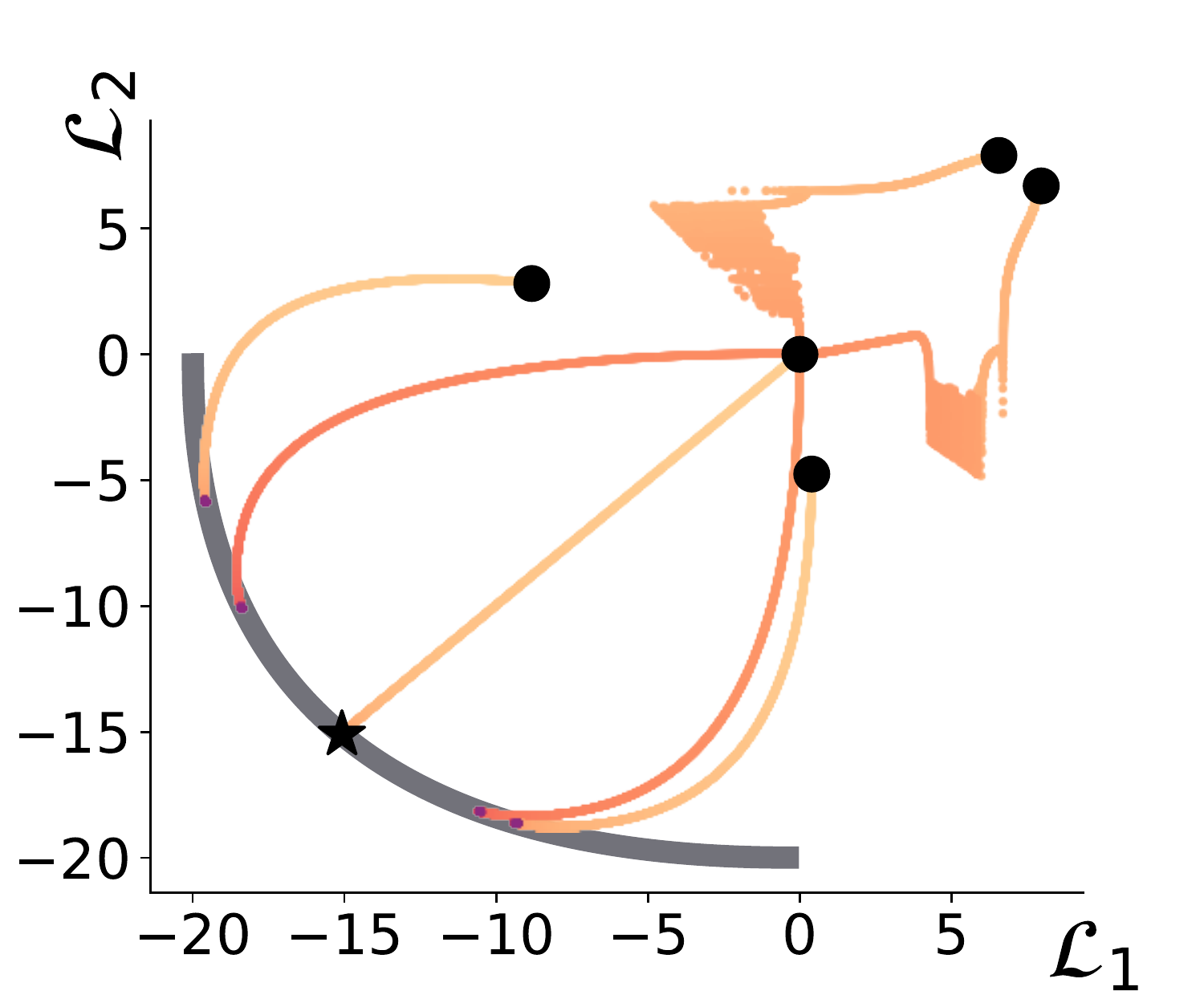}
\end{subfigure}
\begin{subfigure}[t]{.19\textwidth}
  \centering
  \includegraphics[width=1.0\textwidth]{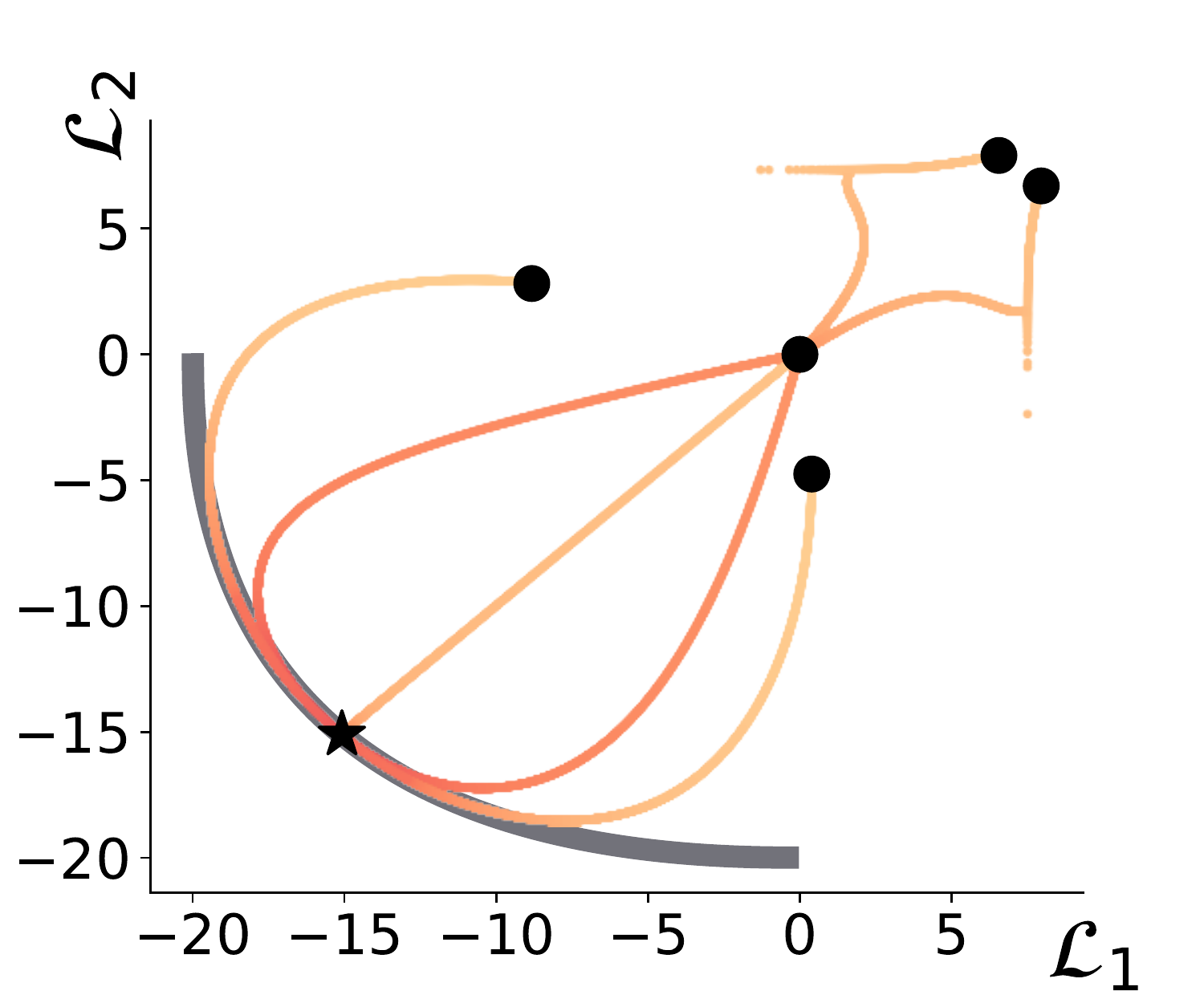}
\end{subfigure}

\begin{subfigure}[t]{\textwidth}
    \centering
\end{subfigure}

\begin{subfigure}[t]{.195\textwidth}
  \centering
  \includegraphics[clip, width=1.0\textwidth, trim=1.5cm 0cm 0.5cm 1.5cm]{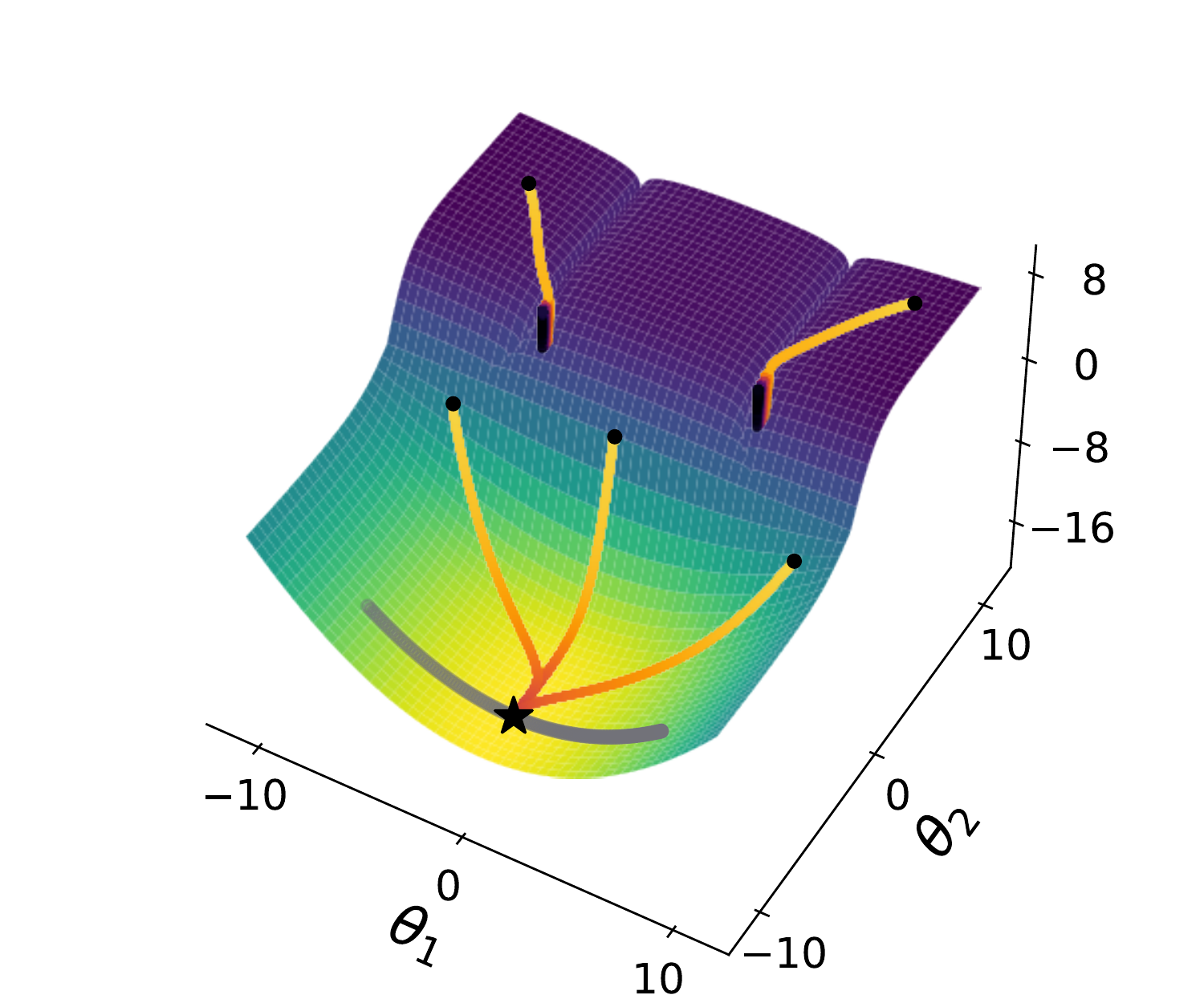}
\captionsetup{font={footnotesize}}
  \caption{Uniform}
  \label{fig:teaser:uniform}
\end{subfigure}
\hfill
\begin{subfigure}[t]{.195\textwidth}
  \centering
  \includegraphics[clip, width=1.0\textwidth, trim=1.5cm 0cm 0.5cm 1.5cm]{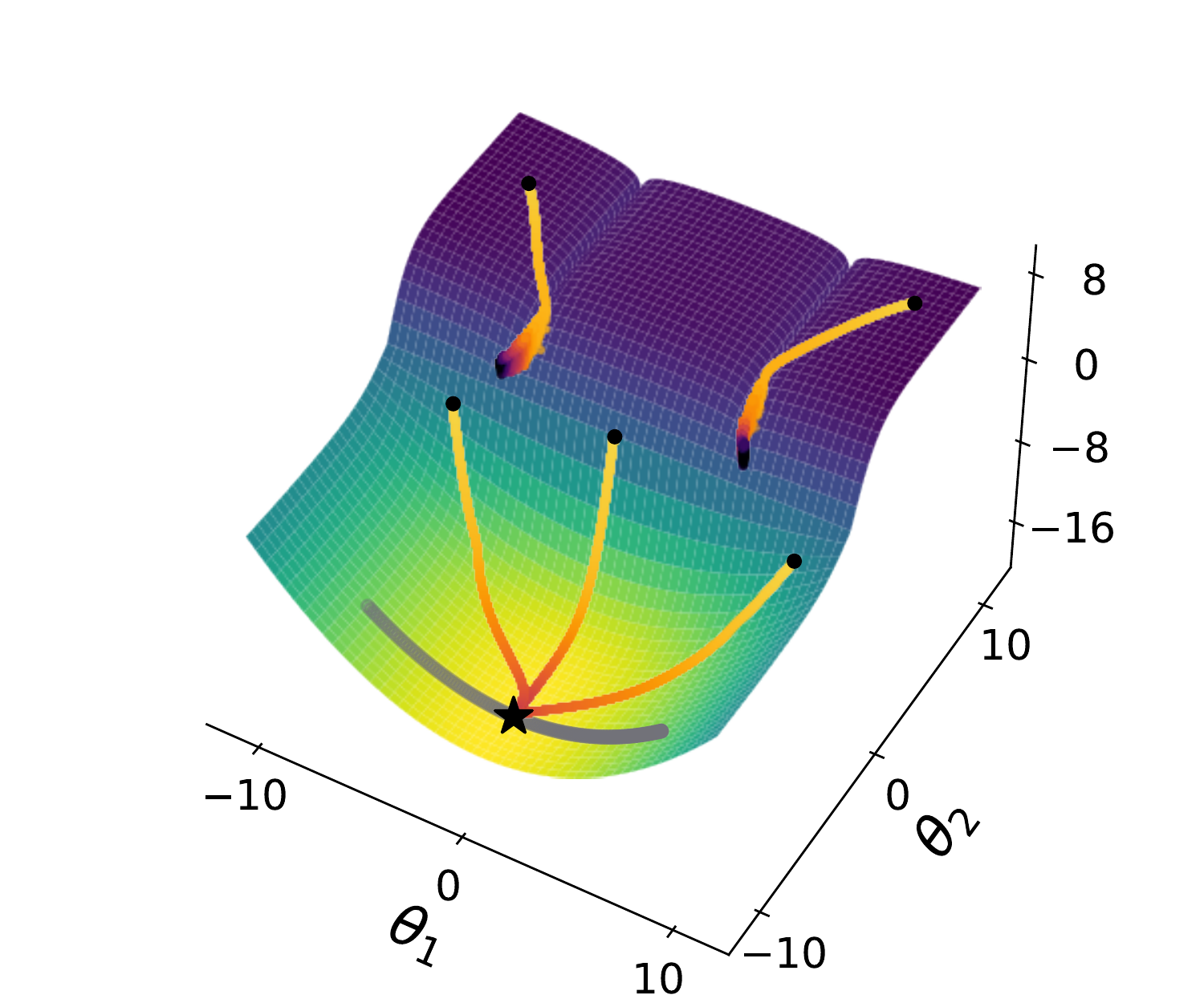}
\captionsetup{font={footnotesize}}
  \caption{CAGrad ($c=0.4$)~\cite{liu2021cagrad}}
  \label{fig:teaser:cagrad}
\end{subfigure}
\hfill
\begin{subfigure}[t]{.195\textwidth}
  \centering
  \includegraphics[clip, width=1.0\textwidth, trim=1.5cm 0cm 0.5cm 1.5cm]{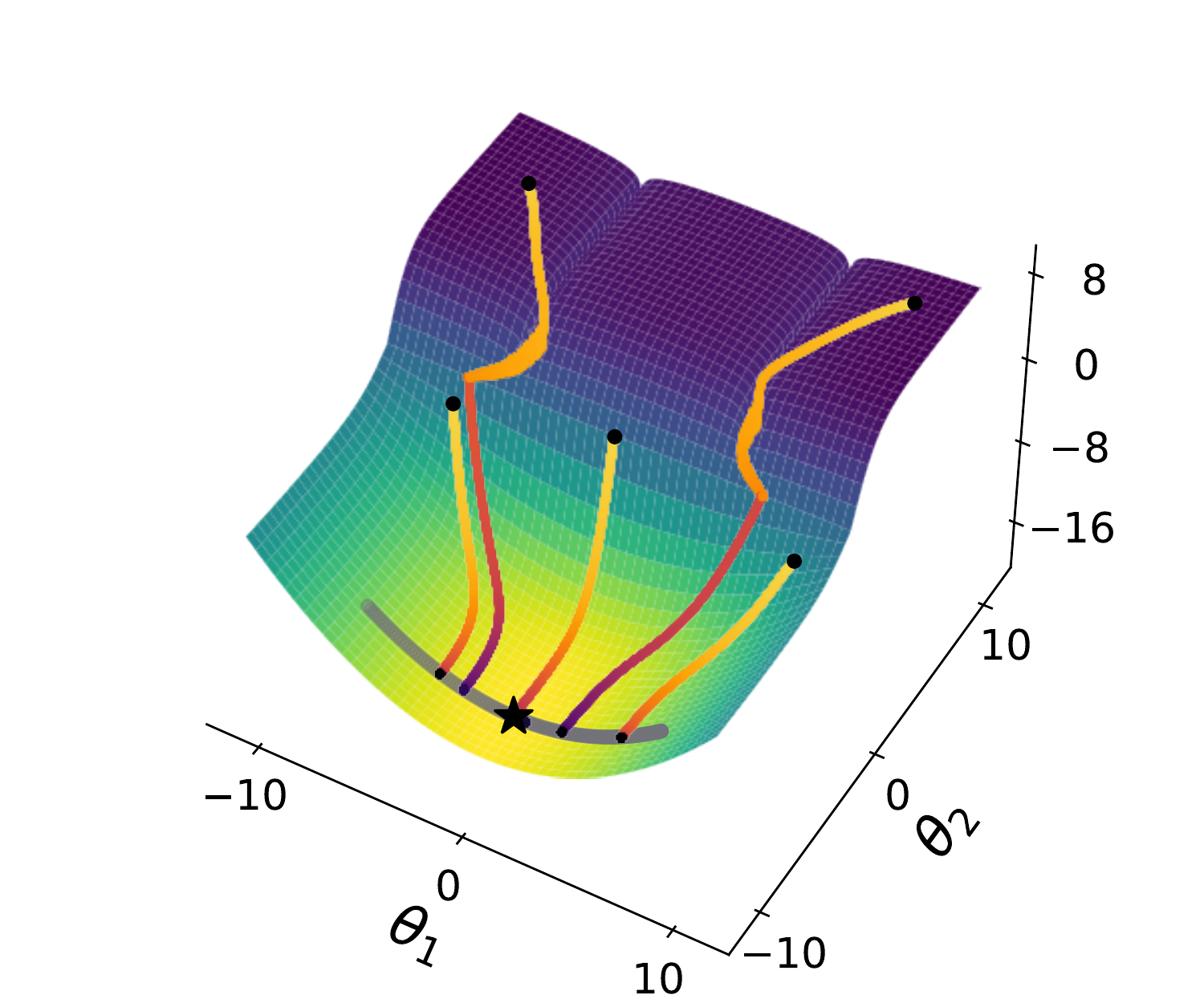}
\captionsetup{font={footnotesize}}
  \caption{IMTL~\cite{liu2021_imtl}}
  \label{fig:teaser:imtl}
\end{subfigure}
\hfill
\begin{subfigure}[t]{.195\textwidth}
  \centering
  \includegraphics[clip, width=1.0\textwidth, trim=1.5cm 0cm 0.5cm 1.5cm]{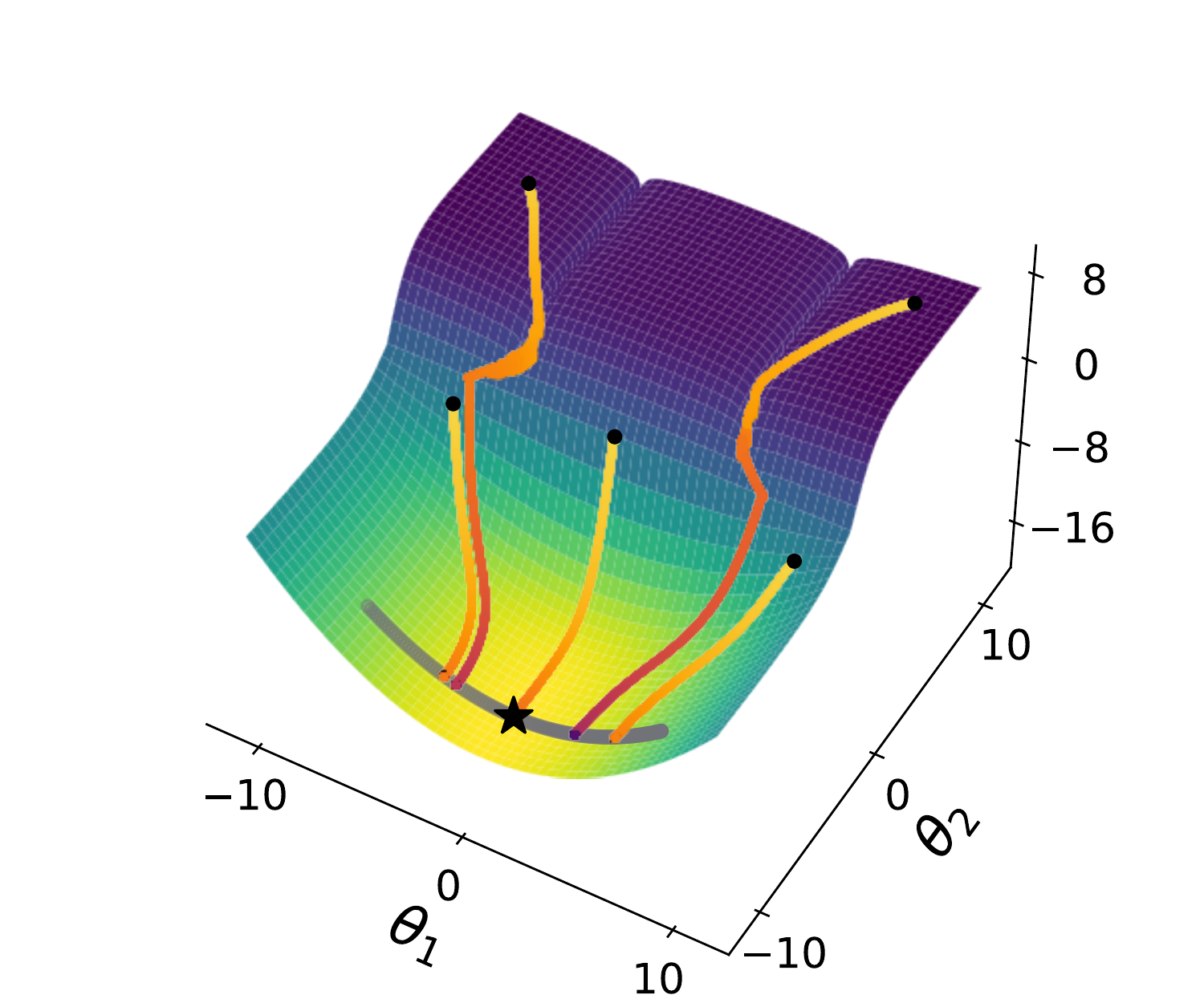}
\captionsetup{font={footnotesize}}
  \caption{Nash-MTL~\cite{navon22a_nashmtl}}
  \label{fig:teaser:nashmtl}
\end{subfigure}
\begin{subfigure}[t]{.195\textwidth}
  \centering
  \includegraphics[clip, width=1.0\textwidth, trim=1.5cm 0cm 0.5cm 1.5cm]{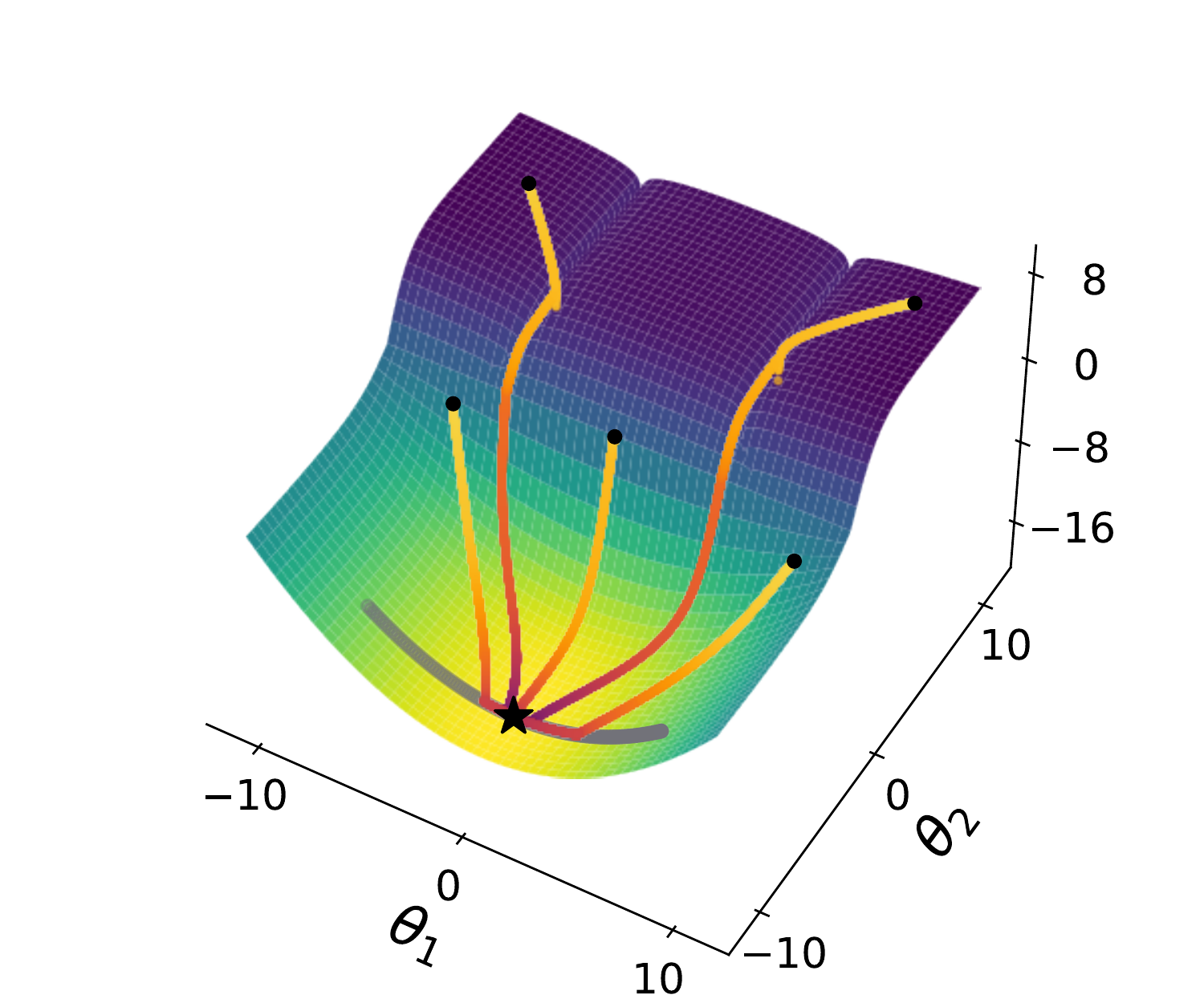}
\captionsetup{font={footnotesize}}
  \caption{Aligned-MTL (ours)}
\end{subfigure}

\caption{Comparison of MTL approaches on a challenging synthetic two-task benchmark~\cite{liu2021cagrad, navon22a_nashmtl}. We visualize optimization trajectories \wrt objectives value ($\mathcal{L}_1$ and $\mathcal{L}_2$, top row), and cumulative objective \wrt parameters ($\theta_1$ and $\theta_2$, bottom row). Initialization points are marked with $\bullet$, the Pareto front (\cref{def:pareto}) is denoted as \protect\tikz[baseline=-.45ex]\protect\node[fill=darkgray,inner sep=0pt,rounded corners=1pt,minimum height=1.9pt,minimum width=15pt]{};. Other MTL approaches produce noisy optimization trajectories (\cref{fig:teaser:uniform,fig:teaser:cagrad,fig:teaser:imtl,fig:teaser:nashmtl}) inside areas with conflicting and dominating gradients~(\cref{fig:synthetic-description}). In contrast, our approach converges to the global optimum ($\bigstar$) robustly. Approaches aiming to find a Pareto-stationary solution (such as \cref{fig:teaser:imtl} and \cref{fig:teaser:nashmtl}) terminate once the Pareto front is first reached, as a result, they might provide a suboptimal solution. Differently, Aligned-MTL drifts along the Pareto front and provably converges to the optimum \wrt pre-defined tasks weights.}

\label{fig:teaser}
\end{figure*}

%% file: 03_method.tex
\section{Multi-Task Learning}

Multi-task learning implies optimizing a single model with respect to multiple objectives. The recent works \cite{tianhe2020pcgrad, desideri2012MGDA, Parisotto16Actor-Mimic, kendall2018} have found that this learning problem is difficult to solve by reducing it to a standard single-task approach. In this section, we introduce a general notation and describe frequent challenges arising in gradient optimization in MTL.

\subsection{Notation}

In MTL, there are $T > 1$ tasks. Each task is associated with an objective~$\calL_i(\vtheta)$ depending on a set of model parameters $\vtheta$ shared between all tasks. The goal of MTL training is to find a parameter $\vtheta$ that minimizes an average loss:
\begin{align}
    \label{eq:opt_task}
    \vtheta^* = \arg\min_{\theta \in \mathbb{R}^m}  \bigg\{\calL_0(\vtheta) \defeq \sum_{i=1}^{T} \frac{1}{T}\calL_i(\vtheta). \bigg\} 
\end{align}
We introduce the following notation: $\vg_i = \nabla \calL_i(\vtheta)$ -- individual task gradients; $\calL_0(\vtheta)$ -- a cumulative objective; $\mG = \{\vg_1, \cdots, \vg_T\}$ -- a gradient matrix; $w_i \defeq \frac{1}{T}$ -- pre-defined task weights. 
The task weights are supposed to be fixed. We omit task-specific parameters in our notation, since they are independent and not supposed to be balanced.

\begin{figure}
    \centering
    \includegraphics[width=0.45\textwidth]{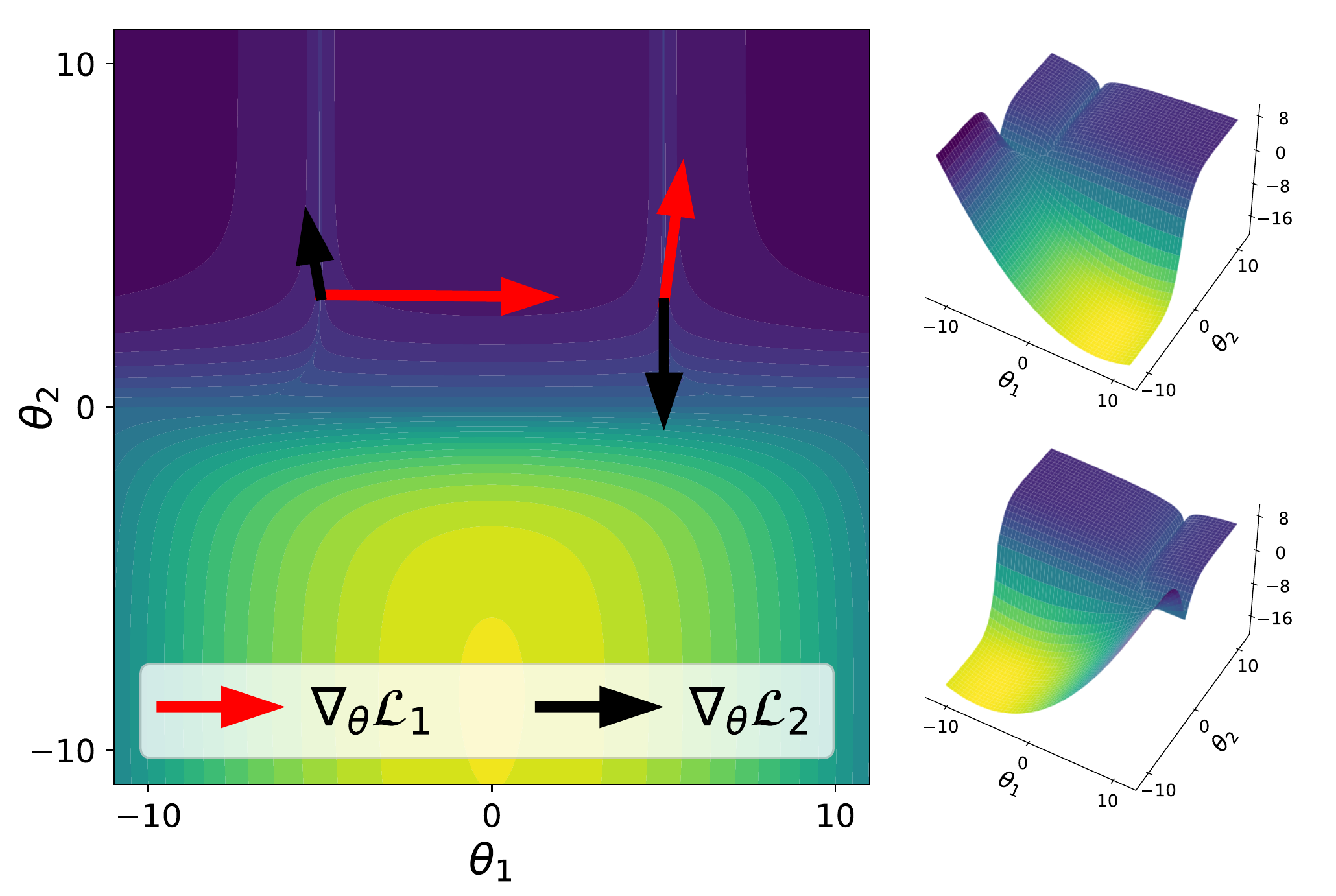}
    \caption{Synthetic two-task MTL benchmark~\cite{liu2021cagrad, navon22a_nashmtl}. Loss landscapes \wrt individual objectives are depicted on the right side. The cumulative loss landscape (on the left side) contains areas with conflicting and dominating gradeints. }
    \label{fig:synthetic-description}
\end{figure}

\subsection{Challenges}
In practice, directly solving a multi-objective optimization problem via gradient descent may significantly compromise the optimization of individual objectives~\cite{tianhe2020pcgrad}. Simple averaging of gradients across tasks makes a cumulative gradient biased towards the gradient with the largest magnitude, which might cause overfitting for a subset of tasks. Conflicting gradients with negative cosine distance complicate the training process as well; along with dominating gradients, they increase inter-step direction volatility that decreases overall performance (\cref{fig:teaser}). To mitigate the undesired effects of conflicting and dominating gradients in MTL, we propose a criterion that is strongly correlated with the presence of such optimization issues. This measure is a \textit{condition number} of a linear system of gradients.

\section{Stability}

The prevailing challenges in MTL are arguably task dominance and conflicting gradients, accordingly, various criteria for indicating and measuring these issues have been formulated. For instance, a gradient dominance can be measured with a gradient magnitude similarity (\cite{tianhe2020pcgrad} Def.~2). Similarly, gradient conflicts can be estimated as a cosine distance between vectors (\cite{tianhe2020pcgrad}~Def. 1, \cite{liu2021cagrad}). However, each of these metrics describes a specific characteristic of a linear system of gradients, and cannot provide a comprehensive assessment if taken separately. We show that our stability criterion indicates the presence of both MTL challenges~(\cref{fig:city:three:triad}); importantly, it describes a whole linear system and can be trivially measured on any set of gradients. Together with magnitude similarity and cosine distance, this criterion accurately describes the training process.

\subsection{Condition Number}

Generally, the stability of an algorithm is its sensitivity to an input perturbation, or, in other words, how much the output changes if an input gets perturbed. In numerical analysis, the stability of a linear system is measured by a \textit{condition number} of its matrix. In a multi-task optimization, a cumulative gradient is a linear combination of task gradients: $\vg = \mG\vw$. Thus, the stability of a linear system of gradients can be measured as the condition number of a gradient matrix $\mG$. The value of this stability criterion is equal to the ratio of the maximum and minimum singular values of the corresponding matrix: 
\begin{equation} \label{eq:cnumber}
    \kappa(\mG) = \frac{\sigma_{max}}{\sigma_{min}}.
\end{equation}

\vspace{-0.5cm}
\paragraph{Remark.} 
A linear system is well-defined if its condition number is equal to one, and ill-posed if it is non-finite. A standard assumption for multi-task optimization is that a gradient system is not ill-posed, \ie task gradients are linearly independent. In this work, we suppose that the linear independence assumption holds unless otherwise stated.

\subsection{Condition Number and MTL Challenges}
The dependence between the stability criterion and MTL challenges is two-fold. Let us consider a gradient system having a minimal condition number. According to the singular value decomposition theorem, its gradient matrix~$\hat{\mG}$ with $\kappa(\hat{\mG}) = 1$ must be orthogonal with equal singular values: 
\begin{equation}\label{eq:cnumber_mtl_challenges}
    \hat{\mG} = \mU\mSigma\mV^\top, \quad \text{where} \quad \mSigma = \sigma\mI
\end{equation}
Moreover, since $\mU, \mV$ matrices are orthonormal, individual task gradients norms are equal to $\sigma$. Thus, minimizing the condition number of the linear system of gradients leads to mitigating dominance and conflicts within this system. 

On the other hand, if an initial linear system of gradients is not well-defined, reducing neither gradient conflict nor dominance only does not guarantee minimizing a condition number. The stability criterion reaches its minimum iff both issues are solved jointly and gradients are orthogonal. This restriction eliminates positive task gradients interference (co-directed gradients may produce $\kappa > 1$), but it can guarantee the absence of negative interaction, which is essential for a stable training. Noisy convergence trajectories \wrt objectives values~(\cref{fig:teaser}, top row) indicate instability of the training process.

To demonstrate the relation between MTL challenges and our stability criterion, we conduct a synthetic experiment as proposed in~\cite{liu2021cagrad, navon22a_nashmtl}. There are two objectives to be optimized, and the optimization landscape contains areas with conflicting and dominating gradients. We compare our approach against recent approaches that do not handle stability issues, yielding noisy trajectories in problematic areas. By enforcing stability, our method performs well on the synthetic benchmark. 

\input{figs/method}

\section{Aligned-MTL}

We suppose that multi-task gradient optimization should successfully resolve the main MTL challenges: conflicts and dominance in gradient system. Unlike existing approaches~\cite{tianhe2020pcgrad, wang2021gradvac} that focus on directly resolving the optimization problems, we develop an algorithm that handles issues related to the stability of a linear system of gradients and accordingly addresses both gradient conflicts and dominance.

Specifically, we aim to find a cumulative gradient $\hat{g_0}$, so that $\| \vg_0 - \hat{\vg}_0 \|^2_2$ is minimal, while a linear system of gradients is stable $\big( \kappa(\hat{\mG}) = 1 \big)$. This constraint is defined up to an arbitrary positive scaling coefficient. Here, we assume $\sigma = 1$ for simplicity. By applying a triangle inequality to the initial problem, we derive $\| \vg_0 - \hat{\vg}_0 \|^2_2 \leq \| \mG - \hat{\mG} \|^2_F \|\vw\|^2_2$. 
Thereby, we consider the following optimization task: \begin{gather}\label{eq:amtl}
    \min_{\hat{\mG}}\| \mG - \hat{\mG} \|^2_F \quad
    \text{s.t.} \quad \hat{\mG}^\top \hat{\mG} = \mI
\end{gather}

\vspace{-0.1cm}
The stability criterion, a condition number, defines a linear system up to an arbitrary positive scale. To alleviate this ambiguity, we choose the largest scale that guarantees convergence to the optimum of an original problem~(\cref{eq:opt_task}): this is a minimal singular value of an initial gradient matrix $\sigma = \sigma_{min}(\mG) > 0$. The final linear system of gradients defined by $\hat{\mG}$ satisfies the optimality condition in terms of a condition number.

\subsection{Gradient Matrix Alignment}
The problem~\cref{eq:amtl} can be treated as a special kind of Procrustes problem~\cite{procrustes}. Fortunately, there exists a closed-form solution of this task. To obtain such a solution, we perform a singular value decomposition (SVD) and rescale singular values corresponding to principal components so that they are equal to the smallest singular value.

Technically, the matrix alignment can be performed in the \textit{parameter} space or in the \textit{task} space; being equivalent, these options have different computational costs. This duality is caused by SVD providing two different eigen decompositions of Gram matrices $\mG^\top \mG$ and $\mG \mG^\top$:
\begin{equation}
    \hat{\mG} = \sigma\mU\mV^\top = \sigma\underbrace{\mU\mSigma^{-1}\mU^\top}_{\text{Parameter space}} \mG = \sigma\mG \underbrace{\mV\mSigma^{-1}\mV^\top}_{\text{Task space}} 
\end{equation}

\vspace{-0.1cm}
We perform the gradient matrix alignment at each optimization step. Since the number of tasks $T$ is relatively small compared to the number of parameters, we operate in a task space. This makes a gradient matrix alignment more computationally efficient as we need to perform an eigen decomposition of a small $T \times T$ matrix. \cref{fig:geom_alg} provides a geometric interpretation of our approach, while pseudo-code is given in \cref{alg}.

\vspace{-0.5cm}
\paragraph{Remark.} If an initial matrix $\mG$ is singular (gradients are linear dependent), then the smallest singular value is zero. Fortunately, the singular value decomposition provides a unique solution even in this case; yet, we need to choose the smallest singular value greater than zero as a global scale.

\subsection{Aligned-MTL: Upper Bound Approximation}\label{sec:fast_alignment}
The major limitation of our approach is the need to run multiple backward passes through the shared part of the model to calculate the gradient matrix. The backward passes are computationally demanding, and the training time depends linearly on the number of tasks: if it is large, our approach may be non-applicable in practice.

This limitation can be mitigated for encoder-decoder networks, where each task prediction is computed using the same shared representation. We can employ the chain rule trick~\cite{sener2018MGDAUB} to upper-bound an original objective~(\cref{eq:amtl}):

\begin{equation}
    \| \mG - \hat{\mG} \|^2_F \leq \bigg\|\frac{\partial \mH}{\partial \theta} \bigg\|^2_F \| \mZ - \hat{\mZ} \|^2_F
\end{equation}
Here, $\mH$ stands for a hidden shared representation, and $\mZ$ and $\hat{\mZ}$ are gradients of objective w.r.t. a shared representation of the initial and aligned linear system of gradients, respectively. 
Thus, the gradient alignment can be performed for a shared representation:
\begin{gather}\label{eq:amtl-ub}
    \min_{\hat{\mZ}}\| \mZ - \hat{\mZ} \|^2_F \quad
    \text{s.t.} \quad \hat{\mZ}^\top \hat{\mZ} = \mI
\end{gather}

Aligning gradients of shared representation does not require additional backward passes, since matrix $\mZ$ is computed during a conventional backward pass. We refer to such an approximation of Aligned-MTL as to Aligned-MTL-UB. With $O(1)$ time complexity \wrt the number of tasks, this approximation tends to be significantly more efficient than the original Aligned-MTL having $O(T)$ time complexity. 

\begin{algorithm}[t!]
\caption{Gradient matrix alignment}\label{alg}
\begin{algorithmic}
\Require{$G \in \mathbb{R}^{|\theta| \times T}$ -- gradient matrix, \\ \quad \quad \quad \ $w \in \mathbb{R}^{T}$ -- task importance}
\
    
    \State{\color{CornflowerBlue}/* Compute task space Gram matrix */}
    \State {$\mM \gets \mG^ \top \mG$ }
    \State{\color{CornflowerBlue} /* Compute eigenvalues and eigenvectors of $M$ */}
    \State {$(\lambda, \mV) \gets eigh(\mM)$}
    \State {$\mSigma^{-1} \gets diag \left( \sqrt{\frac{1}{\lambda_1}}, \cdots \sqrt{\frac{1}{\lambda_R}} \right)$}
    \State {\color{CornflowerBlue}/* Compute balance transformation */}
    \State {$\mB \gets \sqrt{\lambda_R}\mV \mSigma^{-1} \mV^\top$}
    \State {$\valpha \gets \mB \vw$}
    \State {$\textbf{return} \ \mG\valpha$}

\end{algorithmic}
\end{algorithm}

\label{sec:method}

%% file: figs/method.tex
\begin{figure*}[t!]
  \huge
  \centering
  \def\scale{.17}
  \begin{subfigure}[t]{0.32\textwidth}  
    \centering
    \scalebox{\scale}{
        \includegraphics[scale=4.0]{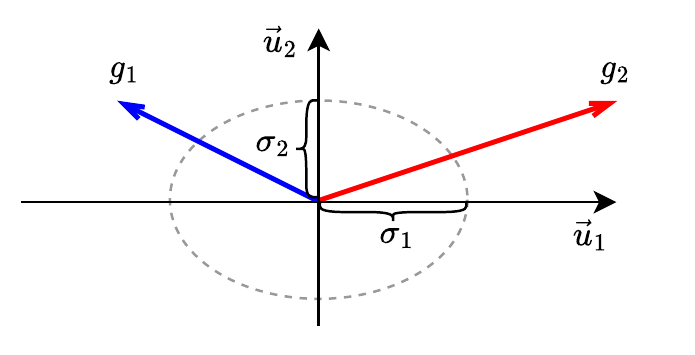}
    }
    \caption{Initial gradients}
    \label{fig:geom_alg:first}    
  \end{subfigure}
  \hfill
  \begin{subfigure}[t]{0.32\textwidth}    
    \centering
    \scalebox{\scale}{
        \includegraphics[clip, scale=4.0, trim=0cm 0.25cm 0cm 0cm]{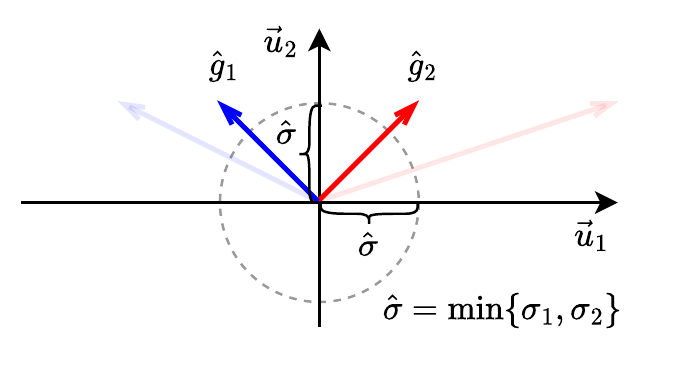}
    }
    \caption{Gradients aligned via Aligned-MTL}    
    \label{fig:geom_alg:second}    
  \end{subfigure}    
  \hfill
  \begin{subfigure}[t]{0.32\textwidth}
    \centering 
    \scalebox{\scale}{
        \includegraphics[clip, scale=4.0, trim=0cm 0.5cm 0.0cm 0cm]{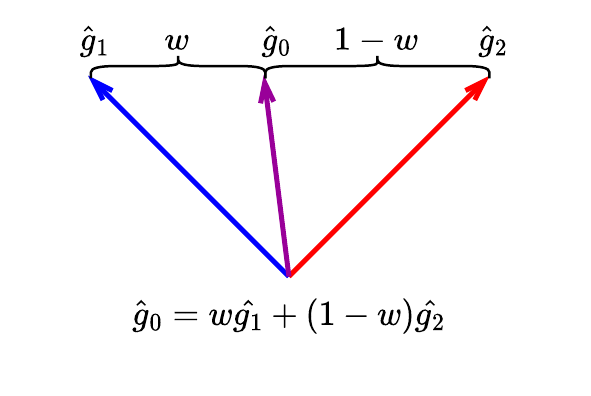}
    }
    \caption{Accumulated MTL gradient}
    \label{fig:geom_alg:third}
  \end{subfigure}    
  \caption{
  Geometric interpretation of our approach on a two-task MTL. Here, individual task gradients $g_1$ and $g_2$ are directed oppositely \textit{(conflict)} and have different magnitude \textit{(dominance)}~(\cref{fig:geom_alg:first}). 
  Aligned-MTL enforces stability via aligning principal components $u_1$, $u_2$ of an initial linear system of gradients. 
  This can be interpreted as re-scaling axes of a coordinate system set by principal components, so that singular values of gradient matrix $\sigma_1$ and $\sigma_2$ are rescaled to be equal to the minimal singular value ($\sigma_2$, in this case). The aligned gradients $\hat{g_1}$, $\hat{g_2}$ are orthogonal (non-conflicting) and of equal magnitude (non-dominant)~(\cref{fig:geom_alg:second}). Finally, the aligned gradients are summed up with pre-defined tasks weights $w$ and $1 - w$, resulting in a cumulative gradient $\hat{g_0}$~(\cref{fig:geom_alg:third}). 
  } \label{fig:geom_alg}
\end{figure*}


%% file: 04_analysis.tex
\subsection{Convergence Analysis}
In this section, we formulate a theorem regarding the convergence of our approach. Similar to single-task optimization converging to a stationary point, our MTL approach converges to a \textit{Pareto}-stationary solution. 

\begin{definition} \label{def:pareto}
A solution $\vtheta^* \in \bm{\Theta}$ is called \textbf{Pareto-stationary} iff there exists a convex combination of the gradient-vectors that is equal to zero. All possible Pareto-stationary solutions form a \textbf{Pareto set} (or \textbf{Pareto front}). 
\end{definition}

The overall model performance may vary significantly within points of the Pareto front. Recent MTL approaches~\cite{navon22a_nashmtl, liu2021_imtl} that provably converge to an arbitrary Pareto-stationary solution, tend to overfit to a subset of tasks. In contrast, our approach converges to a Pareto-stationary point with pre-defined tasks weights, thus providing more control over an optimization result~\cref{eq:opt_task}.

\begin{theorem}
\label{thm:convergence}
Assume $\calL_0(\vtheta), \dots, \calL_T(\vtheta)$ are lower-bounded continuously differentiable functions with Lipschitz continuous gradients with $\Lambda > 0$. A gradient descent with aligned gradient and step size $\alpha \leq \frac{1}{\Lambda}$ converges linearly to a Pareto stationary point where $\nabla \calL_0(\vtheta) = 0$.
\end{theorem}

A similar theorem is valid for aligning gradients in the shared representation space (Aligned-MTL upper-bound approximation is described in \cref{sec:fast_alignment}).
Mathematical proofs of both versions of this theorem versions are provided in supplementary materials. 

%% file: 05_results.tex
\section{Experiments}\label{sec:experiments}

\input{figs/city_metrics}

We empirically demonstrate the effectiveness of the proposed approach on various multi-task learning benchmarks, including scene understanding, multi-target regression, and reinforcement learning. 

\vspace{0.1cm}
\nbf{Competitors} We consider the following MTL approaches: {\em (1)}~\textbf{Linear Scalarization (LS, Uniform baseline)}: optimizing a uniformly weighted sum of individual task objectives, \ie $\frac{1}{T}\sum_{t}{\calL_t}$; 
{\em (2)}~\textbf{Dymanic Weight Average (DWA)} \cite{liu2019DWA}: adjusting task weights based on the rates of loss changes over time; {\em (3)}~\textbf{Uncertainty}~\cite{kendall2018} weighting; {\em (4)}~\textbf{MGDA}~\cite{desideri2012MGDA}: a multi-objective optimization with KKT~\cite{kuhn1951} conditions; {\em (5)}~\textbf{MGDA-UB}~\cite{sener2018MGDAUB}: optimizing an upper bound for the MGDA optimization objective; {\em (6)}~\textbf{GradNorm}~\cite{chen2018GradNorm}: normalizing the gradients to balance the learning of multiple tasks; {\em (7)}~\textbf{GradDrop}~\cite{chen2020graddrop}: forcing the sign consistency between task gradients; {\em (8)}~\textbf{PCGrad}~\cite{tianhe2020pcgrad}: performing gradient projection to avoid the negative interactions between tasks gradients; {\em (9)}~\textbf{GradVac}~\cite{wang2021gradvac}: leveraging task relatedness to set gradient similarity objectives and adaptively align task gradients, {\em (10)}~\textbf{CAGrad}~\cite{liu2021cagrad}: finding a conflict-averse gradients; {\em (11)}~\textbf{IMTL}~\cite{liu2021_imtl}: aligning projections to task gradients; {\em (12)}~\textbf{Nash-MTL}~\cite{navon22a_nashmtl}: utilizing a bargaining games for gradient computation, and {\em (13)}~\textbf{Random  loss weighting (RLW)}~\cite{lin2021RLW} with normal distribution. The proposed approach and the baseline methods are implemented using the PyTorch framework~\cite{Paszke19Pytorch}. The technical details on the training schedules and a complete listing of hyperparameters are provided in supplementary materials.

\vspace{0.1cm}
\nbf{Evaluation metrics} Besides task specific metrics we follow Maninis \etal~\cite{MRK19} and report a model performance drop relative to a single task baseline averaged over tasks: $\Delta m_{task} = \frac{1}{T} \sum_{t=1}^T \sum_{k=1}^{n_t} (-1)^{\sigma_{tk}} (M_{m, tk} - M_{b, tk}) / M_{b, tk}$ -- or over metrics: $\Delta m_{metric} = \frac{1}{T} \sum_{t=1}^{T} (-1)^{\sigma_{t}} (M_{m, t} - M_{b, t}) / M_{b, t}$.  Here, $M_{m, tk}$ denotes the performance of a model $m$ on a task $t$, measured with a metric $k$. Similarly, $M_{b, tk}$ is a performance of a single-task $t$ baseline; $n_t$ denotes number of metrics per task $t$. $\sigma_{tk} = 1$ if higher values of metric is better, and $\sigma_{tk} = 0$ otherwise. We mostly rely on the task-weighted measure since the metric-weighted criterion tends to be biased to a task with high number of metrics.

\subsection{Synthetic Example}\label{ssec:synthetic_mtl_problem}

To illustrate the proposed approach, we consider a synthetic MTL task~(\cref{fig:synthetic-description}) introduced in~\cite{liu2021cagrad} (a formal definition is provided in the supplementary material). As shown in~\cref{fig:teaser}, we perform optimization from five initial points tagged with $\bullet$. IMTL~\cite{liu2021_imtl}, and Nash-MTL~\cite{navon22a_nashmtl} aims at finding Pareto-stationary solution (\cref{def:pareto}). As a result, they terminate optimization once they reach a solution in the Pareto front. Accordingly, the final result strongly depends on an initialization point, and the optimization may not converge to the global optimum~$\bigstar$ in some cases~(\cref{fig:teaser:imtl} and \cref{fig:teaser:nashmtl}). Meanwhile, Aligned-MTL provides a less noisy and more stable trajectory, and provably converges to an optimum.

\input{tables/cityscapes_pspnet}

\input{tables/nyuv2_mtan}

\input{tables/nyuv2_pspnet}

\subsection{Scene Understanding}

The evaluation is performed on \textsc{NYUv2}~\cite{Silberman:ECCV12:nyuv2} and \textsc{CityScapes}~\cite{Cordts2016Cityscapes, Cordts2015Cvprw} datasets. 
We leverage two network architectures: Multi-Task Attention Network (MTAN)~\cite{liu2019DWA} and Pyramid Scene Parsing Network (PSPNet)~\cite{zhao2017, sener2018MGDAUB} on scene understanding benchmarks. MTAN applies a multi-task specific attention mechanism built upon MTL SegNet~\cite{kendall2018}. PSPNet features a dilated ResNet~\cite{he2016} backbone and multiple decoders with pyramid parsing modules~\cite{zhao2017}. Both networks were previously used in MTL benchmarks~\cite{sener2018MGDAUB}.

\indent \nbf{\scshape NYUv2} Following Liu~\etal~\cite{liu2019DWA, liu2021cagrad, navon22a_nashmtl}, we evaluate the performance of our approach on the NYUv2 \cite{Silberman:ECCV12:nyuv2} dataset by jointly solving semantic segmentation, depth estimation, and surface normal estimation tasks. We use both MTAN~\cite{liu2019DWA} and PSPNet~\cite{sener2018MGDAUB} model architectures.

For MTAN, we strictly follow the training procedure described in~\cite{navon22a_nashmtl, liu2021cagrad}: training at 384$\times$288 resolution for 200 epochs with Adam \cite{adam2015} optimizer and $10^{-4}$ initial learning rate, halved after 100 epochs. The evaluation results are presented in~\cref{tbl:mtan:nyuv2}. We report metric values averaged across three random initializations as in previous works. We calculate both metric-weighted measure to compare with previous works alongside a task-weighted $\Delta m$ modification. We claim the latter measure to be more important, as it is not biased towards surface normal estimation, thereby assessing overall performance more fairly. Accordingly, it exposes inconsistent task performance of GradNorm \cite{chen2018GradNorm} and MGDA \cite{sener2018MGDAUB}, which are biased towards surface normal estimation task and perform poorly on semantic segmentation. Although the MTAN model is not encoder-decoder architecture, our Aligned-MTL-UB approach outperforms all previous MTL optimization methods according to task-weighted $\Delta m$. Our original Aligned-MTL approach improves model performance even further in terms of both metrics.

We report results of PSPNet (\cref{tbl:pspnet:nyuv2}), trained on NYUv2~\cite{Silberman:ECCV12:nyuv2} following the same experimental setup. PSPNet architecture establishes much stronger baselines for all three tasks than vanilla SegNet. As a result, most of MTL approaches fail to outperform single-task models. According to the task-weighted metric, only two previous approaches provide solutions better than single-task baselines, while our Aligned-MTL approach demonstrates the best results.

\indent \nbf{{\scshape CityScapes}: two-task} We follow Liu \etal~\cite{liu2021cagrad} experimental setup for Cityscapes~\cite{Cordts2016Cityscapes}, which implies jointly addressing semantic segmentation and depth estimation with a single MTAN~\cite{liu2019DWA} model. According to it, the original 19 semantic segmentation categories are classified into 7 categories. Our Aligned-MTL approach demonstrates the best results according to semantic segmentation and overall $\Delta m$ metric. Our upper bound approximation of our Aligned-MTL again achieves a competitive performance, although MTAN does not satisfy architectural requirements.

\indent \nbf{{\scshape CityScapes}: three-task} We adopt a more challenging experimental setup~\cite{kendall2018, sener2018MGDAUB}, and address MTL with disparity estimation and instance and semantic segmentation tasks. The instance segmentation is reformulated as a centroid regression~\cite{kendall2018}, so that the instance objective has a much larger scale than others. In this benchmark, we utilize the training setup proposed by Sener and Koltun~\cite{sener2018MGDAUB}: 100 epochs, Adam optimizer with learning rate $10^{-4}$. Input images are rescaled to $256\times512$, and a full set of labels is used for semantic segmentation. While many recent approaches experience a considerable performance drop~(\cref{tbl:pspnet:city}), our method performs robustly even in this challenging scenario.

\subsection{Multi-task Reinforcement Learning}

Following~\cite{liu2021cagrad, navon22a_nashmtl, tianhe2020pcgrad}, we consider an MTL reinforcement learning benchmark MT10 in a MetaWorld~\cite{yu2019metaworld} environment. In this benchmark, a robot is being trained to perform actions, \eg pressing a button and opening a window. Each action is treated as a task, and the primary goal is to successfully perform a total of 10 diverse manipulation tasks. In this experiment, we compare against the optimization-based baseline Soft Actor-Critic (SAC)~\cite{Haarnoja2018SAC} trained with various gradient altering methods~\cite{liu2021cagrad, navon22a_nashmtl, tianhe2020pcgrad}. We also consider MTL-RL~\cite{pmlr-v139-sodhani21a-mtlrl}-based approaches: specifically, MTL SAC with a shared model, Multi-task SAC with task encoder (MTL SAC + TE)~\cite{yu2019metaworld}, Multi-headed SAC (MH SAC) with task-specific heads~\cite{yu2019metaworld}, Soft Modularization (SM)~\cite{yang20SM} and CARE~\cite{pmlr-v139-sodhani21a-mtlrl}. The Aligned-MTL method has higher success rates, superseding competitors by a notable margin.

\subsection{Empirical Analysis of Stability Criterion}\label{sec:empirical}

In this section, we analyze gradient magnitude similarity, cosine distance, and condition number empirically. We use \textsc{CityScapes} three-task benchmark for this purpose, which suffers from the dominating gradients. 
According to the gradient magnitude similarity measure, PCGrad~\cite{tianhe2020pcgrad}, Uniform, and CAGrad~\cite{liu2021cagrad} tend to suffer from gradient dominance. For PCGrad and Uniform baseline, imbalanced convergence rates for different tasks result in a suboptimal solution~(\cref{tbl:pspnet:city}). Differently, a well-performing CAGrad is misleadingly marked as problematic by gradient magnitude similarity. 
In contrast, the stability criterion -- \textit{condition number} -- reveals domination issues for PCGrad and Uniform baselines and indicates a sufficient balance of different tasks for CAGrad~($\kappa \approx 5$). Thus, the condition number exposes the training issues more evidently~(\cref{fig:city:three:cnumber}). The experimental evaluation shows that with $\kappa \leq 10$, model tends to converge to an optimum with better overall performance.

\input{tables/cityscapes_mtan}

\input{tables/rl}

%% file: figs/city_metrics.tex
\begin{figure*}[t!]
\centering

\begin{subfigure}[t]{1\textwidth}
\centering
    \includegraphics[width=1.0\textwidth]{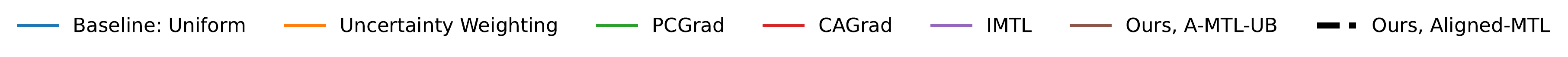}
\end{subfigure}

\begin{subfigure}[t]{.33\textwidth}
    \centering
    \includegraphics[width=1.0\textwidth]{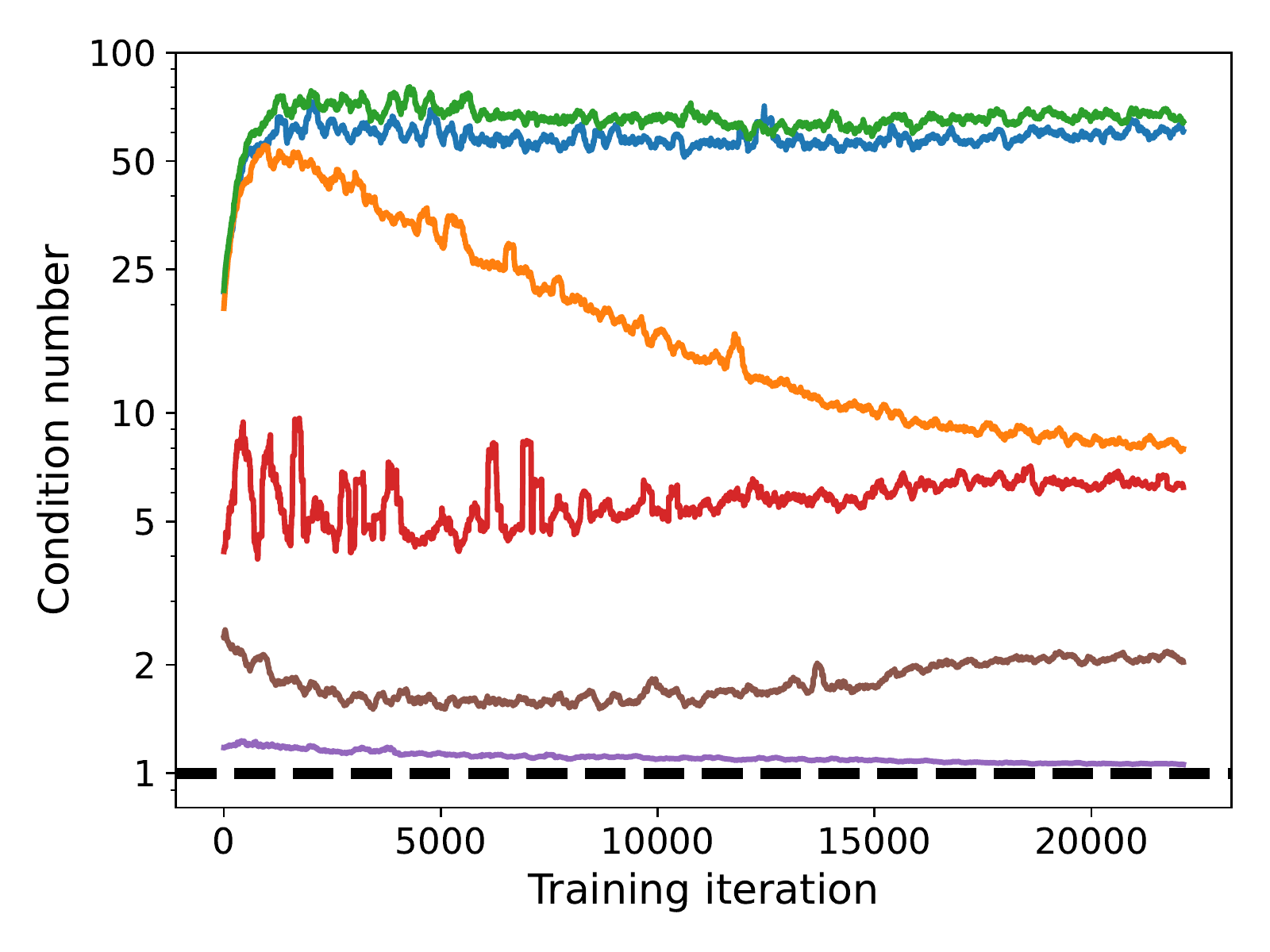}
    \caption{Condition Number}
    \label{fig:city:three:cnumber}
\end{subfigure}
\hfill
\begin{subfigure}[t]{.33\textwidth}
    \centering
    \includegraphics[width=1.0\textwidth]{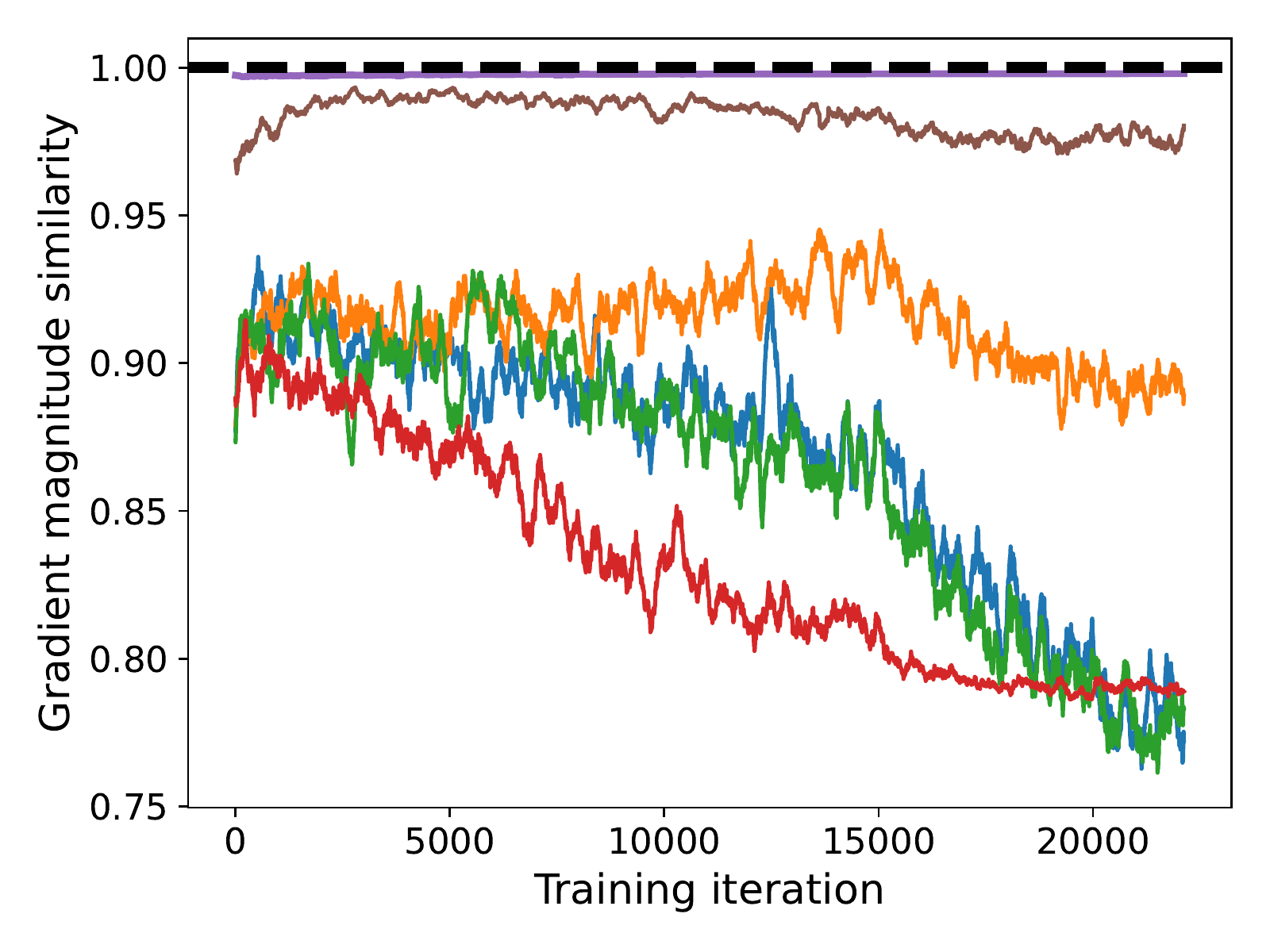}
    \caption{Gradient Magnitude Similarity~\cite{tianhe2020pcgrad}}
    \label{fig:city:three:gms}
\end{subfigure}
\hfill
\begin{subfigure}[t]{.33\textwidth}
\centering
    \includegraphics[width=1.0\textwidth]{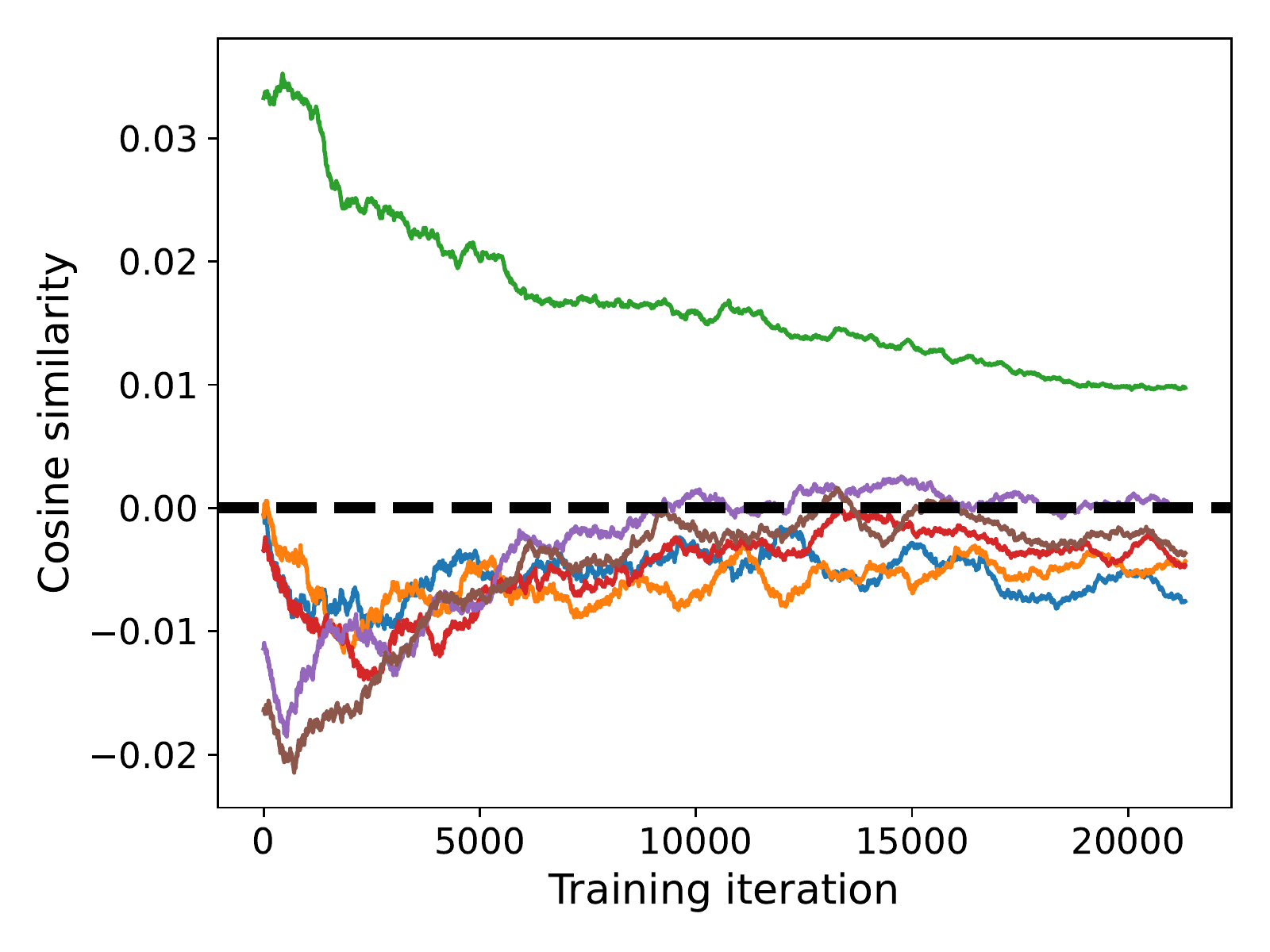}
    \caption{Gradient Conflicts}
    \label{fig:city:three:cosdist}
\end{subfigure}

\caption{Empirical evaluation of a stability criterion. We plot a condition number (\cref{eq:cnumber}), gradient magnitude similarity~\cite{tianhe2020pcgrad}, and minimal cosine distance during training on the \textsc{CityScapes} three-task benchmark. This benchmark suffers from high dominance since instance segmentation loss is of much larger scale than the others. The most intuitive way to define the dominance is the maximum ratio of task gradients magnitudes. The condition number coincides with this definition in a nearly orthogonal case, as in this benchmark~\cref{fig:city:three:cosdist}. However, gradient magnitude similarity measure~\cref{fig:city:three:gms} proposed in~\cite{tianhe2020pcgrad} does not reveal much correlation with a condition number(and, accordingly, with a maximal gradients magnitude ratio)~\cref{fig:city:three:cnumber}, so we assume it does not represent dominance issues comprehensively. From the empirical point of view~\cref{tbl:pspnet:city}, the value of a target metric is more correlated with the condition number, than with the gradient magnitude similarity. 
}
\label{fig:city:three:triad}

\end{figure*}

%% file: tables/cityscapes_pspnet.tex
\sisetup{round-mode = places, round-precision = 2, detect-weight=true}
\begin{table}[!tp]
\caption{\textbf{Scene understanding} ({\sc CityScapes}: three tasks). We report PSPNet~\cite{zhao2017, sener2018MGDAUB} model performance averaged over 3 random seeds. The best scores are provided in \protect\tikz[baseline=-.45ex]\protect\node[fill=mygray,inner sep=2pt,rounded corners=1pt]{gray};.}
    \centering
    \fontsize{7}{8}\selectfont
    \renewcommand{\tabcolsep}{2.4pt}
    \begin{tabular}{l|SSS|S}
    \toprule
    & {Segmentation~$\uparrow$} & {Instance~$\downarrow$} & {Disparity~$\downarrow$} &  {$\mathbf{\Delta m\%}$ $\downarrow$}\\
    {Method} &  {mIoU~[\%]} & {L1~[px]} & {MSE} &  \\
  \midrule
    Single task baselines & 66.73 & 10.55 & 0.33 & {--} \\ 
\midrule
Baseline:Uniform & 52.98 & 10.89 & 0.39 & 14.30 \\ 
RLW~\cite{lin2021RLW} & 51.26 & 10.25 & 0.41 & 15.58 \\ 
DWA~\cite{liu2019DWA} & 53.15 & 10.22 & 0.40 & 13.20 \\ 
Uncertainty~\cite{kendall2018} & 60.12 & \maxf{9.87} & 0.33 & 1.53 \\ 
MGDA~\cite{desideri2012MGDA} & 66.72 & 17.02 & 0.33 & 20.62 \\ 
MGDA-UB~\cite{sener2018MGDAUB} & 66.37 & 18.63 & 0.32 & 25.05\\ 
GradNorm~\cite{chen2018GradNorm} & 57.24 & 10.29 & 0.35 & 6.55\\ 
GradDrop~\cite{chen2020graddrop} & 52.98 & 10.09 & 0.40 & 12.50\\
PCGrad~\cite{tianhe2020pcgrad} & 54.06 & 9.91 & 0.38 & 10.00\\ 
GradVac~\cite{wang2021gradvac} & 54.07 & 10.39 & 0.40 & 12.99\\ 
CAGrad~\cite{liu2021cagrad} & 64.33 & 10.15 & 0.34 & 1.46 \\ 
IMTL~\cite{liu2021_imtl} & 65.13 & 11.58 & \maxf{0.32} & 3.10 \\ 
Nash-MTL~\cite{navon22a_nashmtl} & 64.84 & 11.90 & 0.37 & 9.38 \\ 
\midrule
Aligned-MTL (ours) & \maxf{67.06} & 10.63 & 0.33 & -0.02\\ 
Aligned-MTL-UB (ours) & 66.07 & 10.54 & 0.32 & \maxf{-0.35}  \\ 
  \bottomrule
    \end{tabular}
\label{tbl:pspnet:city}
\end{table}


%% file: tables/nyuv2_mtan.tex
\sisetup{round-mode = places, round-precision = 2, detect-weight=true}

\begin{table*}[t!] 
\caption{\textbf{Scene understanding} (\textsc{NYUv2}, three tasks). We report MTAN~\cite{liu2019DWA} model performance averaged over 3 random seeds. The best scores are provided in \protect\tikz[baseline=-.45ex]\protect\node[fill=mygray,inner sep=2pt,rounded corners=1pt]{gray};.
}
    \centering    \fontsize{7}{8}\selectfont
    \renewcommand{\tabcolsep}{3pt}
    \begin{tabular}{l|SSSSSSSSS|SS} \toprule
    & \multicolumn{2}{c}{Segmentation $\uparrow$} & \multicolumn{2}{c}{Depth~$\downarrow$} & \multicolumn{5}{c|}{Surface normals~$\downarrow$} & & \\
    & & & & & \multicolumn{2}{c}{Angle Dist.~$\downarrow$} & \multicolumn{3}{c|}{Within~$t^{\circ}$~$\uparrow$} & {$\mathbf{\Delta m\%}$ $\downarrow$} & {$\mathbf{\Delta m\%}$ $\downarrow$} \\
  {Method} & {mIoU} & {Pix Acc} & {Abs.} & {Rel.} & {Mean} & {Median} & {11.25} & {22.5} & {30} & {Metric-weighted} & {Task-weighted}\\
  \midrule
Single task baselines & 38.30 & 63.76 & 0.68 & 0.28 & 25.01 & 19.21 & 30.14 & 57.20 & 69.15 & {--} & {--} \\ 
\midrule
Baseline: Uniform & 39.29 & 65.33 & 0.5493 & 0.2263 & 28.15 & 23.96 & 22.09 & 47.50 & 61.08 & \color{purple}5.46 & \color{teal}-1.07 \\
RLW~\cite{lin2021RLW} & 37.17 & 63.77 & 0.58 & 0.24 & 28.27 & 24.18 & 22.26 & 47.05 & 60.62 & \color{purple}7.67 & \color{purple}2.00 \\ 
DWA~\cite{liu2019DWA} & 39.11 & 65.31 & 0.55 & 0.23 & 27.61 & 23.18 & 24.17 & 50.18 & 62.39 & \color{purple}3.49 & \color{teal}-2.06 \\ 
Uncertainty~\cite{kendall2018} & 36.87 & 63.17 & 0.54 & 0.23 & 27.04 & 22.61 & 23.54 & 49.05 & 63.65 & \color{purple}4.01 & \color{teal}-0.97 \\ 
MGDA~\cite{sener2018MGDAUB} & 30.47 & 59.90 & 0.61 & 0.26 & 24.88 & 19.45 & 29.18 & 56.88 & 69.36 & \color{purple}1.47 & \color{purple}1.79 \\ 
GradNorm~\cite{chen2018GradNorm} & 20.09 & 52.06 & 0.72 & 0.28 & \maxf{24.83} & \maxf{18.86} & \maxf{30.81} & \maxf{57.94} & \maxf{69.73} & \color{purple}7.22 & \color{purple}11.51 \\ 
GradDrop~\cite{chen2020graddrop} & 39.39 & 65.12 & 0.55 & 0.23 & 27.48 & 22.96 & 23.38 & 49.44 & 62.87 & \color{purple}3.61 & \color{teal}-2.03 \\ 
PCGrad~\cite{tianhe2020pcgrad} & 38.06 & 64.64 & 0.56 & 0.23 & 27.41 & 22.80 & 23.86 & 49.83 & 63.14 & \color{purple}3.83 & \color{teal}-1.33 \\ 
GradVac~\cite{wang2021gradvac} & 37.53 & 64.35 & 0.56 & 0.24 & 27.66 & 23.38 & 22.83 & 48.66 & 62.21 & \color{purple}5.44 & \color{purple}0.01 \\ 
CAGrad~\cite{liu2021cagrad} & 39.79 & 65.49 & 0.55 & 0.23 & 26.31 & 21.58 & 25.61 & 52.36 & 65.58 & \color{purple}0.29 & \color{teal}-4.18 \\ 
IMTL~\cite{liu2021_imtl} & 39.35 & 65.60 & 0.54 & 0.23 & 26.02 & 21.19 & 26.20 & 53.13 & 66.24 & \color{teal}-0.59 & \color{teal}-4.76 \\ 
Nash-MTL~\cite{navon22a_nashmtl} & 40.13 & 65.93 & 0.53 & 0.22 & 25.26 & 20.08 & 28.40 & 55.47 & 68.15 & \color{teal}-4.04 & \color{teal}-7.56 \\ 
\midrule
Aligned-MTL (ours) & 40.82 & 66.33 & \maxf{0.53} & \maxf{0.22} & 25.19 & 19.71 & 28.88 & 56.23 & 68.54 & \color{teal}\maxf{-4.93} & \color{teal}\maxf{-8.40} \\ 
Aligned-MTL-UB (ours) & \maxf{43.11} & \maxf{67.22} & 0.55 & 0.22 & 25.67 & 20.57 & 27.58 & 54.37 & 67.12 & \color{teal}-3.48 & \color{teal}-7.83 \\ 
  \bottomrule
    \end{tabular}
    \label{tbl:mtan:nyuv2}
\end{table*}


%% file: tables/nyuv2_pspnet.tex
\sisetup{round-mode = places, round-precision = 2, detect-weight=true}

\begin{table*}[t!] 
\caption{\textbf{Scene understanding} (\textsc{NYUv2}, three tasks). We report PSPNet~\cite{zhao2017, sener2018MGDAUB} model performance averaged over 3 random seeds. The best scores are provided in \protect\tikz[baseline=-.45ex]\protect\node[fill=mygray,inner sep=2pt,rounded corners=1pt]{gray};.  }
    \centering    \fontsize{7}{8}\selectfont
    \renewcommand{\tabcolsep}{3pt}
    \begin{tabular}{l|SSSSSSSSS|SS} \toprule
    & \multicolumn{2}{c}{Segmentation $\uparrow$} & \multicolumn{2}{c}{Depth~$\downarrow$} & \multicolumn{5}{c|}{Surface normals~$\downarrow$} & & \\
    & & & & & \multicolumn{2}{c}{Angle Dist.~$\downarrow$} & \multicolumn{3}{c|}{Within~$t^{\circ}$~$\uparrow$} & {$\mathbf{\Delta m\%}$ $\downarrow$} & {$\mathbf{\Delta m\%}$ $\downarrow$} \\
  {Method} & {mIoU} & {Pix Acc} & {Abs.} & {Rel.} & {Mean} & {Median} & {11.25} & {22.5} & {30} & {Metric-weighted} & {Task-weighted}\\
  \midrule
Single task baselines & 49.37 & 72.03 & 0.52 & 0.24 & 22.97 & 16.94 & 0.34 & 0.62 & 0.73 & {--} & {--} \\ 
\midrule
Baseline:Uniform & 45.21 & 69.70 & 0.49 & 0.21 & 26.10 & 21.08 & 0.26 & 0.52 & 0.66 & \color{purple}8.97 & \color{purple}4.72 \\ 
RLW~\cite{lin2021RLW} & 46.19 & 69.71 & \maxf{0.46} & \maxf{0.19} & 26.09 & 21.09 & 0.27 & 0.53 & 0.66 & \color{purple}6.67 & \color{purple}1.73 \\ 
DWA~\cite{liu2019DWA} & 45.83 & 69.65 & 0.50 & 0.22 & 26.10 & 21.27 & 0.26 & 0.52 & 0.66 & \color{purple}9.64 & \color{purple}5.61 \\ 
MGDA~\cite{sener2018MGDAUB} & 40.96 & 65.80 & 0.54 & 0.22 & \maxf{23.36} & \maxf{17.45} & \maxf{0.33} & \maxf{0.61} & \maxf{0.72} & \color{purple}3.54 & \color{purple}4.24 \\ 
MGDA-UB~\cite{sener2018MGDAUB} & 41.15 & 65.10 & 0.53 & 0.22 & 23.42 & 17.60 & 0.32 & 0.60 & \maxf{0.72} & \color{purple}4.02 & \color{purple}4.40 \\ 
GradNorm~\cite{chen2018GradNorm} & 45.63 & 69.64 & 0.48 & 0.20 & 25.46 & 20.06 & 0.28 & 0.55 & 0.67 & \color{purple}5.88 & \color{purple}2.18 \\ 
GradDrop~\cite{chen2020graddrop} & 45.69 & 70.13 & 0.49 & 0.20 & 26.16 & 21.21 & 0.26 & 0.52 & 0.65 & \color{purple}8.60 & \color{purple}3.92 \\ 
PCGrad~\cite{tianhe2020pcgrad} & 46.37 & 69.69 & 0.48 & 0.20 & 26.00 & 21.05 & 0.26 & 0.53 & 0.66 & \color{purple}7.78 & \color{purple}3.17 \\ 
GradVac~\cite{wang2021gradvac} & 46.65 & 69.97 & 0.49 & 0.21 & 25.95 & 20.88 & 0.27 & 0.53 & 0.66 & \color{purple}7.89 & \color{purple}3.75 \\ 
CAGrad~\cite{liu2021cagrad} & 45.46 & 69.35 & 0.47 & 0.20 & 24.28 & 18.73 & 0.30 & 0.58 & 0.70 & \color{purple}2.66 & \color{purple}0.13 \\ 
IMTL~\cite{liu2021_imtl} & 44.02 & 68.56 & 0.47 & \maxf{0.19} & 23.69 & 18.03 & 0.32 & 0.59 & \maxf{0.72} & \color{purple}\maxf{0.76} & \color{teal}-1.02 \\ 
Nash-MTL~\cite{navon22a_nashmtl} & \maxf{47.25} & \maxf{70.38} & \maxf{0.46} & 0.20 & 23.95 & 18.83 & 0.31 & 0.59 & 0.71 & \color{purple}1.13 & \color{teal}-1.48 \\ 
\midrule
Aligned-MTL (ours) & 46.70 & 69.97 & \maxf{0.46}& \maxf{0.19} & 24.19 & 18.77 & 0.30 & 0.58 & 0.71 & \color{purple}1.44 & \color{teal}\maxf{-1.55} \\ 
Aligned-MTL-UB (ours) & 46.47 & 69.92 & 0.48 & 0.20 & 24.37 & 18.88 & 0.30 & 0.58 & 0.70 & \color{purple}2.70 & \color{purple}0.07 \\ 
  \bottomrule
    \end{tabular}
    \label{tbl:pspnet:nyuv2}
\end{table*}



%% file: tables/cityscapes_mtan.tex
\sisetup{round-mode = places, round-precision=2, detect-weight=true}
\begin{table}[!tbp]
\caption{\textbf{Scene understanding} ({\sc CityScapes}: two tasks). MTAN~\cite{liu2019DWA} model performance is reported as average over 3 random seeds. 
The best scores are provided in \protect\tikz[baseline=-.45ex]\protect\node[fill=mygray,inner sep=2pt,rounded corners=1pt]{gray};.}
    \fontsize{7}{8}\selectfont
    \centering
    \renewcommand{\tabcolsep}{2.5pt}
    \begin{tabular}{l|SSS[round-precision=4]S|S}
    \toprule
    & \multicolumn{2}{c}{Segmentation} & \multicolumn{2}{c}{Depth} & {$\mathbf{\Delta_m\%}$ $\downarrow$} \\
     & {mIoU~[\%]~$\uparrow$} & {Pix. Acc~$\uparrow$} & {Abs Err~$\downarrow$} & {Rel Err~$\downarrow$} &  \\
     \midrule
Single task baselines & 74.01 & 93.16 & 0.0125 & 27.77 & {--} \\ 
\midrule
Baseline: Uniform & 75.18 & 93.49 & 0.0155 & 46.77 & 22.60 \\
RLW~\cite{lin2021RLW} & 74.57 & 93.41 & 0.0158 & 47.79 & 24.37 \\ 
DWA~\cite{liu2019DWA} & 75.24 & 93.52 & 0.0160 & 44.37 & 21.43 \\ 
Uncertainty~\cite{kendall2018} & 72.02 & 92.85 & 0.0140 & \maxf{30.13} & 5.88 \\ 
MGDA~\cite{sener2018MGDAUB} & 68.84 & 91.54 & 0.0309 & 33.50 & 44.14 \\ 
GradNorm~\cite{chen2018GradNorm} & 73.72 & 93.04 & \maxf{0.0124} & 34.11 & 5.63 \\ 
GradDrop~\cite{chen2020graddrop} & 75.27 & 93.53 & 0.0157 & 47.54 & 23.67 \\ 
PCGrad~\cite{tianhe2020pcgrad} & 75.13 & 93.48 & 0.0154 & 42.07 & 18.21 \\ 
CAGrad~\cite{liu2021cagrad} & 75.16 & 93.48 & 0.0141 & 37.60 & 11.58 \\ 
IMTL~\cite{liu2021_imtl} & 75.33 & 93.49 & 0.0135 & 38.41 & 11.04 \\ 
Nash-MTL~\cite{navon22a_nashmtl} & 75.41 & 93.66 & 0.0129 & 35.02 & 6.72 \\ 
\midrule
Aligned-MTL (ours) & \maxf{75.77} & \maxf{93.69} & 0.0133 & 32.66 & \maxf{5.27} \\ 
A-MTL-UB* (ours) & 74.89 & 93.46 & 0.0131 & 33.92 & 6.37 \\
\bottomrule
    \end{tabular}
    \label{tbl:mtan:cityscapes}
\end{table}

%% file: tables/rl.tex
\begin{table}[!tbp]
    \centering    \fontsize{8}{9}\selectfont
    \renewcommand{\tabcolsep}{3pt}
\caption{\textbf{Reinforcement learning} ({\sc MT10}). Average success rate on validation over 10 seeds. }
\label{tab:rl-mt10}
\begin{tabular}{l|c}
\toprule
                  & Success \textpm \ SEM     \\
\midrule
STL SAC           & 0.90 \textpm \ 0.032      \\
\midrule
MTL SAC           & 0.49 \textpm \ 0.073      \\
MTL SAC + TE      & 0.54 \textpm \ 0.047      \\
MH SAC            & 0.61 \textpm \ 0.036      \\
SM                & 0.73 \textpm \ 0.043      \\
CARE              & 0.84 \textpm \ 0.051      \\
PCGrad            & 0.72 \textpm \ 0.022      \\
CAGrad            & 0.83 \textpm \ 0.045      \\
Nash-MTL          & 0.91 \textpm \ 0.031      \\
\midrule
Ours, Aligned-MTL & \textbf{0.97 \textpm \ 0.045} \\
\bottomrule
\end{tabular}
\end{table}

%% file: 10_conclusion.tex
\section{Discussion}

The main limitation of Aligned-MTL is its computational optimization cost which scales linearly with the number of tasks. The upper-bound approximation of the Aligned-MTL method can be efficiently applied for encoder-decoder architectures using the same Jacobian over the shared representation. This approximation reduces instability, yet, it does not eliminate it since the Jacobian cannot be aligned. For non-encoder-decoder networks, upper-bound approximation has no theoretical guarantees but still can be leveraged as a heuristic and even provide a decent performance.

\section{Conclusion}
\label{sec:conclusion}

In this work, we introduced a stability criterion for multi-task learning, and proposed a novel gradient manipulation approach that optimizes this criterion. Our Aligned-MTL approach stabilize the training procedure by aligning the principal components of the gradient matrix. In contrast to many previous methods, this approach guarantees convergence to the local optimum with pre-defined task weights, providing a better control over the optimization results. Additionally, we presented a computationally efficient approximation of Aligned-MTL. Through extensive evaluation, we proved our approach consistently outperforms previous MTL optimization methods on various benchmarks including scene understanding and multi-task reinforcement learning.

%% file: 14_acknowledgements.tex
%

\vfill\hrule\medskip\noindent{\small\textbf{Acknowledgements.}
We sincerely thank Anna Vorontsova, Iaroslav Melekhov, Mikhail Romanov, Juho Kannala and Arno Solin for their helpful comments, disscussions and proposed improvements regarding this paper.
}

%% file: 12_appendix.tex
\appendix
\label{sec:appendix}
\input{appendix/01_convergence}

\input{appendix/02_cnumber}

\input{appendix/04_nyuv2_triad}
\input{appendix/03_synthetic}
\input{appendix/05_impl_details}

%% file: appendix/01_convergence.tex
\section{Convergence Analysis}
\paragraph{Synopsis.} In these theorems, we prove that the worst case performance of \textit{Aligned-MTL} and \textit{Aligned-MTL-U}B approaches is no worse than of standard gradient descent. The constraints mentioned in convergence theorems below are mild enough to be satisfied in practice. Our approach converges to a Pareto-stationary point with pre-defined tasks weights, thus providing more control over an optimization result.

\input{appendix/proofs/01_lemma}

\input{appendix/proofs/02_convergence_aligned-mtl}

\input{appendix/proofs/03_convergence_aligned-mtl-ub}

%% file: appendix/proofs/01_lemma.tex
\begin{lemma}
\label{lemma}
Assume $\calL(\vtheta)$ to be continuously differentiable and $\nabla \calL(\vtheta)$ to be Lipschitz continuous with $\Lambda > 0$. Then, the following restriction holds for a gradient descent with a step size $\alpha$ and an update rule $\vr$:
\begin{equation}
    \calL(\vtheta_t) - \calL(\vtheta_{t+1}) \geq \alpha  \lbr \nabla \calL(\vtheta_t), \vr \rbr - \frac{\alpha^2\Lambda}{2} \| \vr \|^2.
\end{equation}
\end{lemma}
\begin{proof-non}
Let us consider a gradient descent $\vtheta_{t+1} = \vtheta_t + \vdelta$, where  $\vdelta = -\alpha \vr$. From the fundamental theorem of calculus, we derive:
\begin{equation}
    \calL(\vtheta_t + \vdelta) - \calL(\vtheta_t) = \int_0^1 \lbr \nabla \calL (\vtheta_t + s \vdelta), \vdelta \rbr \,\mathrm{d}s.
\end{equation}

By adding and subtracting the value $\lbr \nabla \calL(\vtheta_t), \delta \rbr = \int_0^1 \lbr \nabla \calL(\vtheta_t), \delta \rbr \,\mathrm{d}s$, we obtain:
\begin{gather}
    \calL(\vtheta_{t+1}) - \calL(\vtheta_t) = \lbr \nabla \calL(\vtheta_t), \vdelta \rbr \ + \\ + \int_0^1 \lbr \nabla \calL (\vtheta_t + s \vdelta) - \nabla \calL(\vtheta_t), \vdelta \rbr \,\mathrm{d}s.
\end{gather}


Since the gradient satisfies the Lipschitz condition $\| \nabla \calL (\vtheta_t + s \vdelta) - \nabla \calL(\vtheta_t) \| \leq \Lambda \| \vtheta_t + s \vdelta - \vtheta_t \|$ and due to inequality $\lbr x, y \rbr \leq \|x\| \|y\|$, we can transform the integral as following:

\begin{gather*}
    \calL(\vtheta_{t+1}) - \calL(\vtheta_t) = \lbr \nabla \calL(\vtheta_t), \vdelta \rbr \ + \\ + \int_0^1 \lbr \nabla \calL (\vtheta_t + s \vdelta) - \nabla \calL(\vtheta_t), \vdelta \rbr \,\mathrm{d}s \leq \\
    \lbr \nabla \calL(\vtheta_t), \vdelta \rbr \ + \ \int_0^1 \Lambda \| \vtheta_t + s \vdelta - \vtheta_t \| \| \vdelta \| \mathrm{d}s \leq \\
    -\alpha \lbr \nabla \calL (\vtheta_t), \vr \rbr \ + \ \Lambda \int_0^1 \| - s \alpha \vr \|_2 \| - \alpha \vr \|_2 \mathrm{d}s \leq \\
    -\alpha \lbr \nabla \calL (\vtheta_t), \vr \rbr \ + \ \alpha^2\Lambda \|\vr\|^2 \int_0^1 s \mathrm{d}s \leq \\ 
    -\alpha \lbr \nabla \calL (\vtheta_t), \vr \rbr + \alpha^2\Lambda \|\vr\|^2
\end{gather*}

Therefore, we obtain the final constraint:
\begin{equation}
    \calL(\vtheta_{t+1}) - \calL(\vtheta_t) \leq -\alpha \lbr \nabla \calL (\vtheta_t), \vr \rbr + \alpha^2\Lambda \|\vr\|^2.
\end{equation}

\end{proof-non}

%% file: appendix/proofs/02_convergence_aligned-mtl.tex
\begin{theorem}[Aligned-MTL]
\label{thm:convergence_aligned_mtl}
Assume $\calL_0(\vtheta), \dots, \calL_T(\vtheta)$ are lower-bounded continuously differentiable functions with Lipschitz continuous gradients with $\Lambda > 0$. A gradient descent with an aligned gradient and a step size $\alpha \leq \frac{1}{\Lambda}$ converges linearly to a Pareto-stationary point where $\nabla \calL_0(\vtheta) = 0$.
\end{theorem}

\begin{proof-non}[\textbf{Aligned-MTL}]
Given the aforementioned assumptions, the cumulative objective satisfies \cref{lemma} with $\vr = \hat{\mG}\vw = \hat{g}_0$ and $\nabla \calL_0(\vtheta) = \mG\vw = \vg_0$:
\begin{equation}
    \label{thm1:main}
    \calL(\vtheta_t) - \calL(\vtheta_{t+1}) \geq \alpha \vg_0^\top \hat{\vg}_0 - \frac{\alpha^2\Lambda}{2} \| \hat{\vg_0} \|^2.
\end{equation}

According to SVD, $\mG = \mU \mSigma \mV^\top$, $\mSigma=\diag\{\sigma_1, \ldots, \sigma_R\}$ where $R = \rank \mG$, and $\mU^\top\mU = \mI$. By definition of the Aligned-MTL, we get:
\begin{gather*}
    \vg_0^\top \hat{\vg}_0 = \sigma_R \vw^\top\mV \mSigma \mU^\top \mU \mV^\top \vw = \\ = \sigma_R \vw^\top\mV \mSigma \mV^\top \vw = \sum_{r=1}^{R} \sigma_R \sigma_r (\vw^\top \vv_r)^2
\label{thm1:term2}
\end{gather*}
Similarly, $\| \hat{\vg}_0 \|^2 = \sum_{r=1}^{R} \sigma_R^2 (\vw^\top \vv_r)^2$. Since $\alpha \leq \frac{1}{\Lambda}$ and $\vw^\top \vv_r > \varepsilon$, \cref{thm1:main} can be further bounded:
\begin{gather*}
    \label{thm1:main_full}
    \calL(\vtheta_t) - \calL(\vtheta_{t+1}) \geq \sigma_R^2\frac{\alpha}{2} \underbrace{\sum_{r=1}^R \underbrace{\bigg(2\frac{\sigma_r}{\sigma_R} - 1\bigg)}_{>1} \bigg(\vw^\top\vv_r\bigg)^2}_{>\|\mV\vw\|^2 > \varepsilon^2} > \\ > \frac{\alpha \sigma_R^2}{2} \frac{\varepsilon^2}{\sigma_1^2} \sigma_1^2.
\end{gather*}

The dominance is always finite: $\frac{\sigma_R}{\sigma_1} > C$. Moreover, $\sigma_1 = \max_{\vx\neq0} \frac{\|\mG\vx\|}{\|\vx\|}$, therefore $\sigma_1 \geq \frac{\| \vg_0 \|}{\| \vw \|}$. Respectively:
\begin{equation}
    \label{thm1:main_final}
    \calL(\vtheta_t) - \calL(\vtheta_{t+1}) > \frac{\alpha \varepsilon^2 C^2}{2\|\vw\|^2} \| \vg_0 \|^2.
\end{equation}

The sequence of $\calL(\vtheta_t)$ is monotonically decreasing and bounded (under assumption), and hence converging. Then $\calL(\vtheta_t) - \calL(\vtheta_{t+1}) \rightarrow 0$ if $t \rightarrow \infty$. Thereby, we have a local convergence of the gradient descent:
\begin{equation}
    \|\vg_0\|^2  < \frac{2 \|\vw\|^2}{\alpha C^2 \epsilon^2} \bigg( \calL(\vtheta_t) - \calL(\vtheta_{t+1}) \bigg) \rightarrow 0\quad \text{as}\quad t \rightarrow \infty.
\end{equation}

The same estimate appears in case of the gradient descent. Accordingly, the convergence of Aligned-MTL is similar to that of the gradient descent, \ie \textit{linear} -- $\mathcal{O}(\frac{1}{T})$.
\end{proof-non}

%% file: appendix/proofs/03_convergence_aligned-mtl-ub.tex
\begin{theorem}[A-MTL-UB]
\label{thm:convergence_aligned_mtl_ub}
Assume $\calL_0(\vtheta), \dots, \calL_T(\vtheta)$ are lower-bounded continuously differentiable functions with Lipschitz continuous gradients with $\Lambda > 0$. Suppose $\mJ = \frac{\partial \mH}{\partial \vtheta}$ to be a full rank, \ie $\rank \mJ = \min\{|\vtheta|, |\mH|\}$. A gradient descent with an aligned gradient and a step size $\alpha \leq \frac{1}{\Lambda}$ converges linearly to a Pareto-stationary point where $\nabla \calL_0(\vtheta) = 0$.
\end{theorem}

\begin{proof-non}[Aligned-MTL-UB]
Similarly to the \cref{thm:convergence_aligned_mtl}, under the aforementioned assumptions, the cumulative objective satisfies \cref{lemma} with $\vr = \sigma_R\mJ\hat{\mZ}\vw = \hat{\vg}_0$ and $\nabla \calL_0(\vtheta) = \mJ \mZ \vw = \vg_0$:
\begin{equation}
    \label{thm2:main}
    \calL(\vtheta_t) - \calL(\vtheta_{t+1}) \geq \alpha \vg_0^\top \hat{\vg}_0 - \frac{\alpha^2\Lambda}{2} \| \hat{\vg_0} \|^2.
\end{equation}

According to SVD, $\mZ = \mU \mSigma \mV^\top$, $\mSigma=\diag\{\sigma_1, \ldots, \sigma_R\}$ where $R = \rank \mZ$, and $\mU^\top\mU = \mI$. By definition of the Aligned-MTL-UB, we get:
\begin{gather*}
    \vg_0^\top \hat{\vg}_0 = \sigma_R \vw^\top\mV \mSigma \mU^\top\mJ^\top\mJ \mU \mV^\top \vw \\
    \hat{\vg}_0^\top \hat{\vg}_0 = \sigma_R^2 \vw^\top\mV \mU^\top\mJ^\top\mJ \mU \mV^\top \vw
\label{thm2:term2}
\end{gather*}

Since $\mJ$ is full rank, $\mJ^\top\mJ$ is positive definite. Any positive definite matrix is congruent to a diagonal~($\mD$) with positive and ordered eigenvalues on the main diagonal. Thus, replacing all eigenvalues $\lambda_i^2$ with the smallest one $\lambda_K^2$ does not increase the inner product produced by this matrix: $\vx\mD\vx \geq \lambda_K \vx^\top\vx$. By taking this into consideration, we can bound the right side of \cref{thm2:main}:
\begin{gather*}
    \calL(\vtheta_t) - \calL(\vtheta_{t+1}) \geq \frac{\alpha}{2}(2\vg_0 - \hat{\vg}_0)^\top\hat{\vg}_0 \geq \\
    \frac{\alpha}{2} (2\sigma_R \vw^\top \mV\mSigma\mU^\top - \sigma_R^2 \vw^\top \mV\mU^\top) \mJ^\top\mJ \mU\mV^\top\vw \geq \\
    \sigma_R^2 \lambda_K^2 \underbrace{\sum_{r=1}^R \underbrace{\bigg(2\frac{\sigma_r}{\sigma_R} - 1\bigg)}_{>1} \bigg(\vw^\top\vv_r\bigg)^2}_{>\|\mV\vw\|^2>\varepsilon^2}
\label{thm2:term3}
\end{gather*}
Thus:
\begin{equation}
    \calL(\vtheta_t) - \calL(\vtheta_{t+1}) \geq \frac{\alpha \varepsilon^2\sigma_R^2 \lambda_K^2}{2\sigma_1^2 \lambda_1^2} \sigma_1^2 \lambda_1^2.
\end{equation}

Following the assumption, $\frac{\sigma_R}{\sigma_1} > C_{\sigma}$ and $\frac{\lambda_K}{\lambda_1} > C_{\lambda}$. Moreover, $\sigma_1 = \max_{\vx\neq0} \frac{\|\mZ\vx\|}{\|\vx\|} \geq \frac{\| \mZ\vw \|}{\| \vw \|}$ and $\lambda_1 = \| \mJ \|$. Therefore, we obtain the final bound:
\begin{gather*}
    \calL(\vtheta_t) - \calL(\vtheta_{t+1}) \geq \frac{\alpha \varepsilon^2 C_{\sigma}^2 C_{\lambda}^2}{2\|\vw\|^2} \| \mG_{\mZ}\vw \|^2 \| \mJ \|^2 \geq \\ \frac{\alpha \varepsilon^2 C_{\sigma}^2 C_{\lambda}^2}{2\|\vw\|^2} \| \vg_0 \|^2.
\end{gather*}

The sequence of $\calL(\vtheta_t)$ is monotonically decreasing and bounded (under assumption), and hence converging. Then $\calL(\vtheta_t) - \calL(\vtheta_{t+1}) \rightarrow 0$ if $t \rightarrow \infty$. Thereby, we have a local convergence of the gradient descent:
\begin{equation}
    \|\vg_0\|^2  < \frac{2 \|\vw\|^2}{\alpha C_{\sigma}^2 C_{\lambda}^2 \varepsilon^2} \bigg( \calL(\vtheta_t) - \calL(\vtheta_{t+1}) \bigg) \to 0 \quad \text{as} \quad t \to \infty.
\end{equation}
\end{proof-non}

%% file: appendix/02_cnumber.tex
\section{Condition Number}

The stability criterion is closely related to the dominance and conflicts. We can find a functional dependence between them for some special cases: \textbf{a)} gradients $\vg_1$ and $\vg_2$ have equal magnitude but not orthogonal, \textbf{b)} they are othogonal but have different norms. To this end, we formulate the following colloraries.
\begin{collorary} Given $\vg_1 \perp \vg_2$ condition number $\kappa$ is 
$$\kappa= \max \bigg\{ \frac{\| \vg_1 \|} {\| \vg_2 \|}, \frac{\| \vg_2 \|} {\| \vg_1 \|} \bigg\}$$
\end{collorary}
\begin{proof-non}
By initial assumtions the Gram matrix $\mG^\top\mG$ is diagonal: 
$$\mG^\top\mG = \diag\{\| \vg_1 \|^2, \| \vg_2 \|^2\}$$
At the same time, this matrix can be factorized using eigen decomposition:
$$\mG^\top\mG = \mV\mSigma^2\mV^\top, \quad \mV\mV^\top = \mI, \quad \mSigma = \diag\{\sigma_1, \sigma_2\}$$
Thus, the singular values are proportional to the gradient magnitudes up to a symmetric swap to keep ordering of singular values. The coefficient of proportionality is not valuable, since the condition number is invariant to the global scale. Therefore, we derive:
$$\kappa= \max \bigg\{ \frac{\| \vg_1 \|} {\| \vg_2 \|}, \frac{\| \vg_2 \|} {\| \vg_1 \|} \bigg\}$$
\end{proof-non}

\input{appendix/figs/cnumber_cos}

\begin{collorary} Given $\vg_1$ and  $\vg_2$ with equal magnitudes, \ie $\| \vg_1 \| = \| \vg_2 \|$, and with $\alpha$ angle in between the condition number  $\kappa$ is 
\begin{equation}
\kappa = 
    \begin{cases}
    \tan(\alpha / 2) & \frac{\pi}{4} < \alpha / 2 \leq \frac{\pi}{2} \\
    \ctan(\alpha / 2) & 0 < \alpha / 2 < \frac{\pi}{4} \\
    \end{cases}
\end{equation}
\end{collorary}
\begin{proof-non}
The direct collorary of SVD states, that the principal components $\vu_i$ are direction with maximum norm of projections over all gradients. Formally:
\begin{gather*}
    \sigma_1 = \max_{\|\vx\|=1} \| \mG^\top \vx \| = \| \mG^\top \vu_1 \| \\
    \sigma_2 = \max_{\|\vx\|=1, \vx \perp \vu_1} \| \mG^\top \vx \| = \| \mG^\top \vu_2 \|
\end{gather*}
Since the gradients have the same length, one of the principal components is the bisectrix of angle between them. For clarity, we suppose, that the bisectrix is the second component. Then, the singular values can be computed trivially~(\cref{fig:cnumber_cos}):
\begin{gather*}
    \sigma_1 = \sqrt{2} \sin(\alpha / 2) \| \vg_1 \| \\
    \sigma_2 = \sqrt{2} \cos(\alpha / 2) \| \vg_1 \|
\end{gather*}
Accroding to these expressions the condition number is tangent or cotangent up to a symmetric swap to keep ordering of singular values. In orthoginal case, the condition number is unit.
\end{proof-non}

%% file: appendix/figs/cnumber_cos.tex
\begin{figure}[t]
    \centering
    \includegraphics[scale = 0.8]{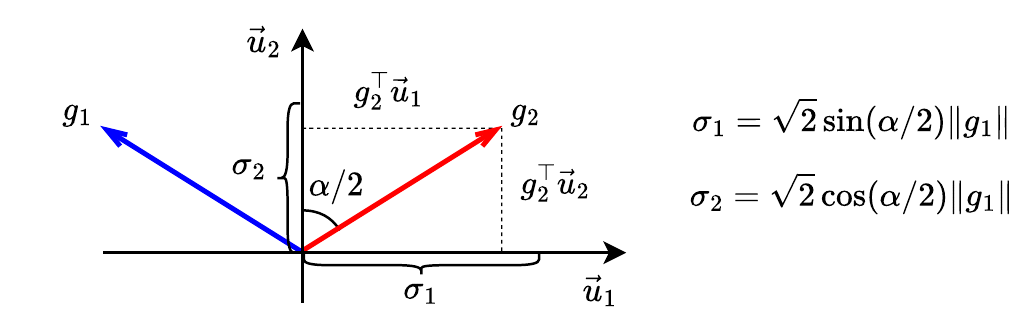}
    \caption{The condition number depends on the angle between gradient vectors. Due to the symmetry one of the principal components is a bisectrix of this angle.}
    \label{fig:cnumber_cos}
\end{figure}

%% file: appendix/03_synthetic.tex
\section{Synthetic Example}

\begin{figure*}
\centering
\caption{Comparison of MTL optimization methods on synthetic two-task benchmark \cite{liu2021cagrad, navon22a_nashmtl}. We explore convergence of various methods with varying pre-defined task weights. Methods that guarantee only Pareto-front convergence (such as IMTL\cite{liu2021_imtl} and NashMTL \cite{navon22a_nashmtl}) fail to achieve global optimum  (defined by $\bigstar$) and converge to an arbitrary Pareto-front solution with unknown task balance. Unlike previous methods, our Aligned-MTL approach respects pre-defined task weights and converges to the global optimum for all task weights combinations and initialization points ($\bullet$), except one extreme case. Moreover, our method provides stable and less noisy trajectories than other methods.}
\label{fig: synthetic-varying}
\setkeys{Gin}{width=\linewidth}
\setlength\tabcolsep{5pt}
\begin{tabularx}{\textwidth}{>{\centering\arraybackslash} m{0.2cm} >{\centering\arraybackslash} m{3.1cm}>{\centering\arraybackslash} m{3.1cm}>{\centering\arraybackslash} m{3.1cm}>{\centering\arraybackslash} m{3.1cm}>{\centering\arraybackslash} m{3.1cm}}
\toprule
& $0.1\mathcal{L}_1 + 0.9\mathcal{L}_2$ & $0.3\mathcal{L}_1 + 0.7\mathcal{L}_2$ & $0.5\mathcal{L}_1 + 0.5\mathcal{L}_2$ & $0.7\mathcal{L}_1 + 0.3\mathcal{L}_2$ & $0.9\mathcal{L}_1 + 0.1\mathcal{L}_2$\\
\midrule
\rotatebox{90}{Uniform} & \includegraphics[clip, trim=1.0cm 0.5cm 1.5cm 1.2cm]{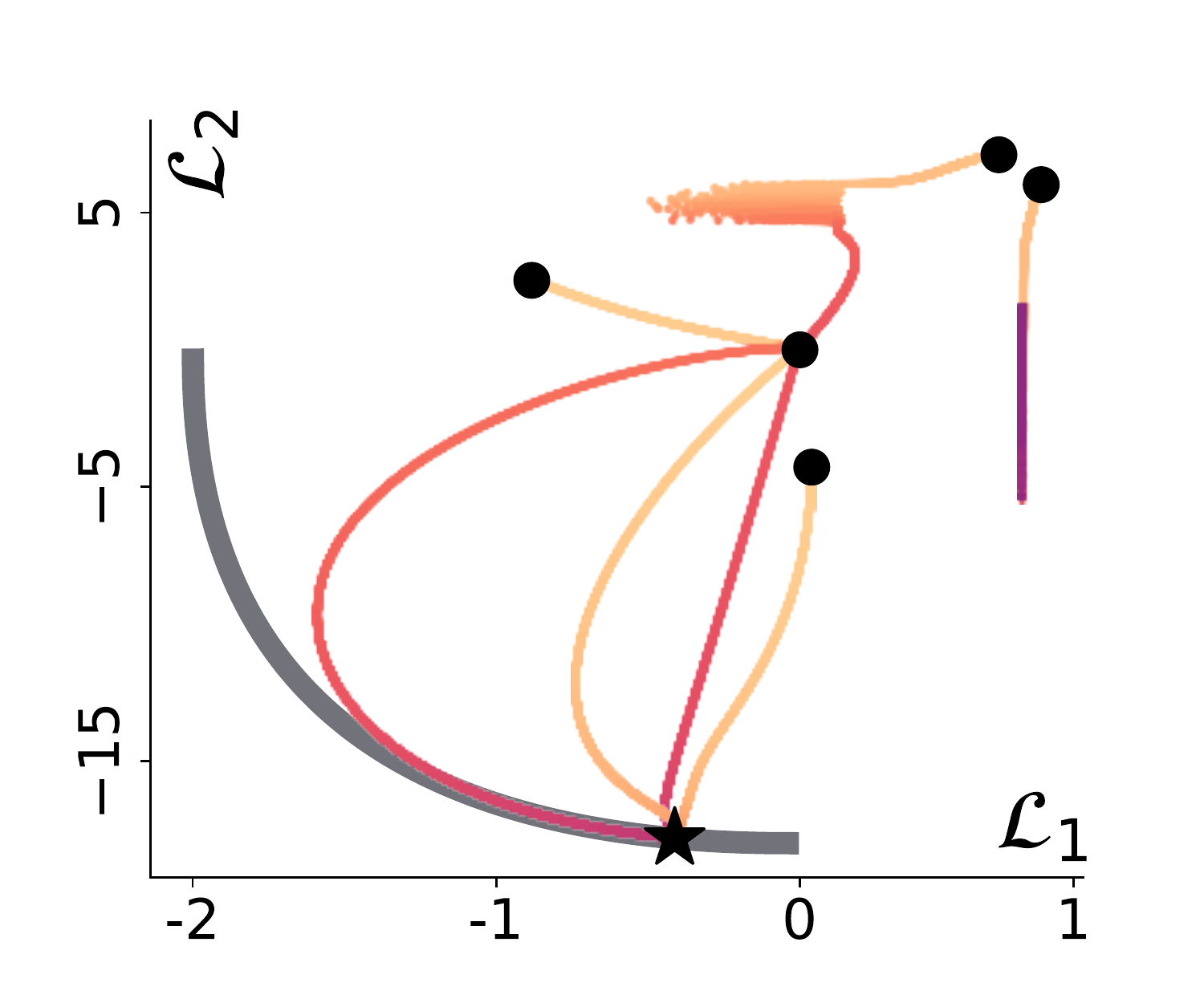} & \includegraphics[clip, trim=1.0cm 0.5cm 1.5cm 1.2cm]{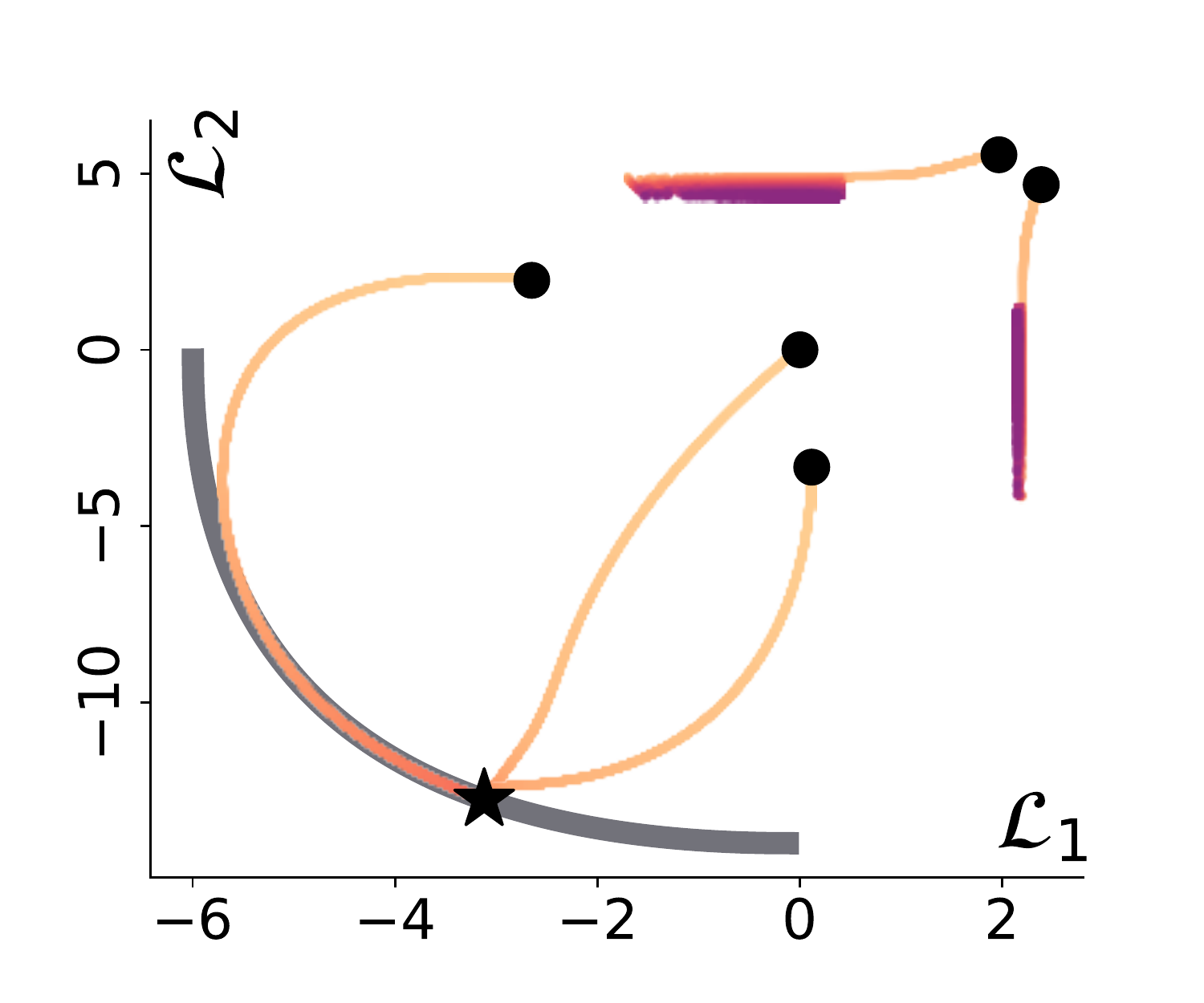} & \includegraphics[clip, trim=1.0cm 0.5cm 1.5cm 1.2cm]{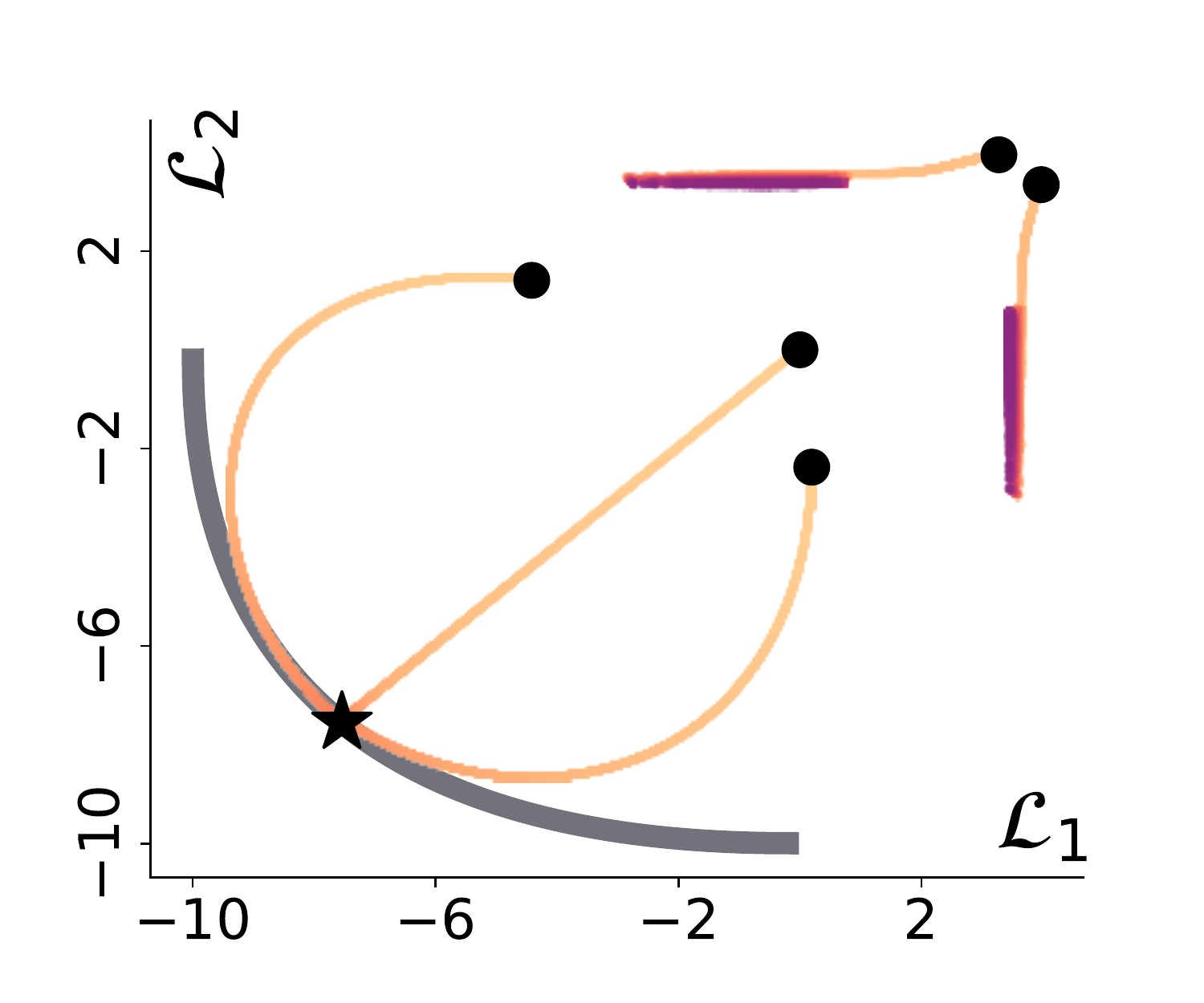} & \includegraphics[clip, trim=1.0cm 0.5cm 1.5cm 1.2cm]{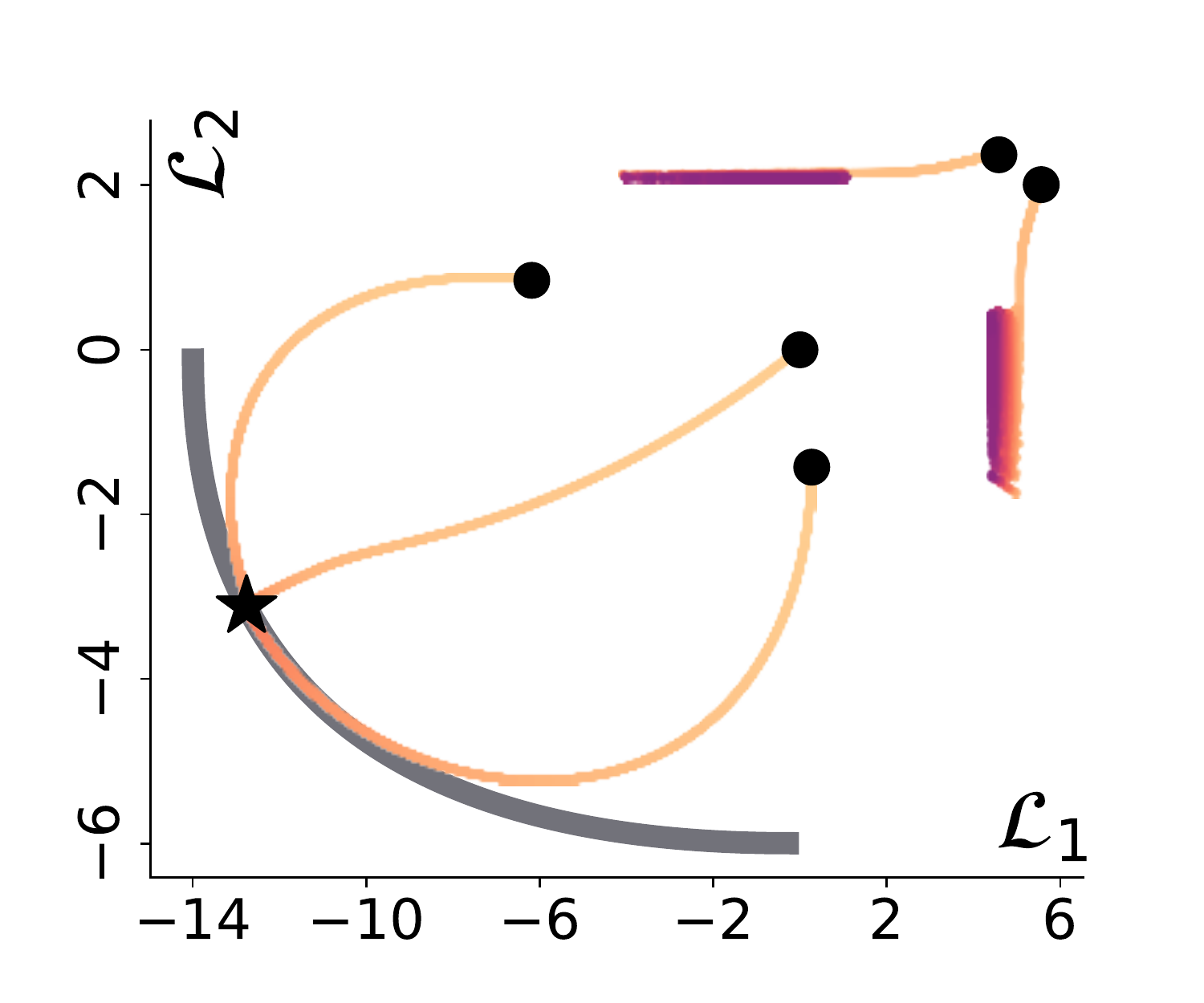} & \includegraphics[clip, trim=1.0cm 0.5cm 1.5cm 1.2cm]{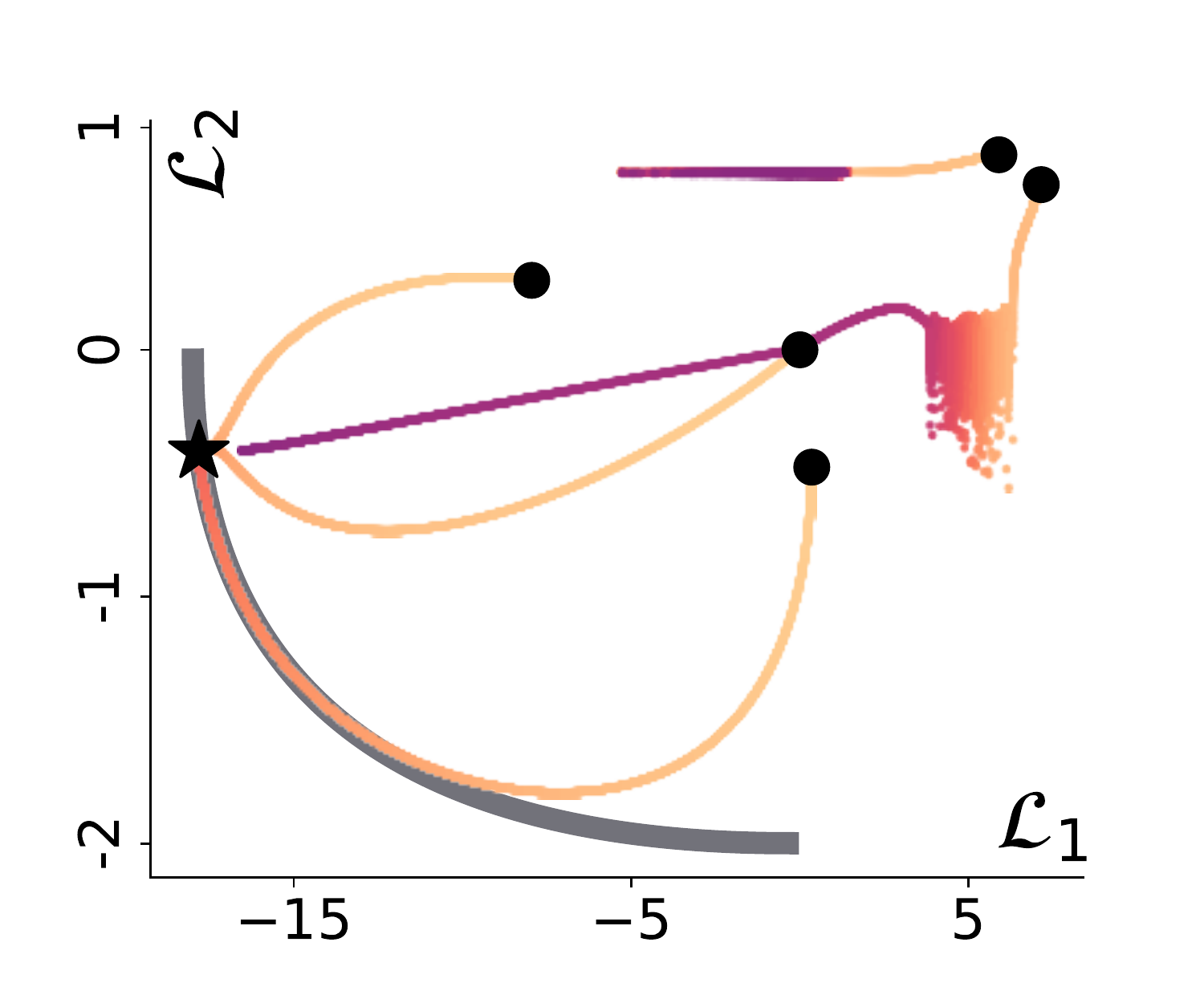} \\

\rotatebox{90}{PCGrad\cite{tianhe2020pcgrad}} & \includegraphics[clip, trim=1.0cm 0.5cm 1.5cm 1.2cm]{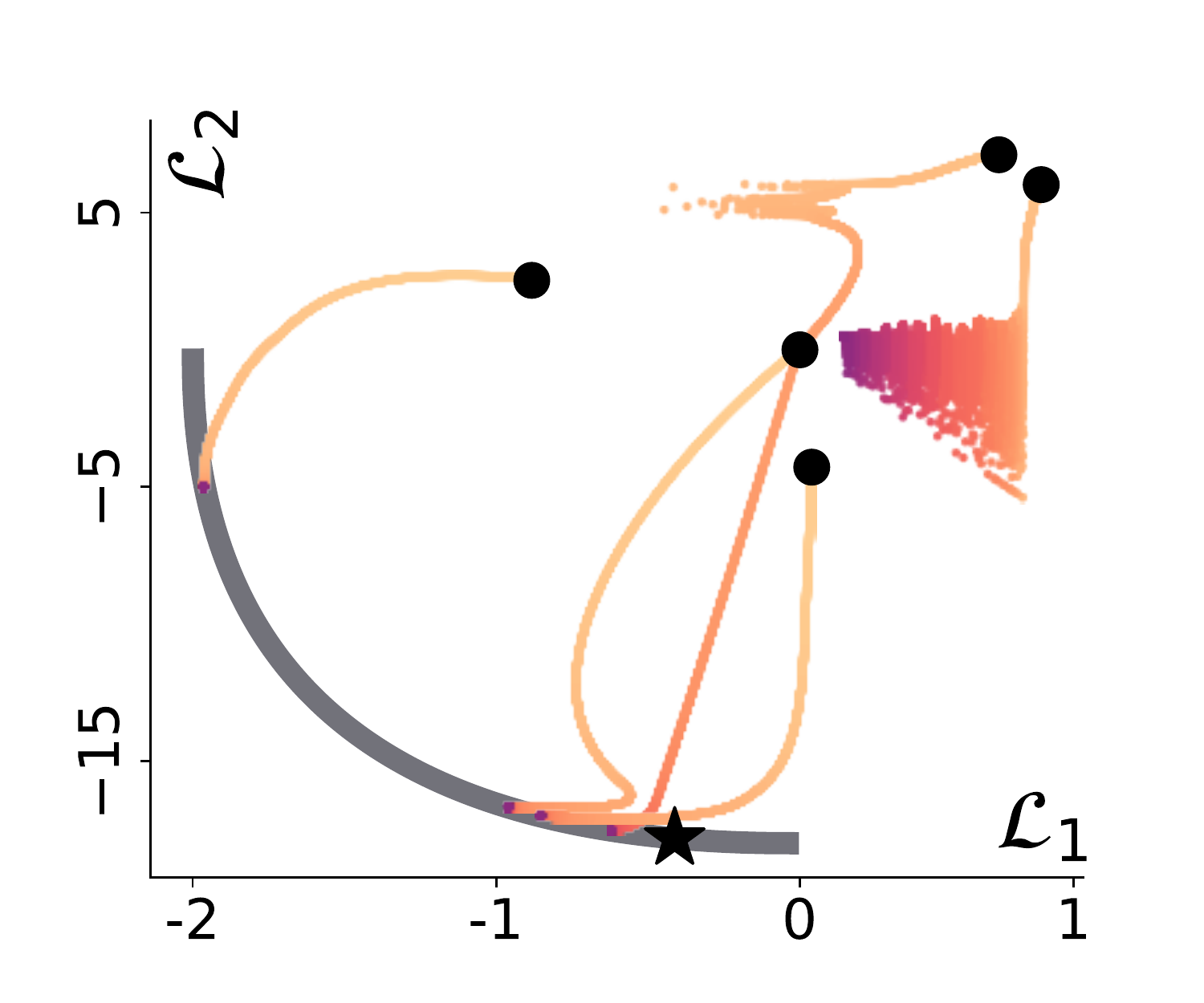} & \includegraphics[clip, trim=1.0cm 0.5cm 1.5cm 1.2cm]{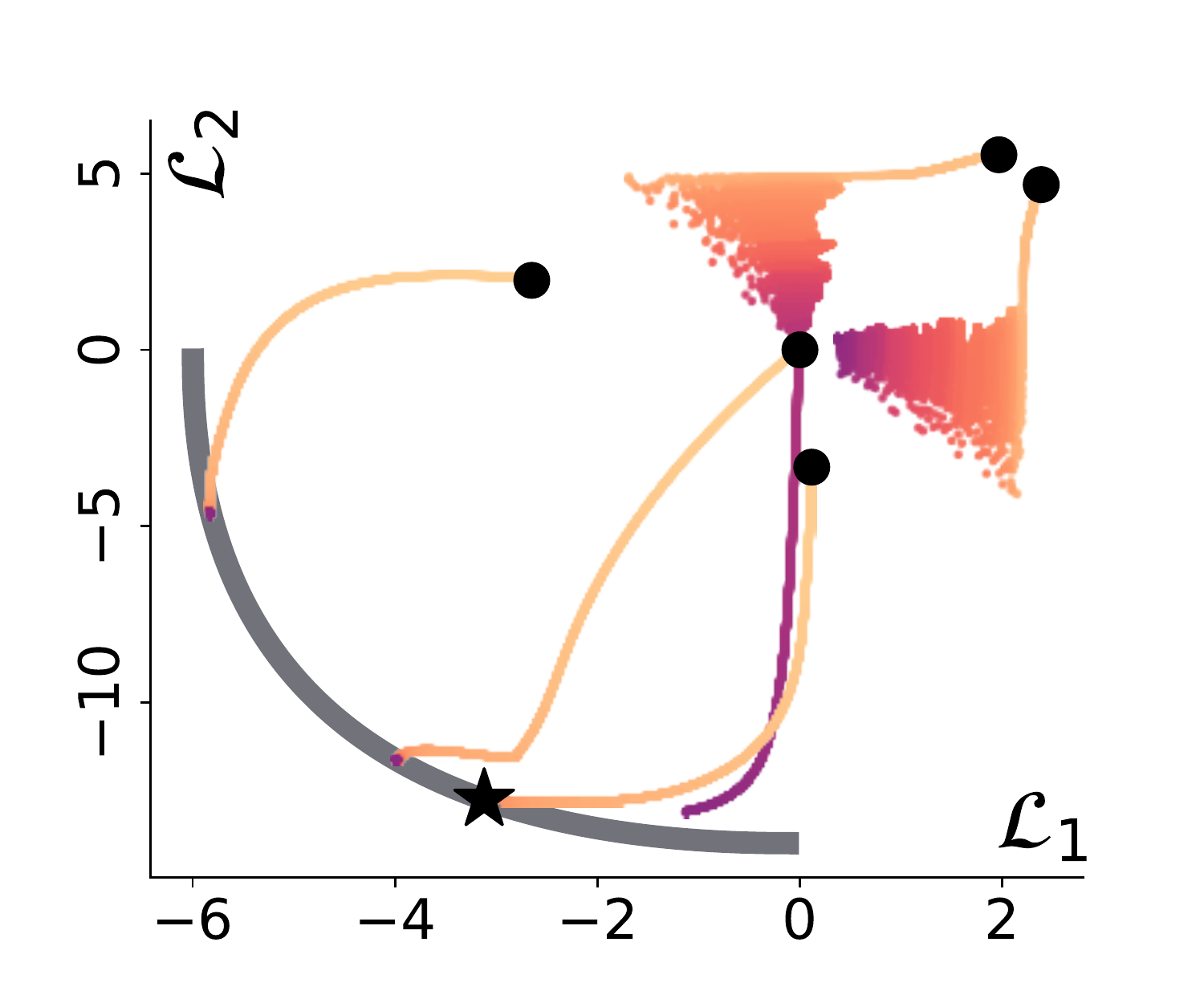} & \includegraphics[clip, trim=1.0cm 0.5cm 1.5cm 1.2cm]{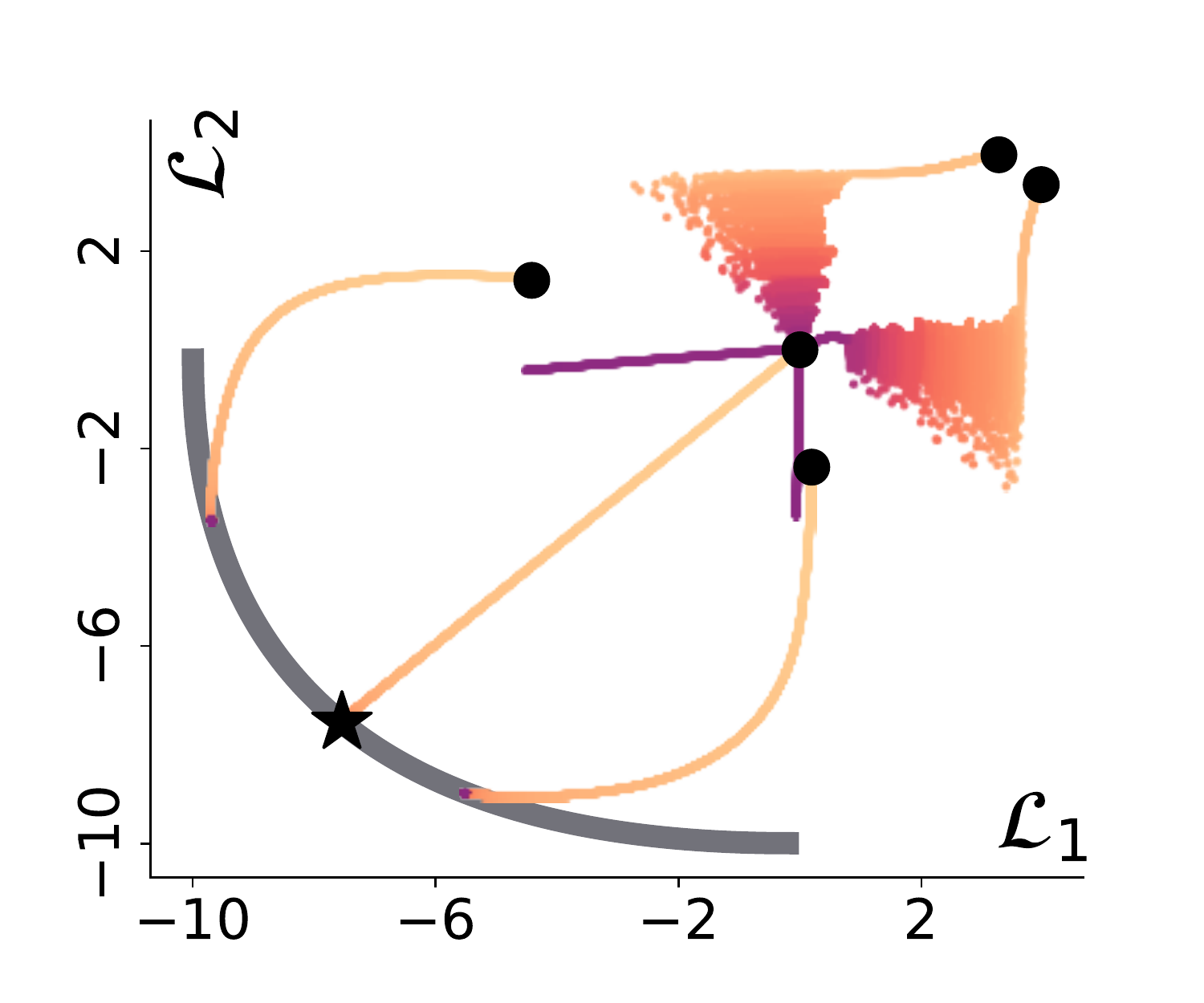} & \includegraphics[clip, trim=1.0cm 0.5cm 1.5cm 1.2cm]{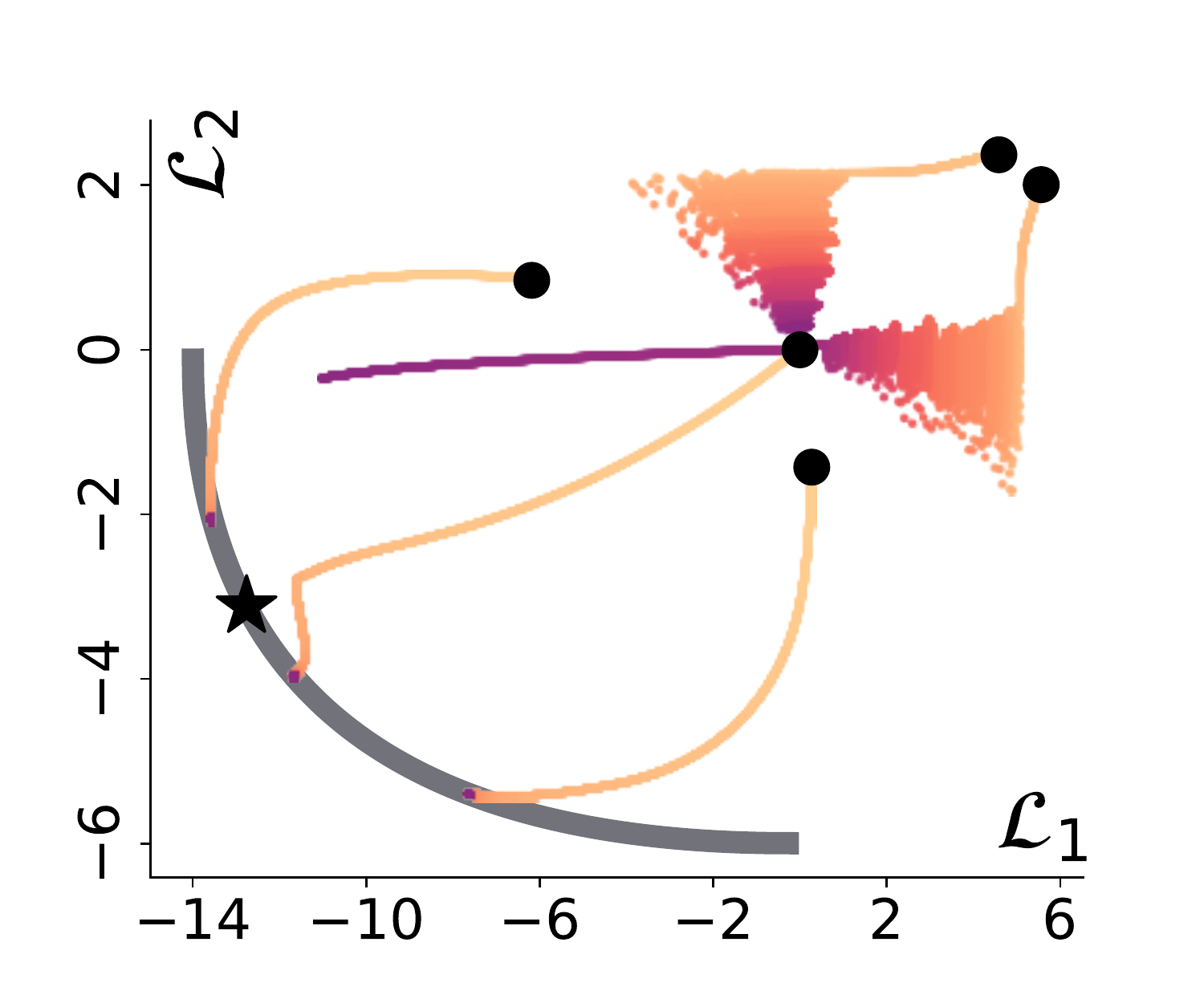} & \includegraphics[clip, trim=1.0cm 0.5cm 1.5cm 1.2cm]{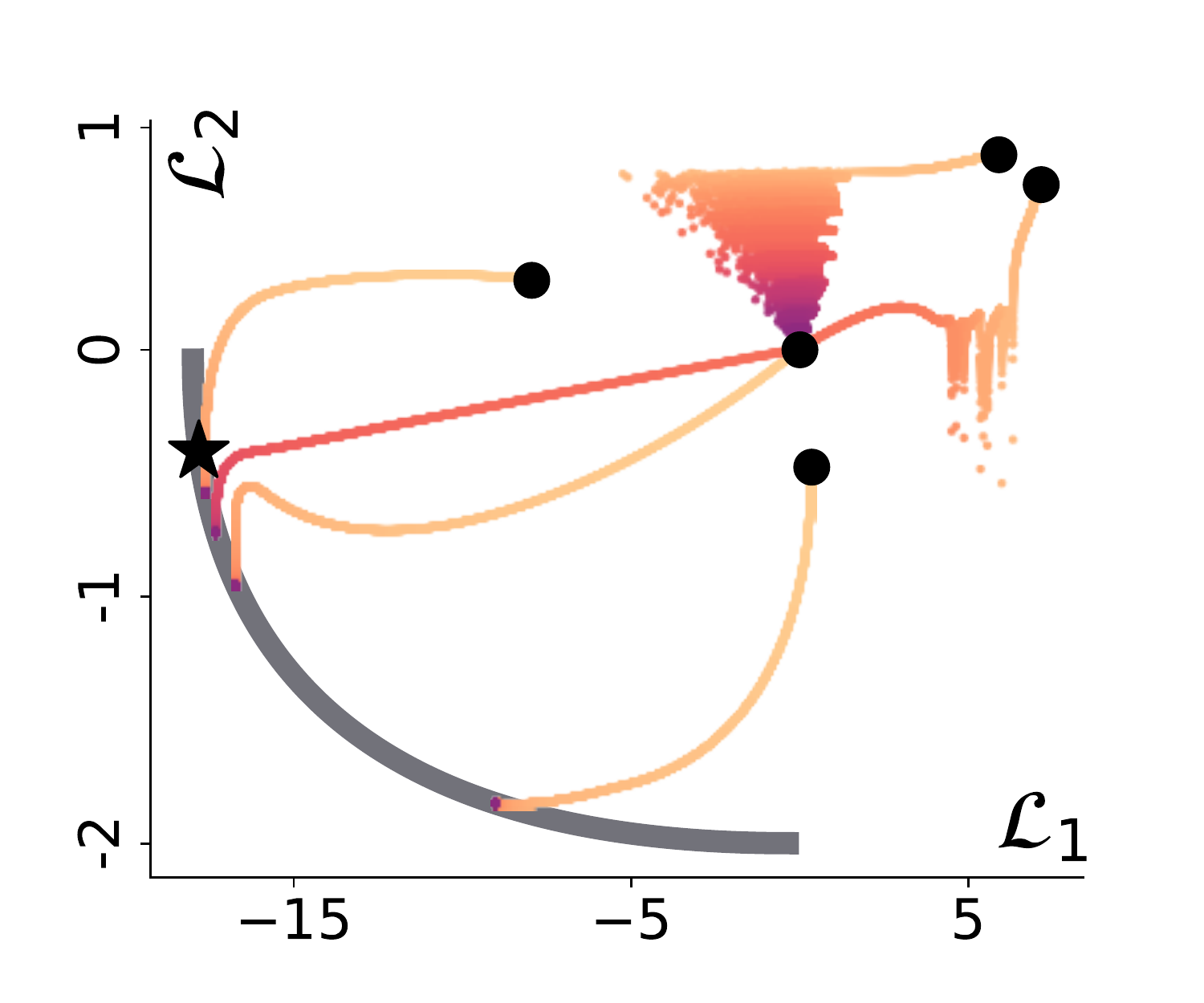} \\

\rotatebox{90}{CAGrad\cite{liu2021cagrad}} & \includegraphics[clip, trim=1.0cm 0.5cm 1.5cm 1.2cm]{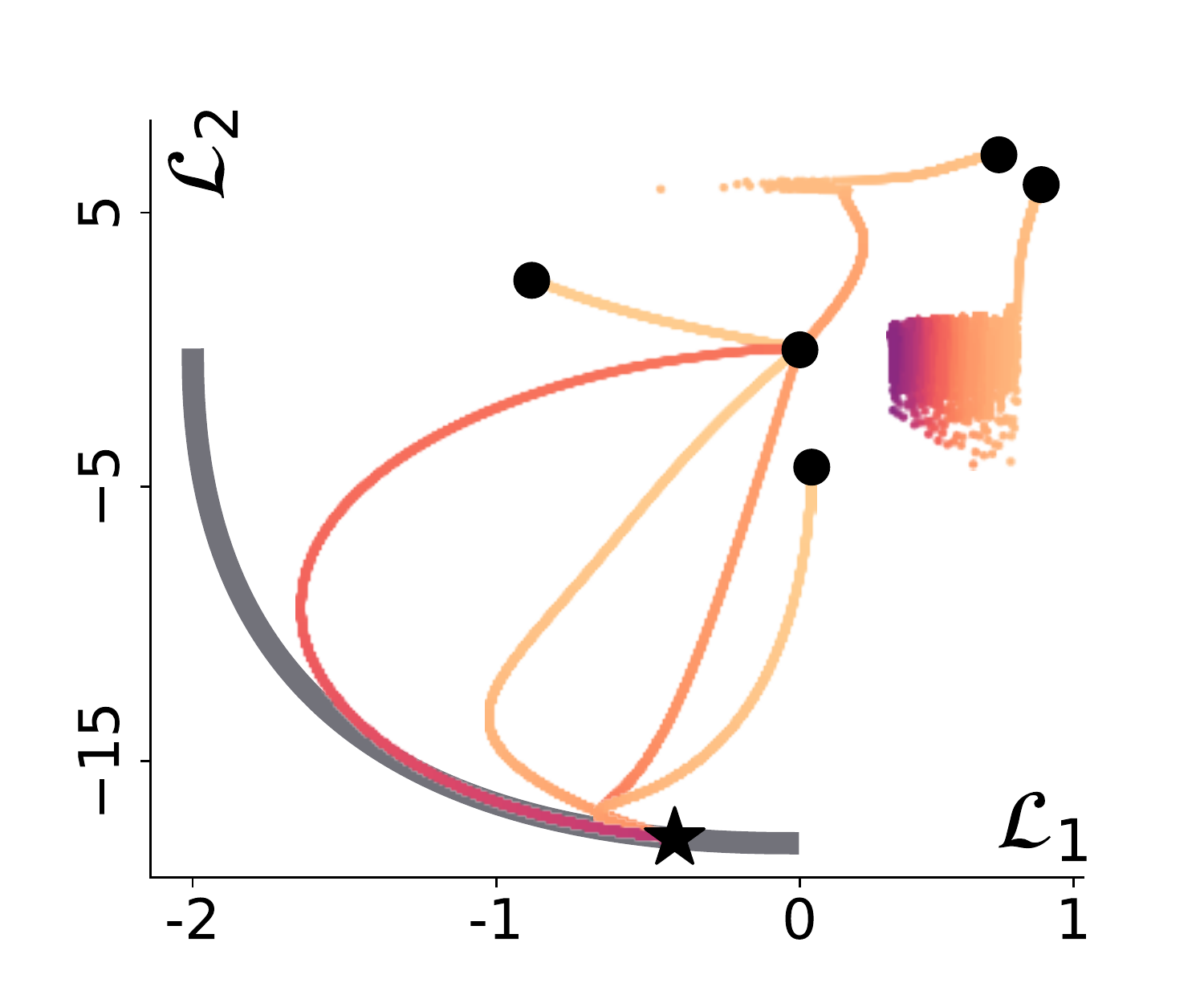} & \includegraphics[clip, trim=1.0cm 0.5cm 1.5cm 1.2cm]{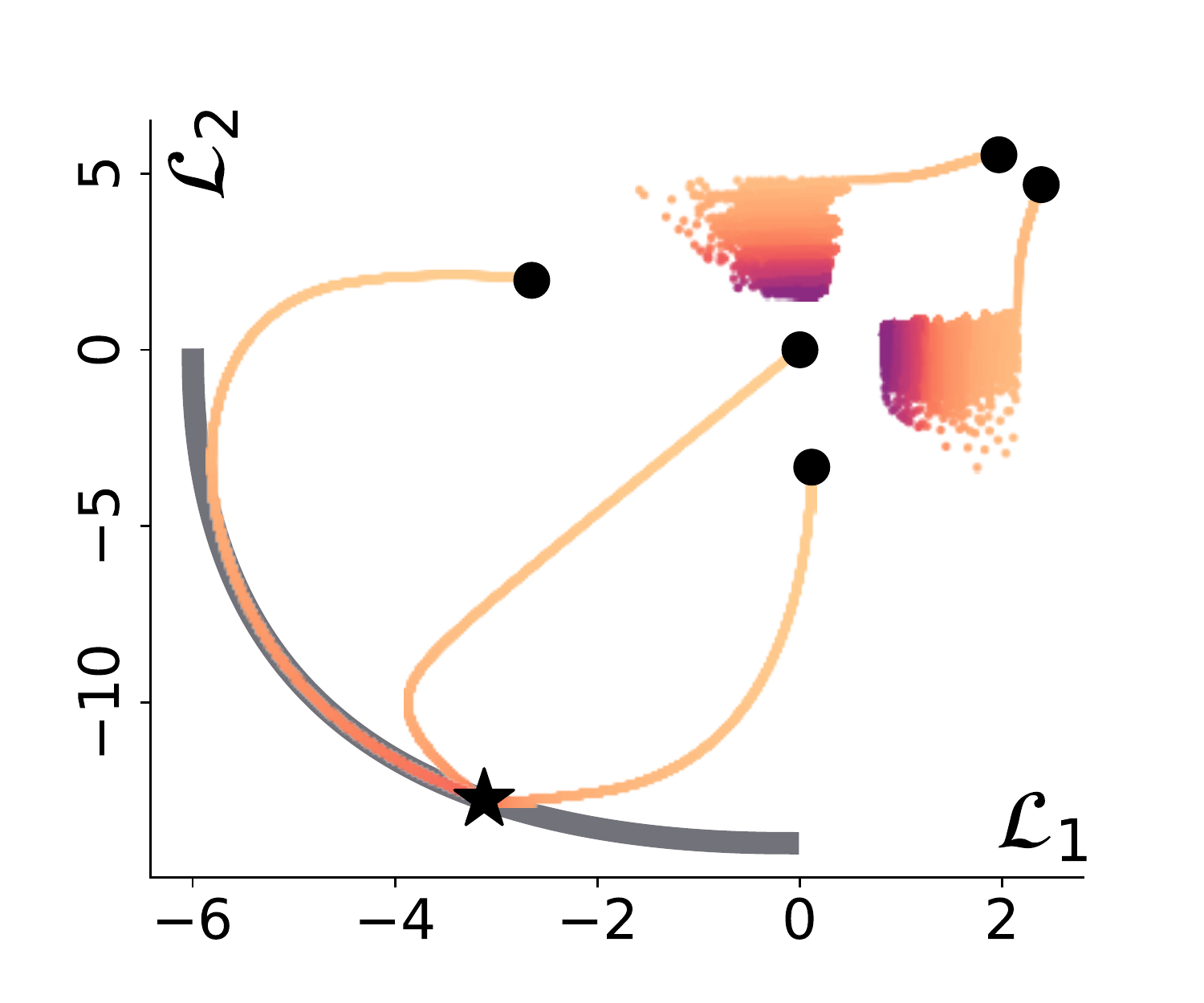} & \includegraphics[clip, trim=1.0cm 0.5cm 1.5cm 1.2cm]{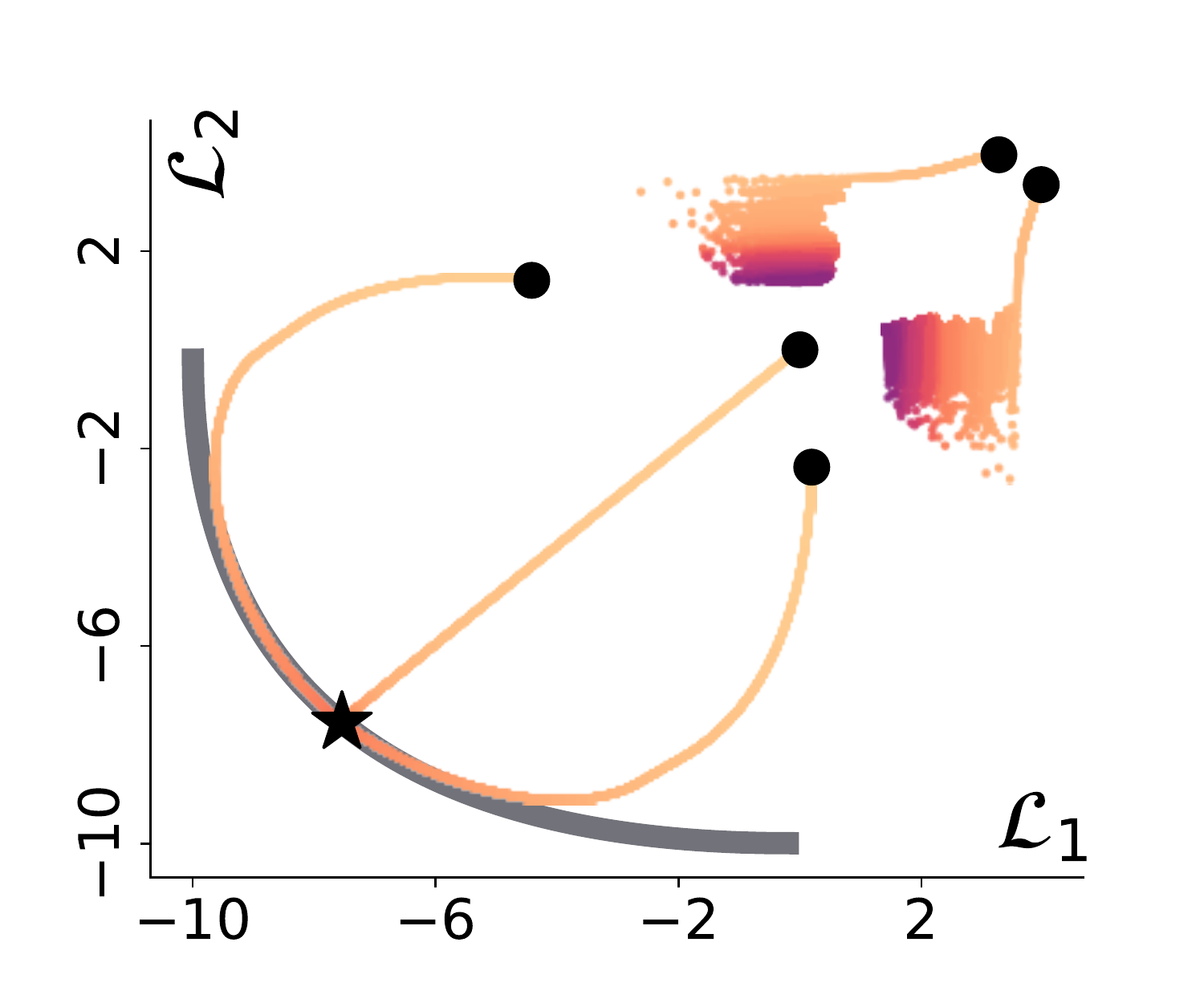} & \includegraphics[clip, trim=1.0cm 0.5cm 1.5cm 1.2cm]{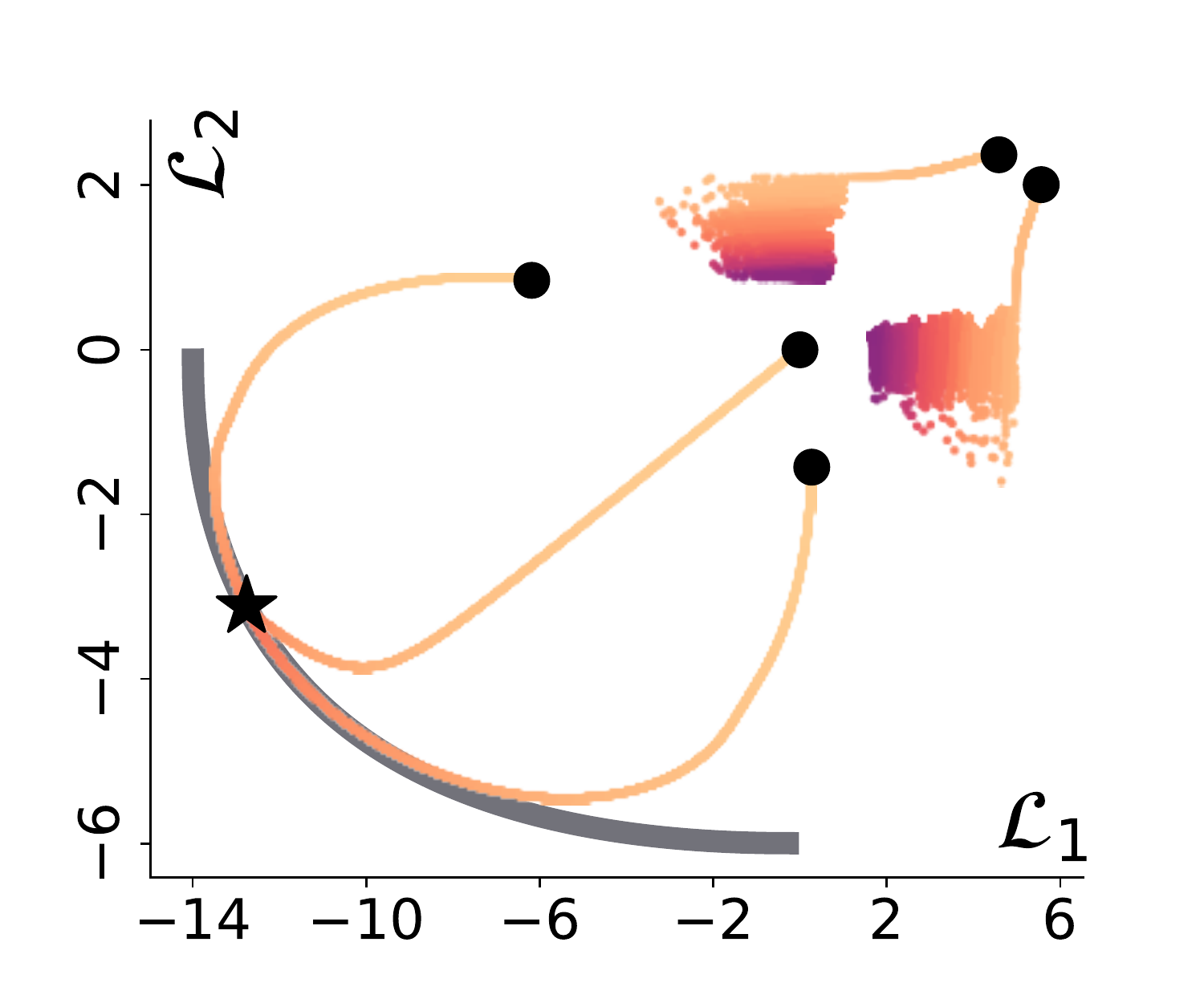} & \includegraphics[clip, trim=1.0cm 0.5cm 1.5cm 1.2cm]{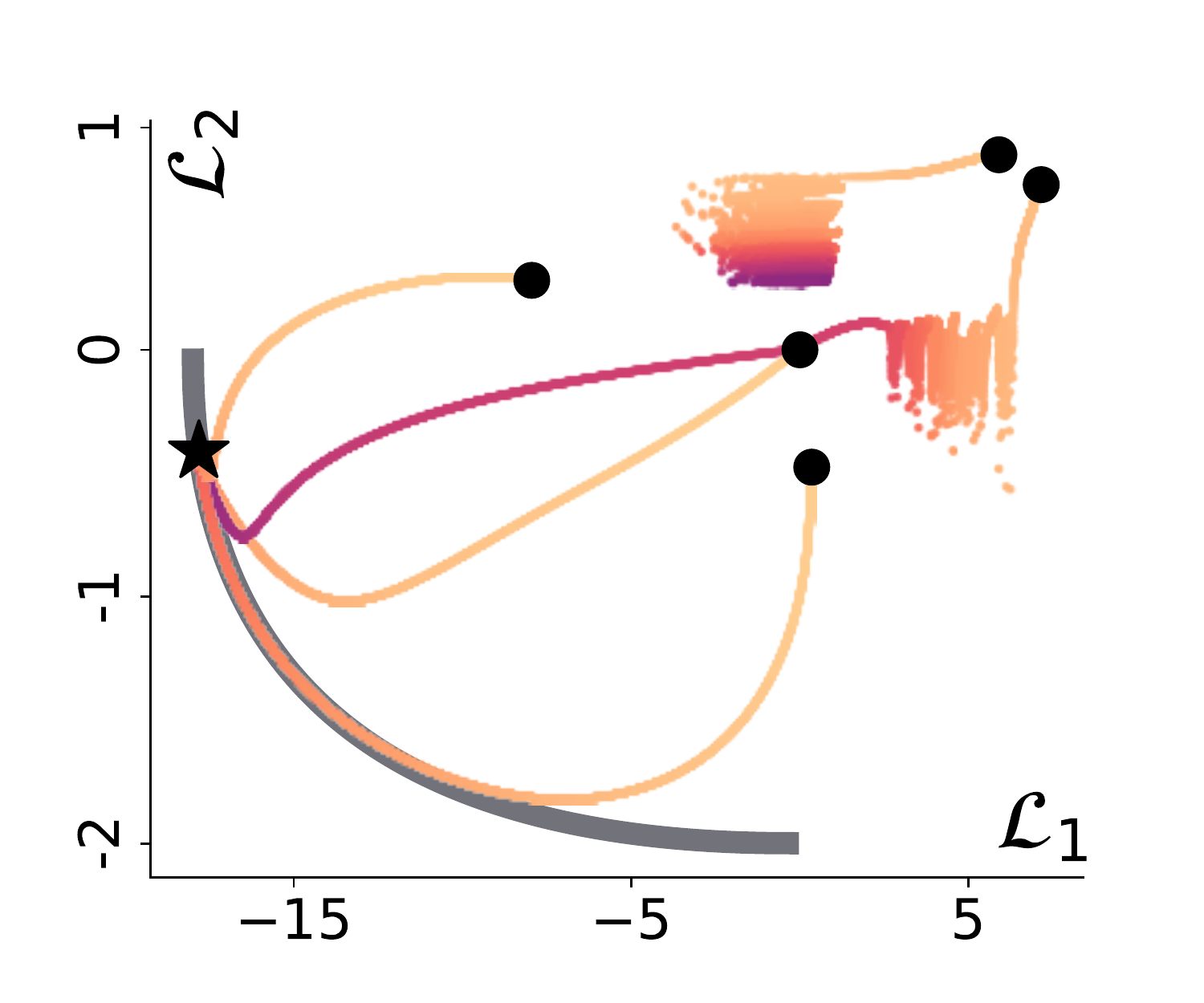} \\

\rotatebox{90}{IMTL\cite{liu2021_imtl}} & \includegraphics[clip, trim=1.0cm 0.5cm 1.5cm 1.2cm]{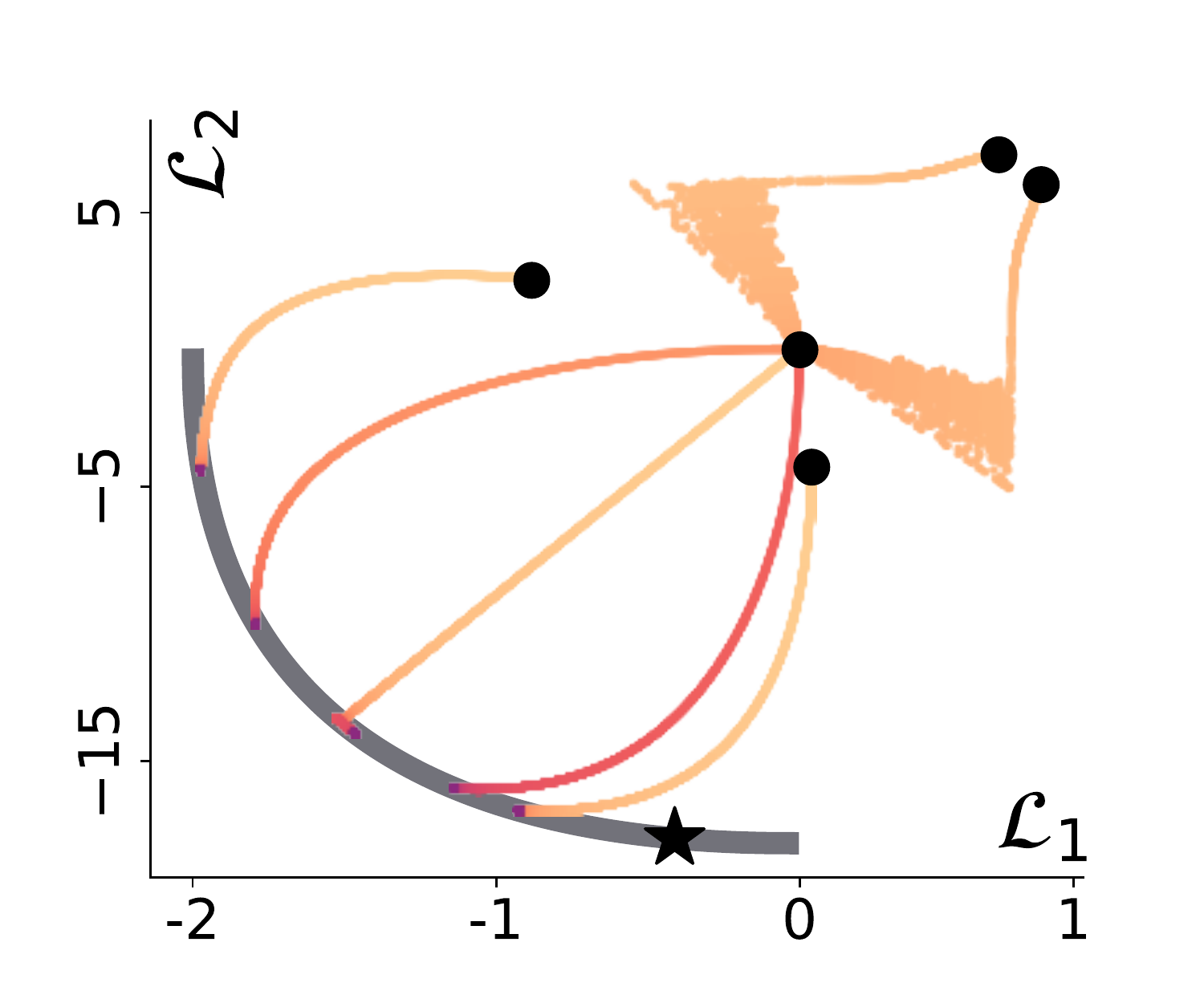} & \includegraphics[clip, trim=1.0cm 0.5cm 1.5cm 1.2cm]{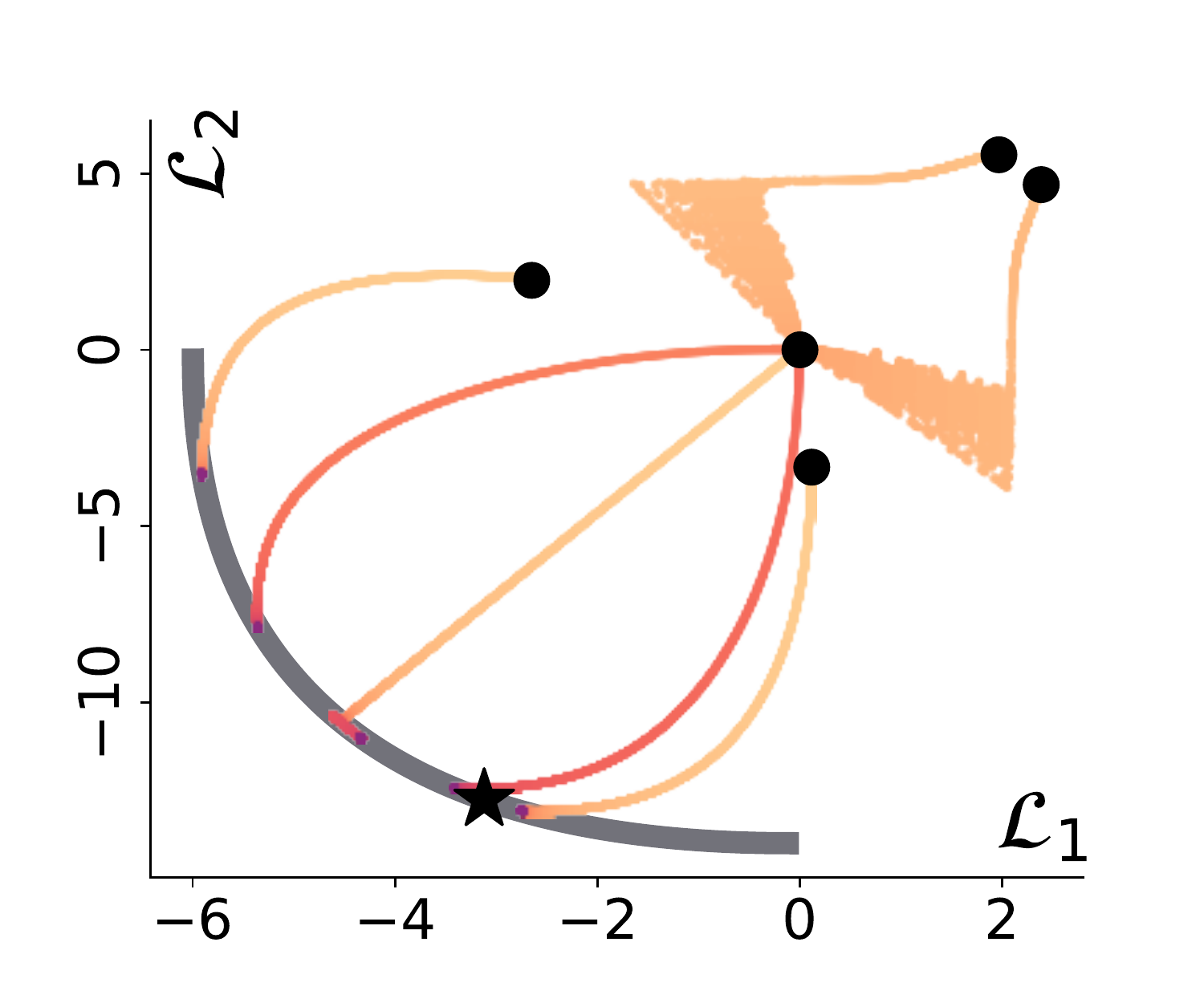} & \includegraphics[clip, trim=1.0cm 0.5cm 1.5cm 1.2cm]{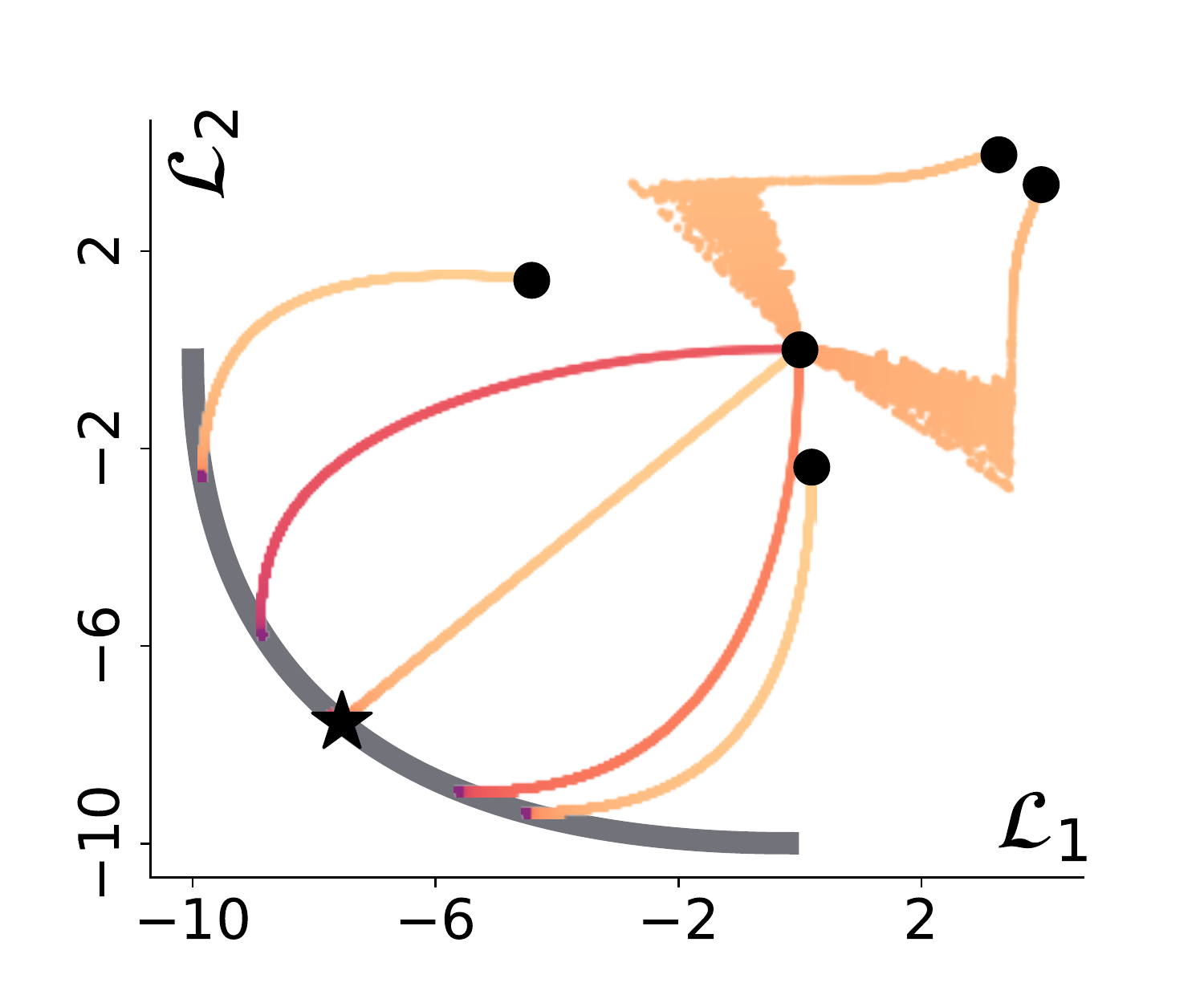} & \includegraphics[clip, trim=1.0cm 0.5cm 1.5cm 1.2cm]{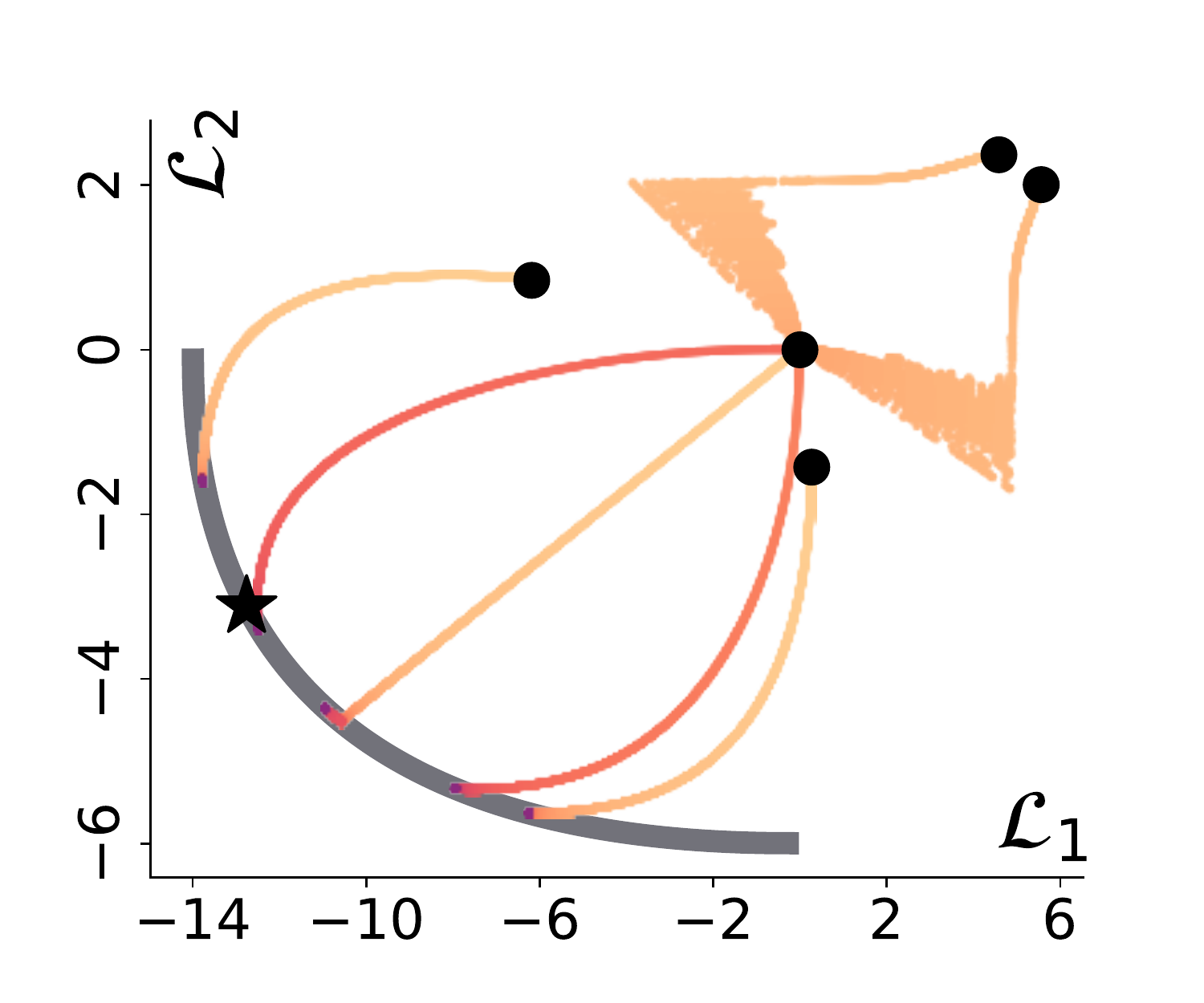} & \includegraphics[clip, trim=1.0cm 0.5cm 1.5cm 1.2cm]{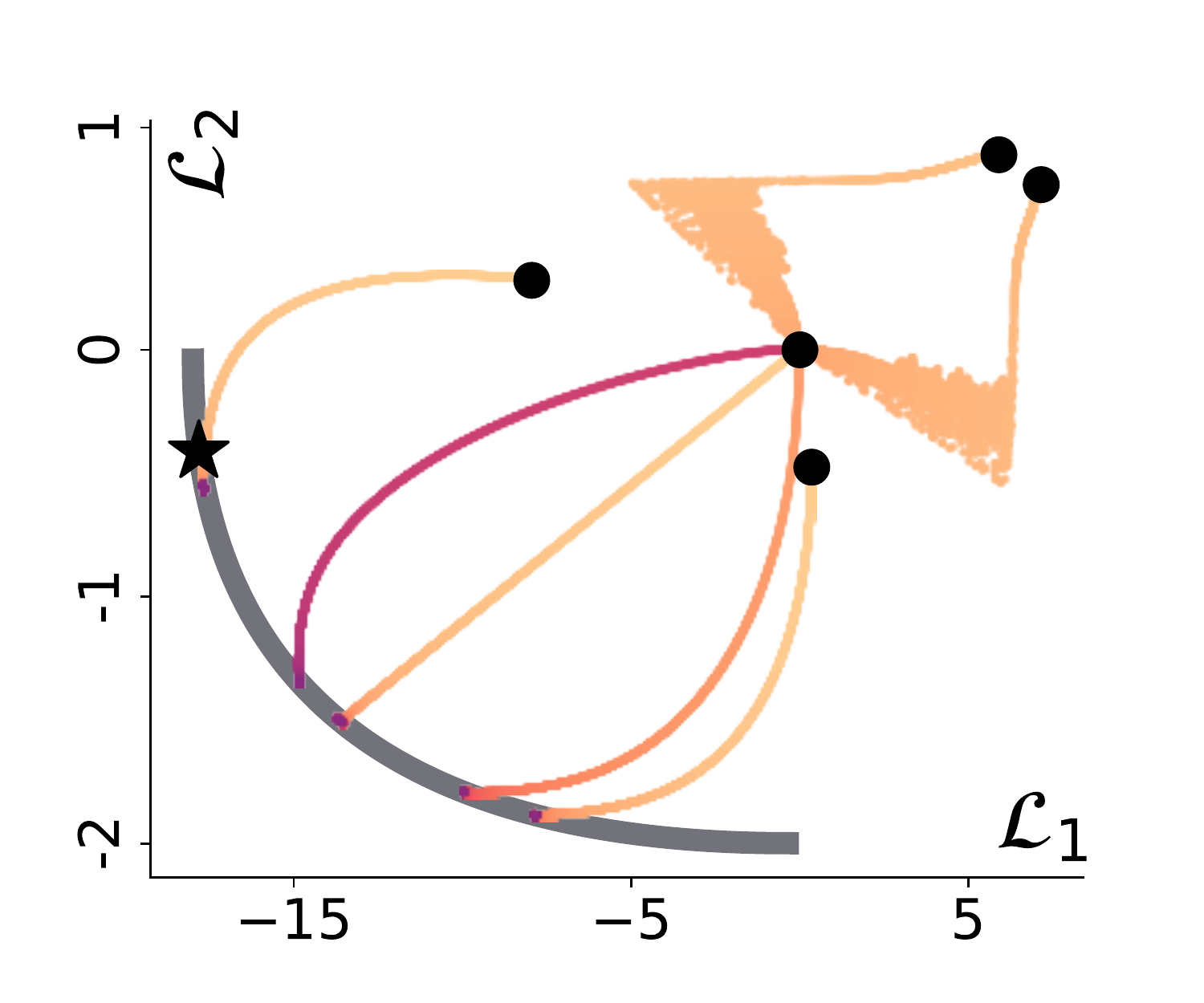} \\

\rotatebox{90}{NashMTL\cite{navon22a_nashmtl}} & \includegraphics[clip, trim=1.0cm 0.5cm 1.5cm 1.2cm]{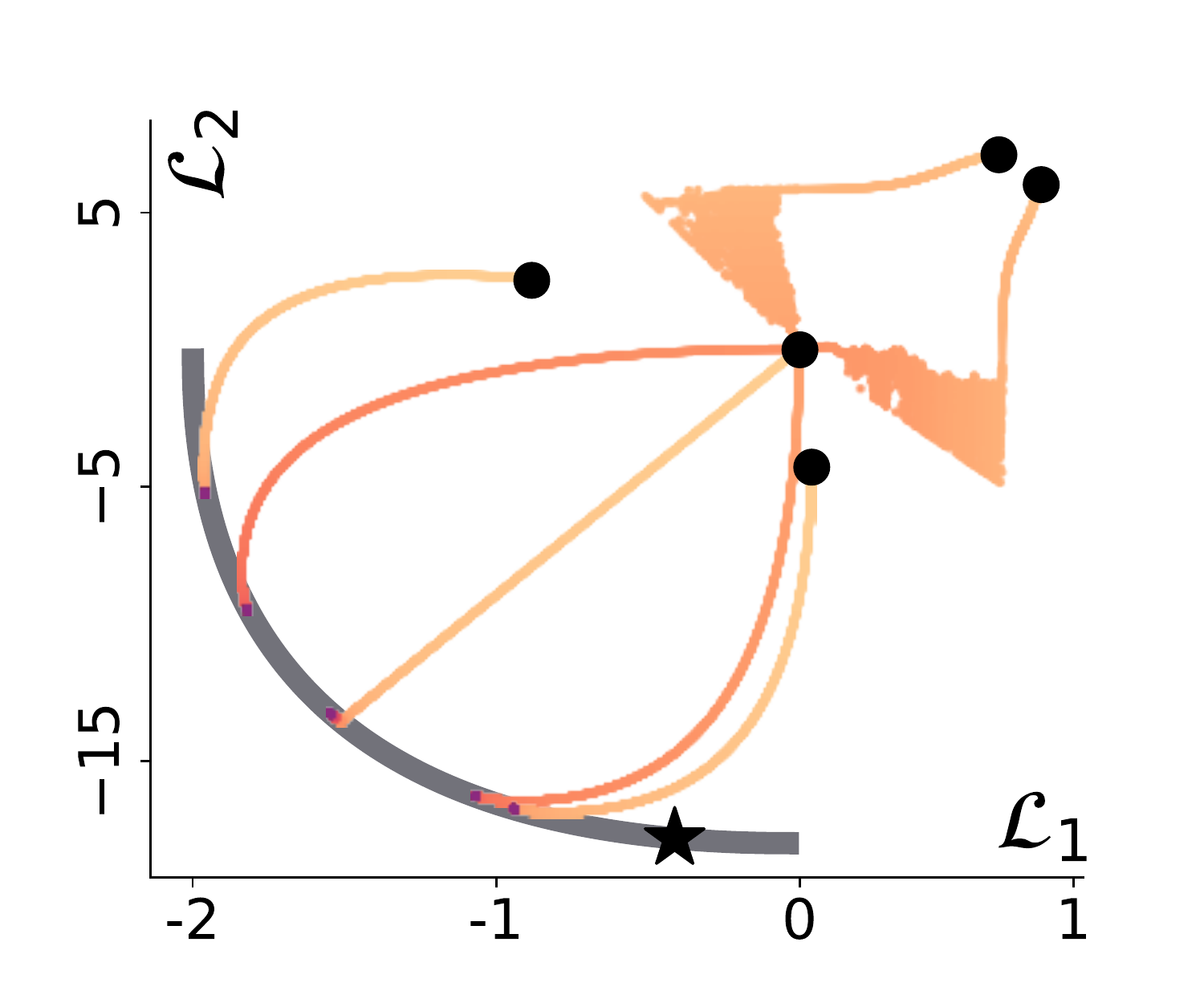} & \includegraphics[clip, trim=1.0cm 0.5cm 1.5cm 1.2cm]{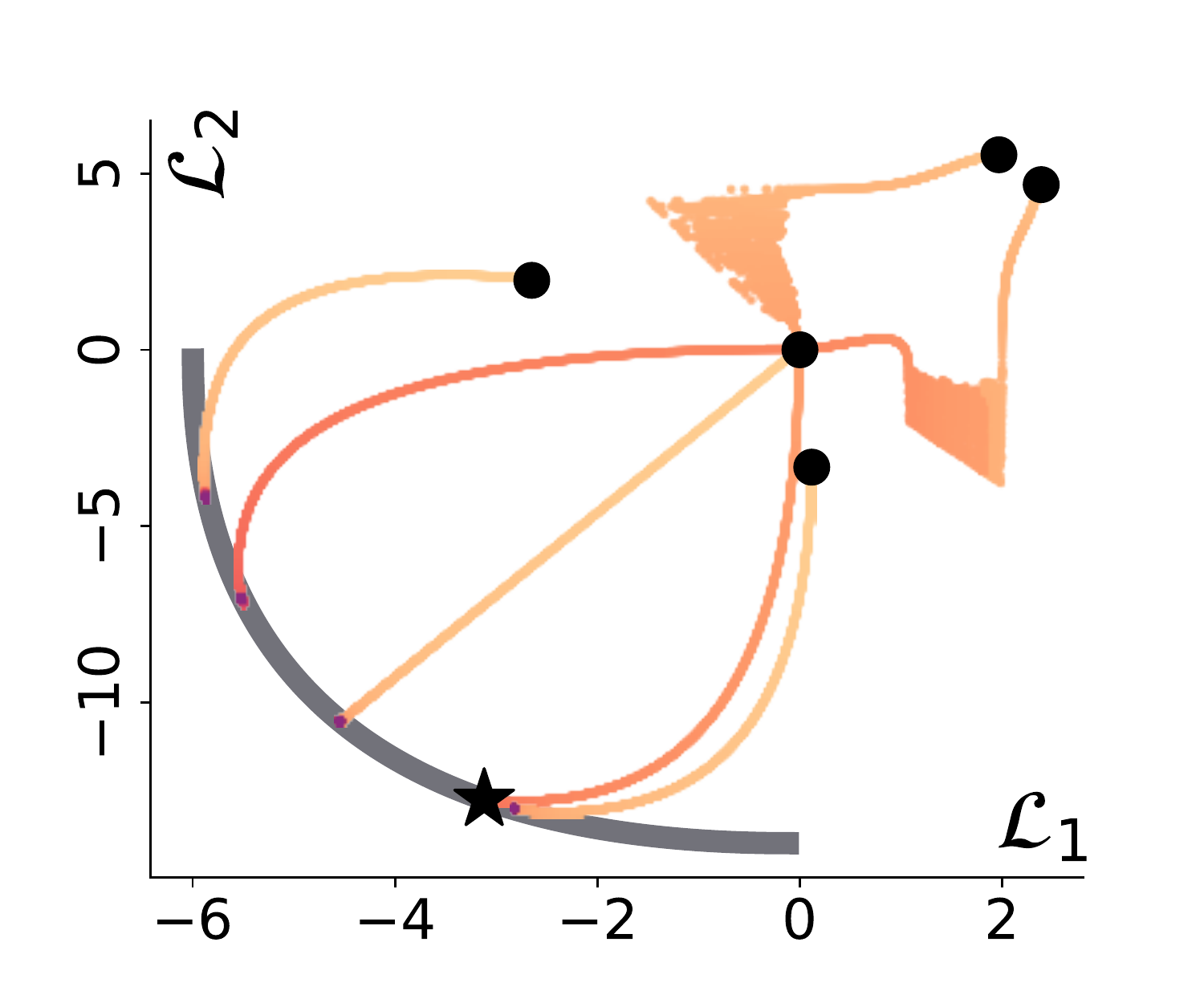} & \includegraphics[clip, trim=1.0cm 0.5cm 1.5cm 1.2cm]{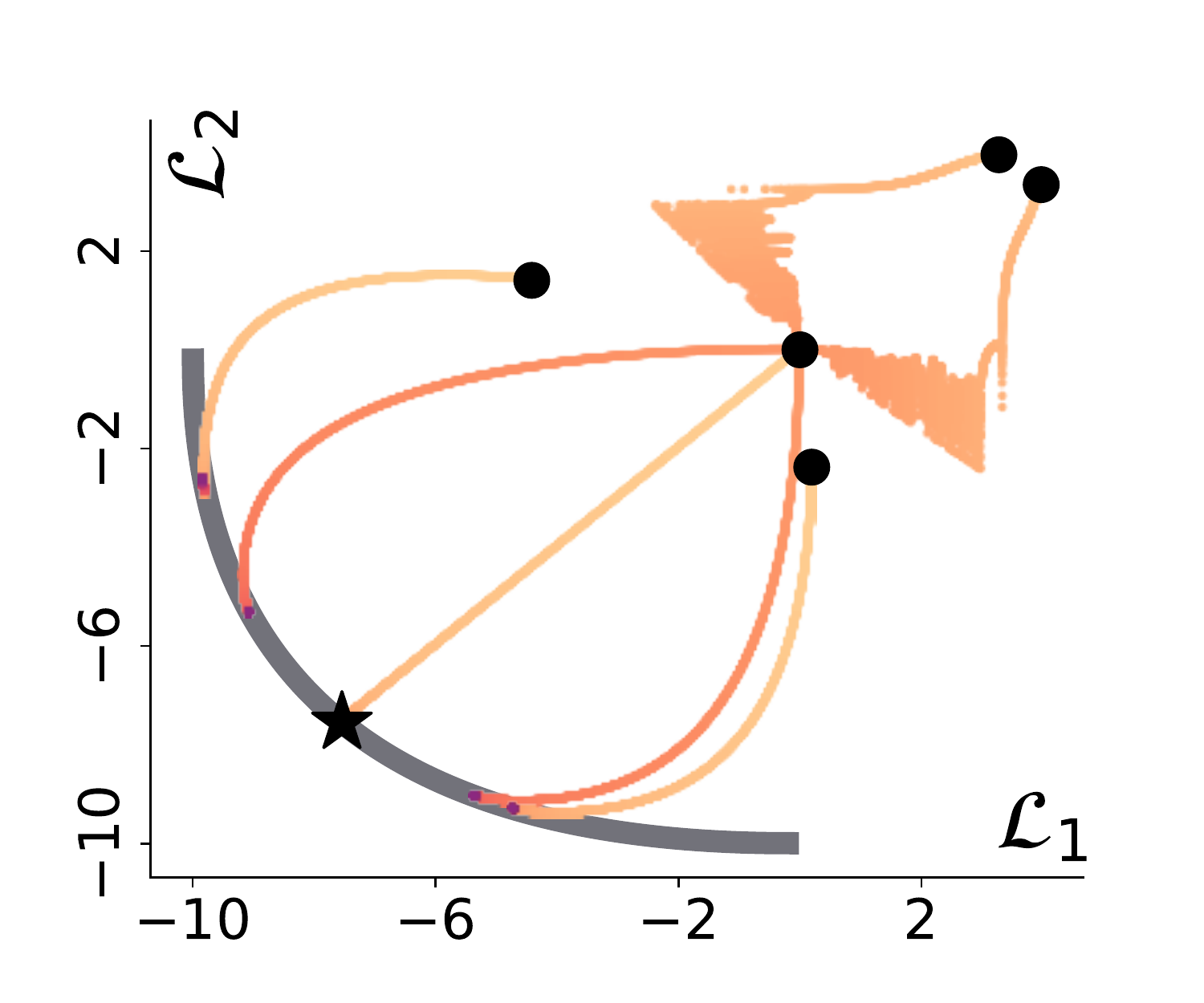} & \includegraphics[clip, trim=1.0cm 0.5cm 1.5cm 1.2cm]{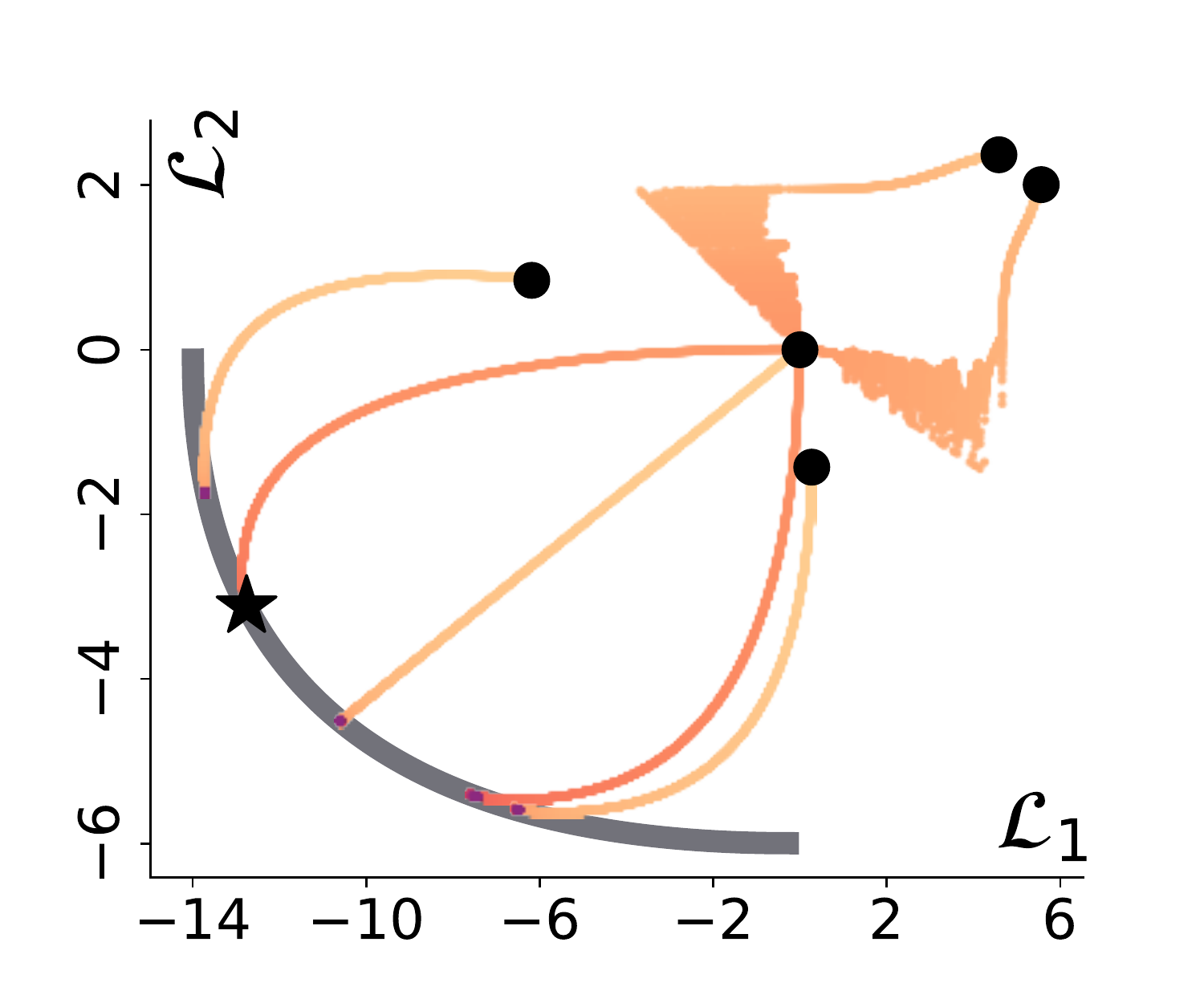} & \includegraphics[clip, trim=1.0cm 0.5cm 1.5cm 1.2cm]{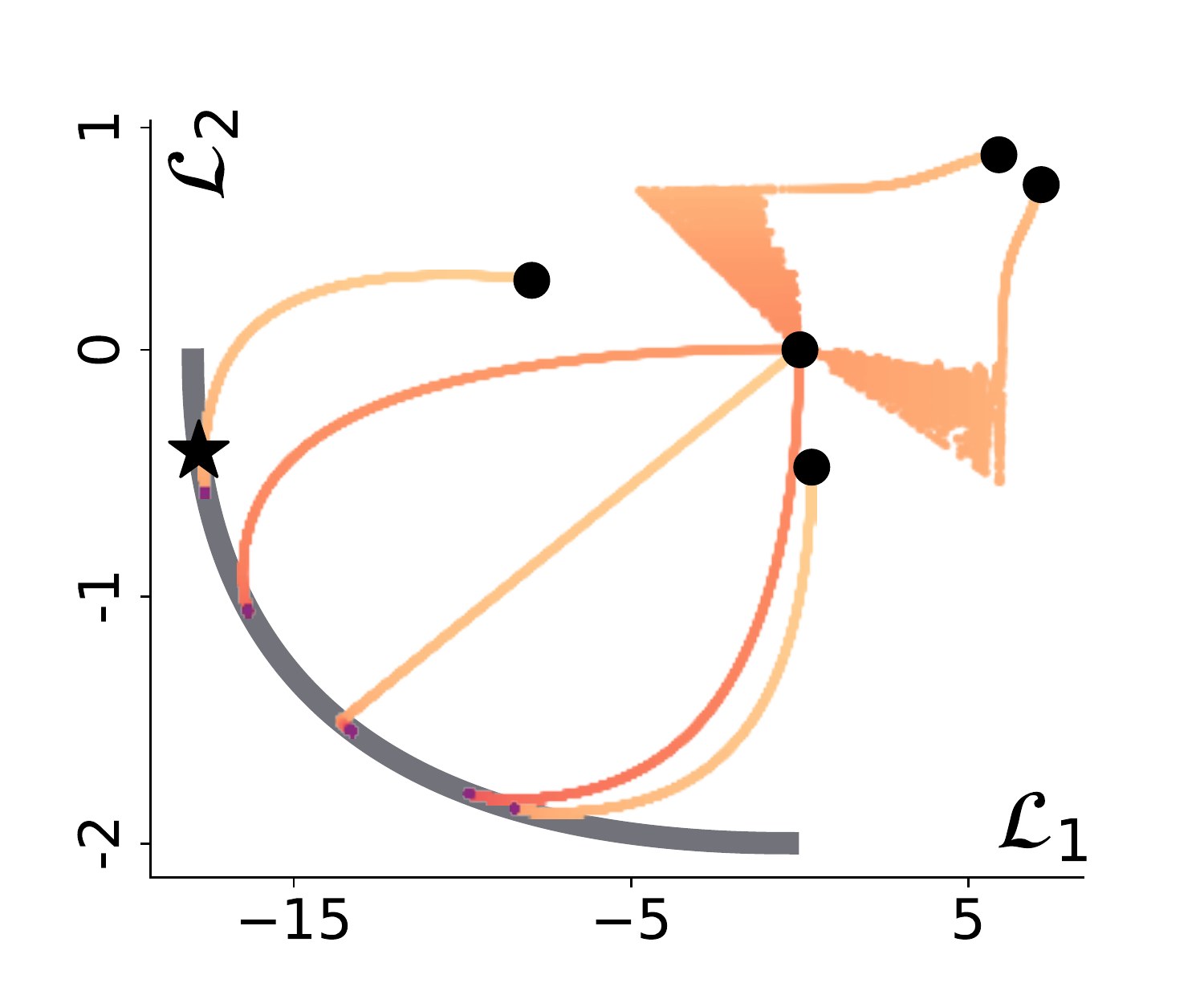} \\
\rotatebox{90}{Ours, Align-MTL} & \includegraphics[clip, trim=1.0cm 0.5cm 1.5cm 1.2cm]{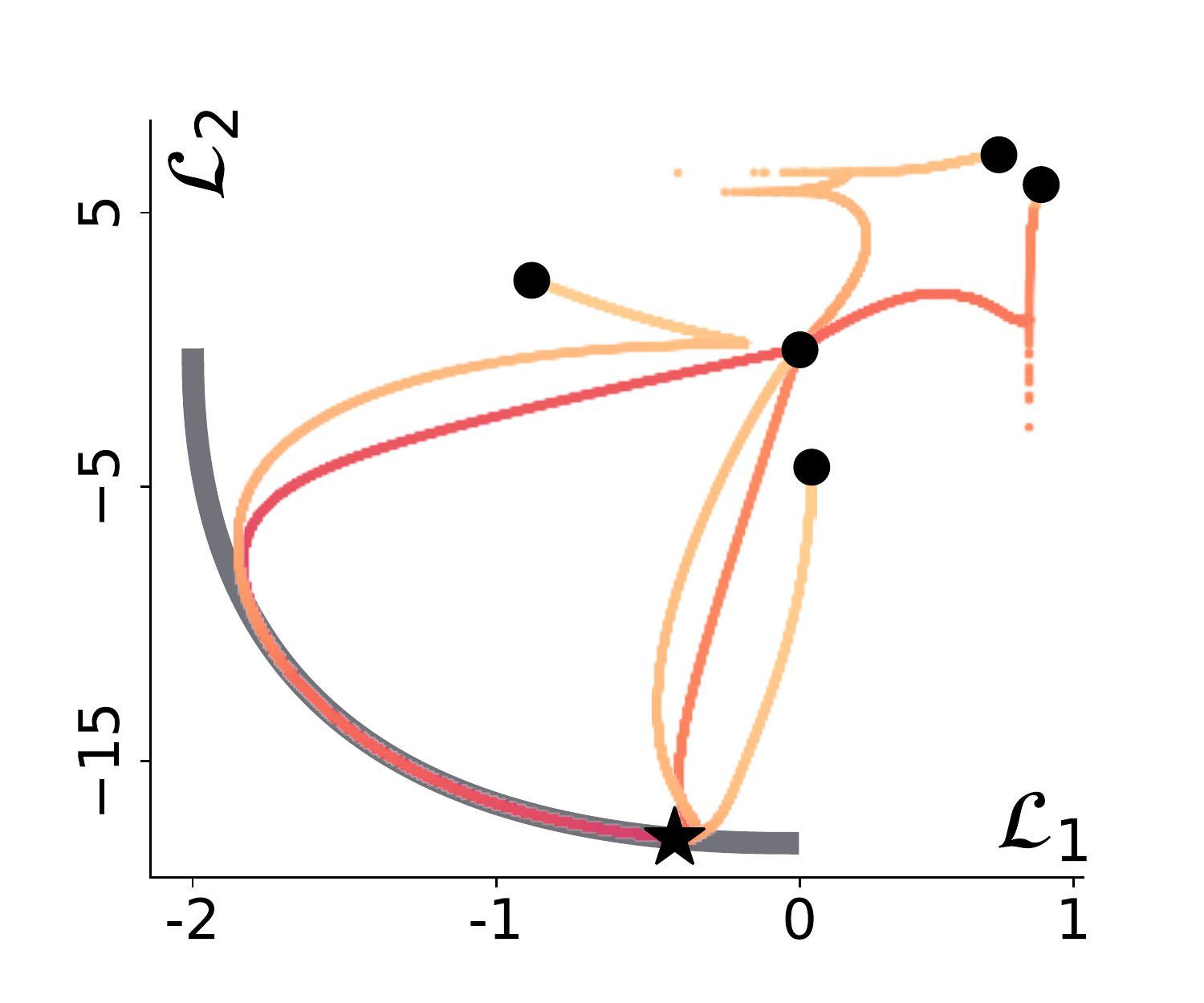} & \includegraphics[clip, trim=1.0cm 0.5cm 1.5cm 1.2cm]{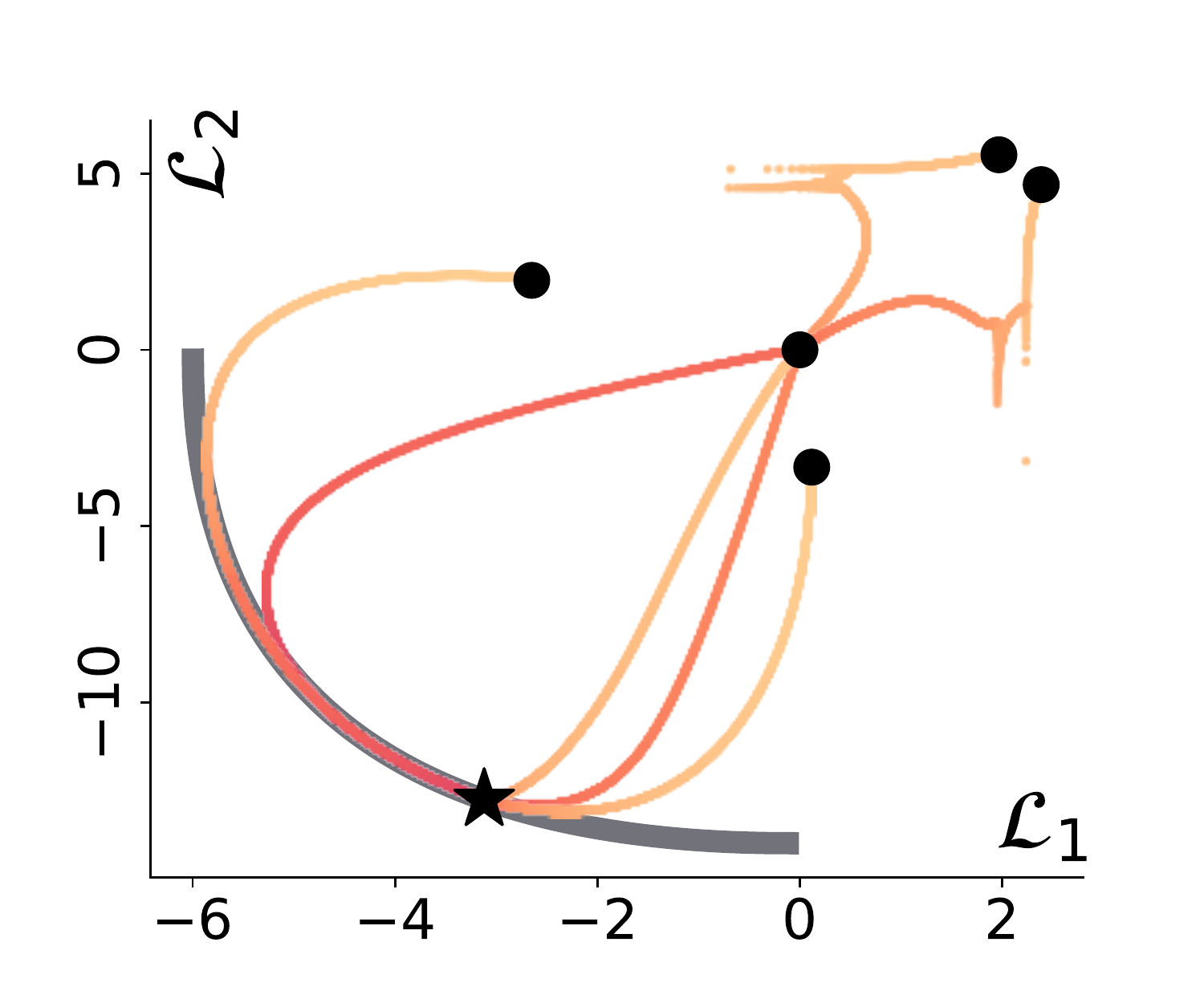} & \includegraphics[clip, trim=1.0cm 0.5cm 1.5cm 1.2cm]{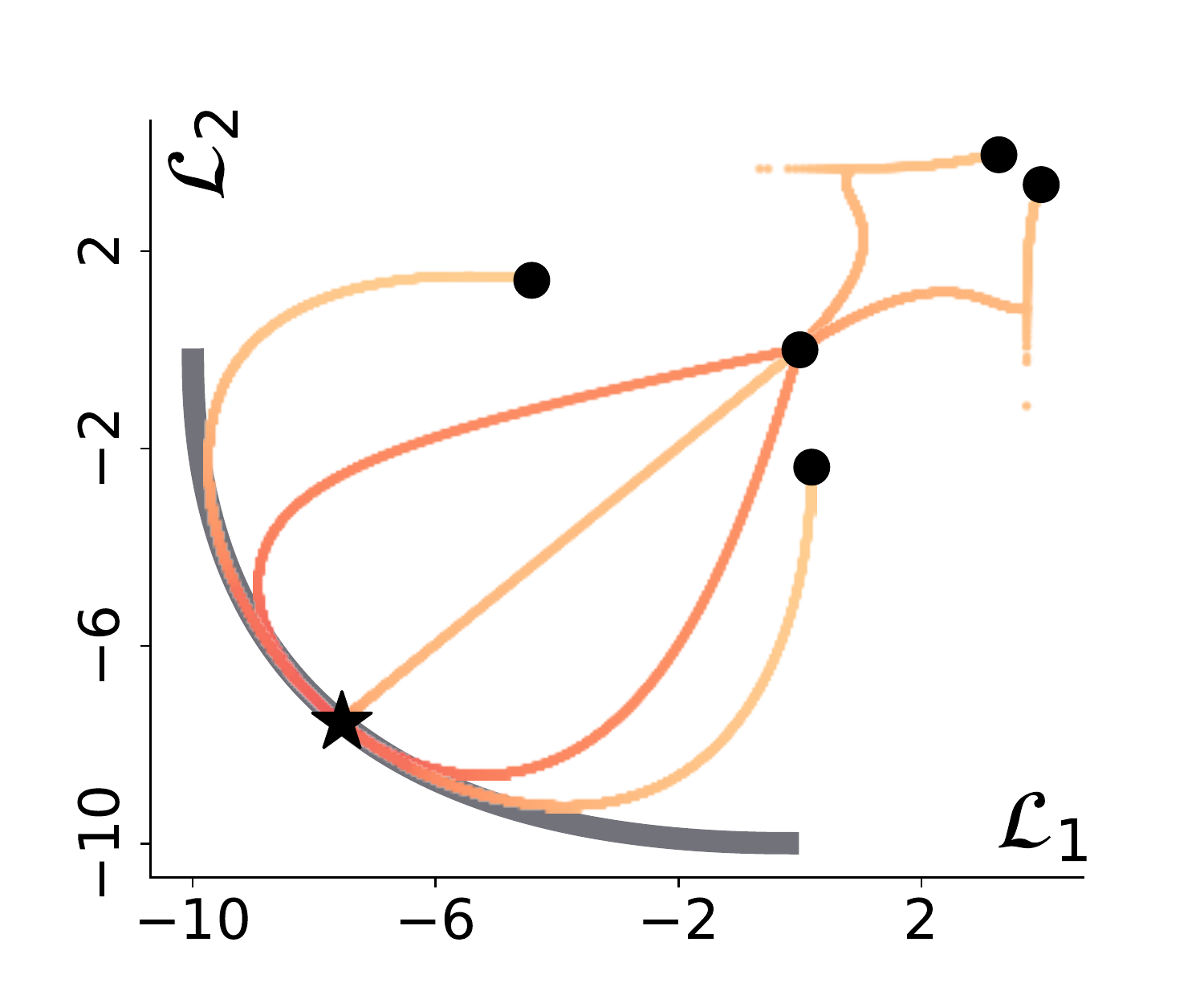} & \includegraphics[clip, trim=1.0cm 0.5cm 1.5cm 1.2cm]{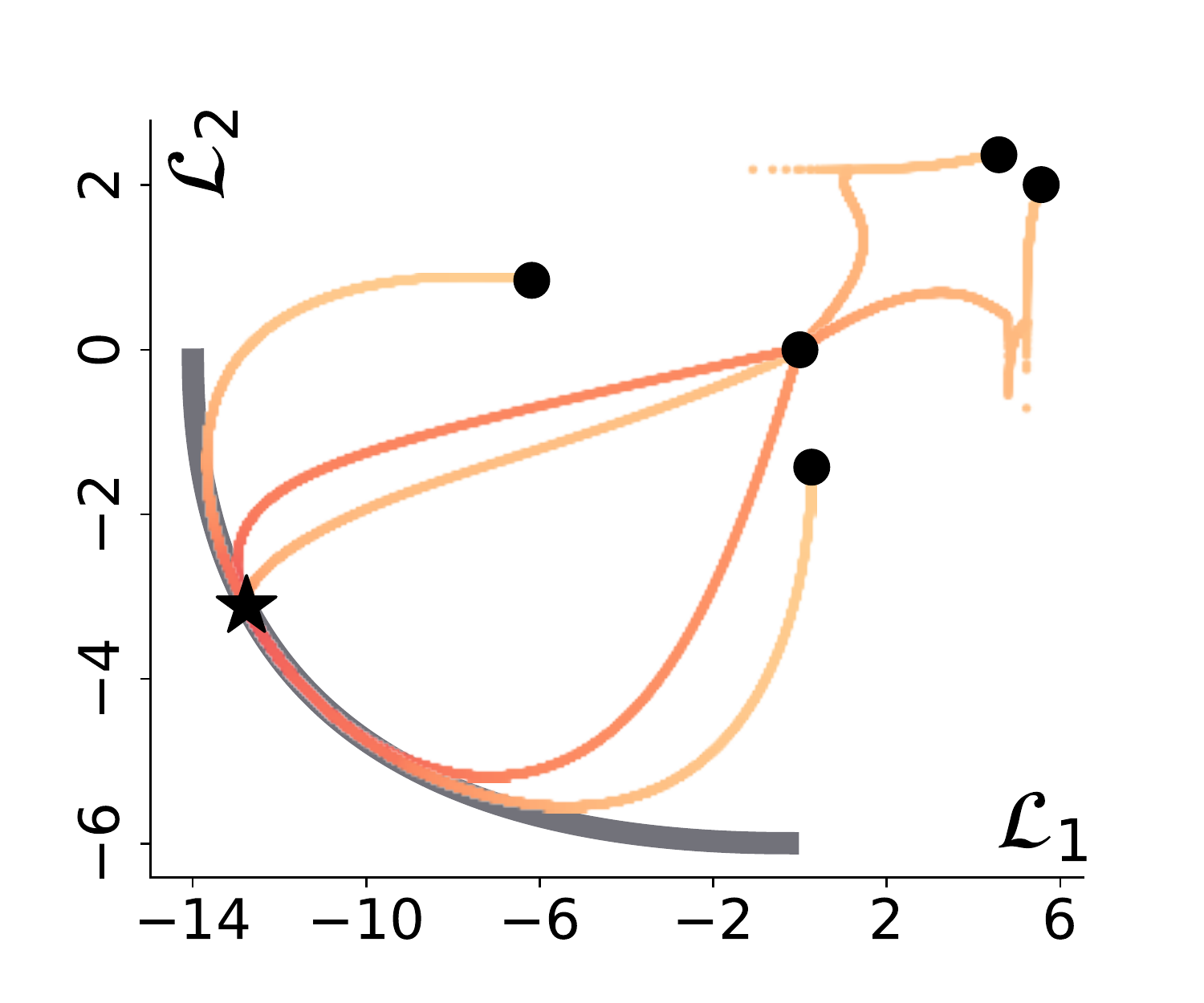} & \includegraphics[clip, trim=1.0cm 0.5cm 1.5cm 1.2cm]{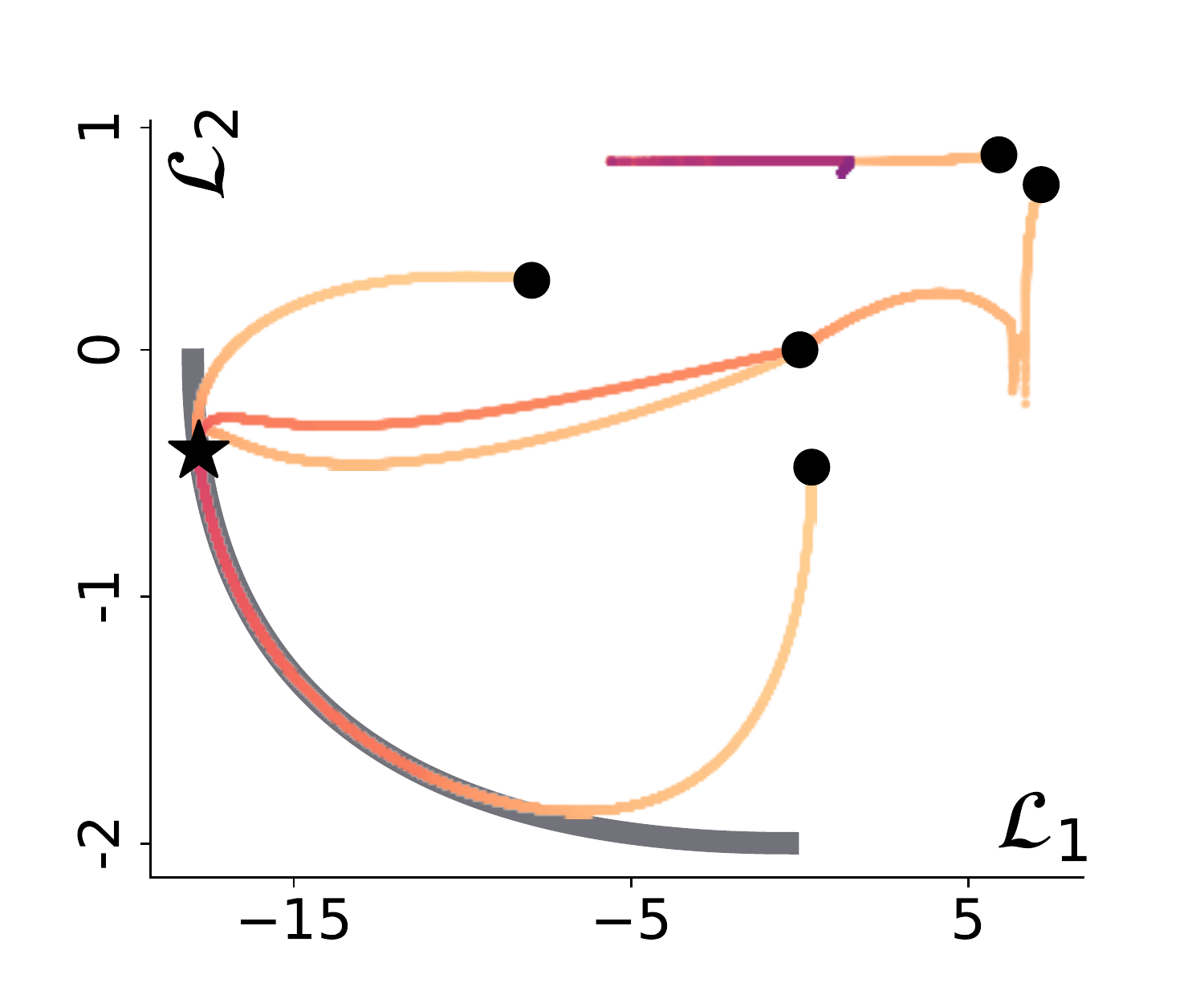} \\
\\ 
\\ 
\\
\\
\end{tabularx}
\end{figure*}

The synthetic example is a two-task objective containing areas with the presence of conflicting and dominating gradients between loss components. Formally, we use the same objective as in previous works\cite{liu2021cagrad, navon22a_nashmtl}:
\begin{gather*}
    \calL_1 = c_1(\vtheta) f_1(\vtheta) + c_2(\vtheta)g_1(\vtheta) \\ \calL_2 = c_1(\vtheta) f_2(\vtheta) + c_2(\vtheta)g_2(\vtheta) \\
    \vtheta \in \mathbb{R}^{2}
\end{gather*}
where
\begin{gather*}
    h_1(\vtheta) = \left|\frac{(-\theta_1-7)}{2}-\tanh{(-\theta_2)}\right| \\
    h_2(\vtheta) = \bigg|\frac{(-\theta_1+3)}{2}-\tanh{(-\theta_2) + 2}\bigg| \\
    c_1(\theta) = \max(\tanh\bigg(\frac{\theta_2}{2}\bigg), 0) \\
    c_2(\theta) = \max(\tanh\bigg(\frac{-\theta_2}{2}\bigg), 0) \\
    f_1(\vtheta) = \log \max \left(h_1(\vtheta), 5 \cdot 10^{-6} \right) + 6 \\
    f_2(\vtheta) = \log \max \left(h_2(\vtheta), 5 \cdot 10^{-6} \right) + 6\\
    g_1(\vtheta) = \frac{(-\theta-7)^2 + 0.1(-\theta_2-8)^2}{10} - 20 \\
    g_2(\vtheta) = \frac{(-\theta+7)^2 + 0.1(-\theta_2-8)^2}{10} - 20 \\
\end{gather*}
We perform minimization starting from five initial points: $[-8.5, 7.5], [0.0, 0.0], [9.0, 9.0], [-7.5, -0.5], [9, -1.0]$. We use Adam~\cite{adam2015} optimizer with learning rate~$10^{-3}$ and optimize for 35k iterations. We demonstrate that our method is able to converge to the optimums with varying pre-defined task weights in \cref{fig: synthetic-varying}. For this purpose we explore a number of task convex combinations, such that $\mathcal{L}_0 = \alpha \mathcal{L}_1 + (1-\alpha) \mathcal{L}_2$

%% file: appendix/05_impl_details.tex
\section{Implementation details}

\noindent\textbf{\textsc{CityScapes} three-task.} Following MGDA-UB training setup \cite{sener2018MGDAUB}, we train PSPNet\cite{zhao2017} model for $100$ epochs using Adam optimizer with learning rate $10^{-4}$. Train batch size is set to $8$. Images from training set are resized into $512\times 256$ resolution. We augment training set using random rotation and horizontal flips. The performance is averaged across $3$ random initializations. 

\noindent\textbf{\textsc{CityScapes} two-task.} We follow CAGrad \cite{liu2021cagrad} training setup and train MTAN \cite{liu2019DWA} model. Semantic labels are groupped into $7$ classes. Batch size is set to $8$, learning rate of Adam optimizer is set to $10^{-4}$. Models are trained for $200$ epochs and learning rate is halved after $100$ epochs. The performance is averaged over last $10$ epochs and $3$ random seeds.

\begin{figure*}[!p]
\centering
\caption{Empirical evaluation of a stability criterion. We plot a condition number, gradient magnitude similarity~\cite{tianhe2020pcgrad}, minimal cosine between gradient pairs (conflicts) and maximum gradient norm ratio, \ie $\max_{i\neq j} \{ \|\vg_i\| / \|\vg_j\| \}$, during training of PSPNet~\cite{zhao2017, sener2018MGDAUB} and MTAN~\cite{liu2019DWA} on the \textsc{NYUv2} benchmark. Unlike Cityscapes with three tasks (figure from the main paper), on NYUv2 gradients do not differ drastically in magnitudes but tend to have more conflicts (the cosine between gradients are negative, except for PCGrad). These figures indicate a high correlation between condition number, gradient norm ratios and gradient magnitude similarity. Our Aligned-MTL approach eliminates dominance ($\kappa = 1$, $r = 1$, $GMS = 1$) and conflicts ($\min_{i\neq j} cos(g_i,g_j) = 0$) by design.}
\label{tab:nyu-metrics}

\includegraphics[width=0.90\textwidth]{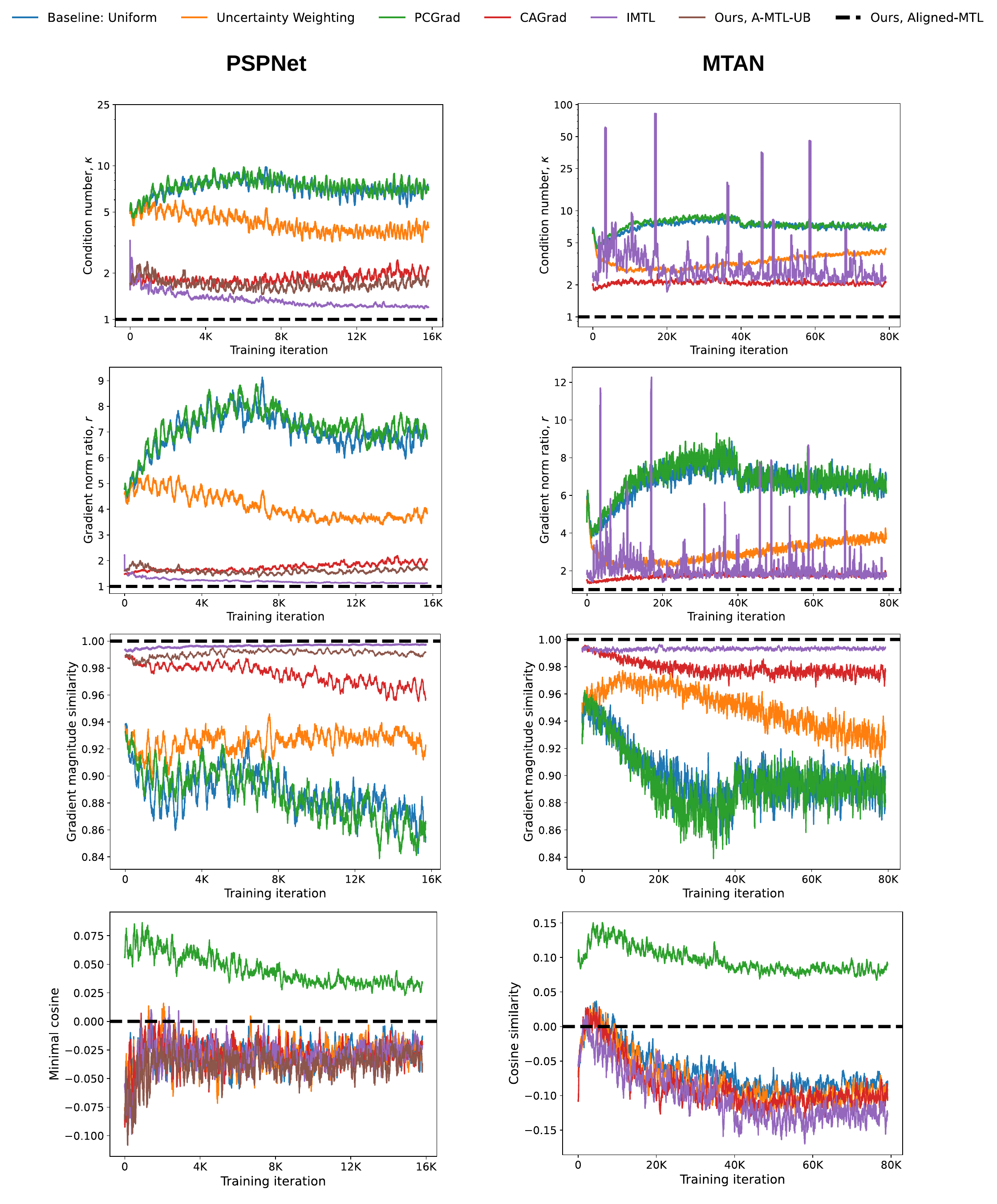}
\end{figure*}

\noindent\textbf{\textsc{NYUv2} three-task.} \cite{liu2019DWA, liu2021cagrad, navon22a_nashmtl} We train both PSPNet models \cite{zhao2017, sener2018MGDAUB} and MTAN \cite{liu2019DWA} models in our training setup with the same hyperparameters set. We use Adam \cite{adam2015} optimizer with learning rate $10^{-4}$. Models are trained for $200$ epochs and batch size $2$. Images from training set are randomly scaled and cropped into $384 \times 288$ resolution. The performance is averaged across $3$ random seeds. 

\noindent\textbf{Reinforcement learning.} We follow CAGrad~\cite{liu2021cagrad} and use the implementation originally proposed and developed by \cite{pmlr-v139-sodhani21a-mtlrl}. The execution config was adapded from CAGrad~\cite{liu2021cagrad}. The global evaluation pipeline is similar to previous works~\cite{navon22a_nashmtl, liu2021cagrad}. The performance is averaged over 10 random seeds.